\documentclass[10pt,a4paper,oneside]{article}
\usepackage[top=25mm, bottom=25mm, left=20mm, right=20mm, a4paper]{geometry}

\usepackage[sort&compress,numbers]{natbib}
\usepackage{authblk}

\usepackage{amssymb}
\usepackage{placeins}
\usepackage[dvipsnames]{xcolor}
\usepackage{xspace}
\usepackage{hyperref}
\usepackage[acronym]{glossaries}
\usepackage[normalem]{ulem}
\usepackage{algorithm2e}
\usepackage{algpseudocode}
\usepackage{mathtools}

\newcommand{\ie}{{i.e.}\xspace}
\newcommand{\Ie}{{I.e.}\xspace}
\newcommand{\cf}{{cf.}\xspace}
\newcommand{\wrt}{{w.r.t.}\xspace}
\newcommand{\eg}{{e.g.}\xspace}
\newcommand{\Eg}{{E.g.}\xspace}
\newcommand{\etc}{{etc.}\xspace}

\newcommand{\x}{\boldsymbol{x}}

\def\RR{\mathrm{R}}

\glsdisablehyper
\makenoidxglossaries

\newacronym{ai}{AI}{Artificial Intelligence}
\newacronym{auc}{AUC}{Area Under Curve}
\newacronym{bd}{BD}{Backdoor}
\newacronym{cav}{CAV}{Concept Activation Vector}
\newacronym{cap}{CAP}{Class artifact Projection}
\newacronym{cd}{CD}{Contextual Decomposition}
\newacronym{cdep}{CDEP}{Contextual Decomposition Explanation Penalization}
\newacronym{clarc}{ClArC}{Class Artifact Compensation}
    \newacronym{aclarc}{A-ClArC}{Augmentative Class Artifact Compensation}
    \newacronym{pclarc}{P-ClArC}{Projective Class Artifact Compensation}
\newacronym{ch}{CH}{Clever Hans}
\newacronym{dnn}{DNN}{Deep Neural Network}
\newacronym{fda}{FDA}{Fisher Discriminant Analysis}
\newacronym{lrp}{LRP}{Layer-wise Relevance Propagation}
\newacronym{ml}{ML}{Machine Learning}
\newacronym{roc}{ROC}{Receiver Operating Characteristic}
\newacronym{rrr}{RRR}{Right for the Right Reasons}
\newacronym{sc}{SC}{Spectral Clustering}
\newacronym{ss}{SpeSig}{Spectral Signature}
\newacronym{sgd}{SGD}{Stochastic Gradient Descent}
\newacronym{spray}{SpRAy}{Spectral Relevance Analysis}
\newacronym{xai}{XAI}{eXplainable Artificial Intelligence}
\newacronym{xil}{XIL}{eXplanatory Interactive Learning}

\definecolor{cmnist_0}{RGB}{0, 0, 255}
\definecolor{cmnist_1}{RGB}{0, 128, 255}
\definecolor{cmnist_2}{RGB}{0, 255, 255}
\definecolor{cmnist_3}{RGB}{0, 255, 0}
\definecolor{cmnist_4}{RGB}{255, 255, 0}
\definecolor{cmnist_5}{RGB}{255, 128, 0}
\definecolor{cmnist_6}{RGB}{255, 0, 0}
\definecolor{cmnist_7}{RGB}{255, 0, 127}
\definecolor{cmnist_8}{RGB}{255, 0, 255}
\definecolor{cmnist_9}{RGB}{127, 0, 255}

\begin{document}

\title{
Finding and Removing Clever Hans:\\
Using Explanation Methods to Debug and Improve Deep Models
}

\renewcommand*{\thefootnote}{\fnsymbol{footnote}}

\author[1]{Christopher~J.~Anders\footnote{contributed equally}}
\newcommand\CoAuthorMark{\footnotemark[\arabic{footnote}]}
\author[2,3]{Leander~Weber\protect\CoAuthorMark}
\author[3]{David~Neumann}
\author[3]{Wojciech~Samek}
\author[1,4,5]{Klaus-Robert~Müller}
\author[3]{Sebastian~Lapuschkin}

\affil[1]{\footnotesize Machine Learning Group, Technische Universit\"at Berlin, 10587 Berlin, Germany}
\affil[2]{\footnotesize Media Technology Group, Technische Universit\"at Berlin, 10587 Berlin, Germany}
\affil[3]{\footnotesize Department of Artificial Intelligence, Fraunhofer Heinrich Hertz Institute, 10587 Berlin, Germany}
\affil[4]{\footnotesize Department of Artificial Intelligence, Korea University, Seoul 136-713, Korea}
\affil[5]{\footnotesize Max Planck Institut f\"ur Informatik, 66123 Saarbr\"ucken, Germany}

\maketitle

\begin{abstract}
    Contemporary learning models for computer vision are typically trained on very large (benchmark) datasets with millions of samples.
    These may, however, contain biases, artifacts,
    or errors that have gone unnoticed and are exploitable by the model.
    In the worst case, the trained model does not learn a valid and generalizable strategy to solve the problem it was trained for,
    and becomes a `Clever-Hans' predictor that bases its decisions on spurious correlations in the training data,
    potentially yielding an unrepresentative or unfair, and possibly even hazardous predictor.
    In this paper, we contribute by providing a comprehensive analysis framework based on a scalable statistical analysis of attributions from explanation methods for large data corpora.
    Based on a recent technique -- Spectral Relevance Analysis -- we propose the following technical contributions and resulting findings:
    (a) a scalable quantification of artifactual and poisoned classes where the machine learning models under study exhibit Clever-Hans behavior,
    (b) several approaches we collectively denote as Class Artifact Compensation,
    which are able to effectively and significantly reduce a model's \glsdesc{ch} behavior.
    \Ie, we are able to \emph{un-Hans} models trained on (poisoned) datasets,
    such as the popular ImageNet data corpus. 
    We demonstrate that Class Artifact Compensation,
    defined in a simple theoretical framework,
    may be implemented as part of a Neural Network's training or fine-tuning process,
    or in a post-hoc manner by injecting additional layers, preventing any further propagation of undesired \glsdesc{ch} features,
    into the network architecture.
    Using our proposed methods,
    we provide qualitative and quantitative analyses of the biases and artifacts in, \eg, the ImageNet dataset,
    the Adience benchmark dataset of unfiltered faces
    and the ISIC 2019 skin lesion analysis dataset.
    We demonstrate that these insights can give rise to improved,
    more representative and fairer models operating on implicitly
    cleaned data corpora.

    \vspace{5mm}
    \includegraphics[width=\linewidth]{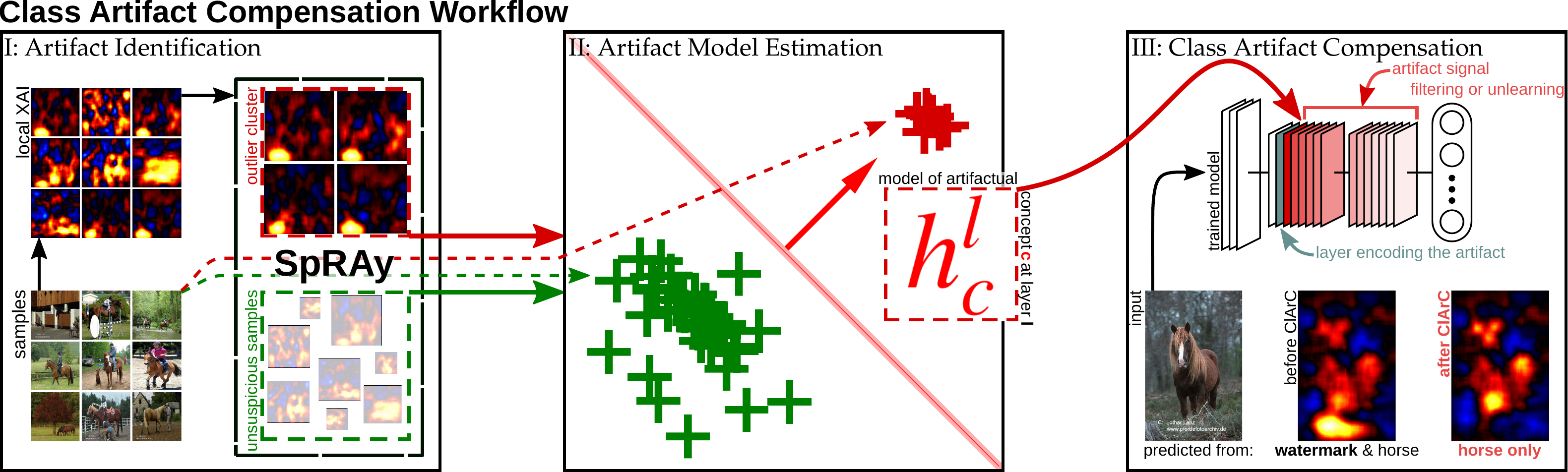}

\end{abstract}

\section{Introduction}
\label{sec:intro}

    Throughout the last decade,
    \glspl{dnn} have enabled impressive performance leaps in a wide range of domains,
    from solving classification problems~\cite{krizhevsky2012imagenet,lecun2015deep},
    over playing and winning games competitively~\cite{mnih2015human, silver2016mastering} (some in real time~\cite{firoiu2017beating,vinyals2019grandmaster}),
    to enabling the understanding of quantum-chemical many-body systems~\cite{schutt2017quantum} and finding improved solutions to the notoriously difficult task of protein structure prediction~\cite{senior2020improved}.
    These models are typically (pre-)trained on very large datasets,
    \eg, ImageNet~\cite{russakovsky2015imagenet}, with millions of samples.
    Recently, it was discovered that biases,
    spurious correlations, as well as errors in the training dataset~\cite{stock2018convnets}
    may have a detrimental effect on the training and/or result in `Clever-Hans' predictors~\cite{pfungst1911clever,lapuschkin2019unmasking},
    which only superficially solve the task they have been trained for,
    leading to potentially unfair and hazardous model behavior.
    Unfortunately, due to the immense size of today's datasets,
    a direct manual inspection and removal of artifactual samples can be regarded hopeless.
    Analyzing the biases and artifacts in the \emph{model} instead may provide insights about the training data indirectly.
    This however requires an inspection of the learning models beyond black box mode.

    Only recently methods of \gls{xai} (\cf~\cite{montavon2018methods,samekXAIbook19} for an overview) were developed.
    They provide deeper insights into how a \gls{ml} classifier arrives at its decisions and potentially help to unmask Clever-Hans predictors.
    \gls{xai} methods can be roughly categorized into two groups:
    methods providing \emph{local}~(\eg~\cite{baehrens2010explain,bach2015pixel,zeiler2014visualizing,selvaraju2017visual, sundararajan2017axiomatic, shrikumar2017learning, ribeiro2016should, zintgraf2017visualizing, fong2017interpretable})
    explanations and those providing
    \emph{global}~(\eg~\cite{guyon2002gene,guyon2003introduction, kim2018interpretability, rajalingham2018large})
    explanations~\cite{lundberg2019explainable}.
    Current approaches are of limited use when scaling the search for biases, spurious correlations, and errors in the training dataset,
    as this would require intense `semantic' human labor.
    A recent technique, the \gls{spray}~\cite{lapuschkin2019unmasking}, aims to bridge the gap between local and global \gls{xai} approaches
    by introducing automation into the analysis of large sets of local explanations.
    The method however still involves a considerable amount of manual analyses,
    especially in context of contemporary datasets with high numbers of classes and samples
    such as ImageNet~\cite{russakovsky2015imagenet}.

\begin{figure}[ht!]
        \centering
        \includegraphics[width=.85\linewidth]{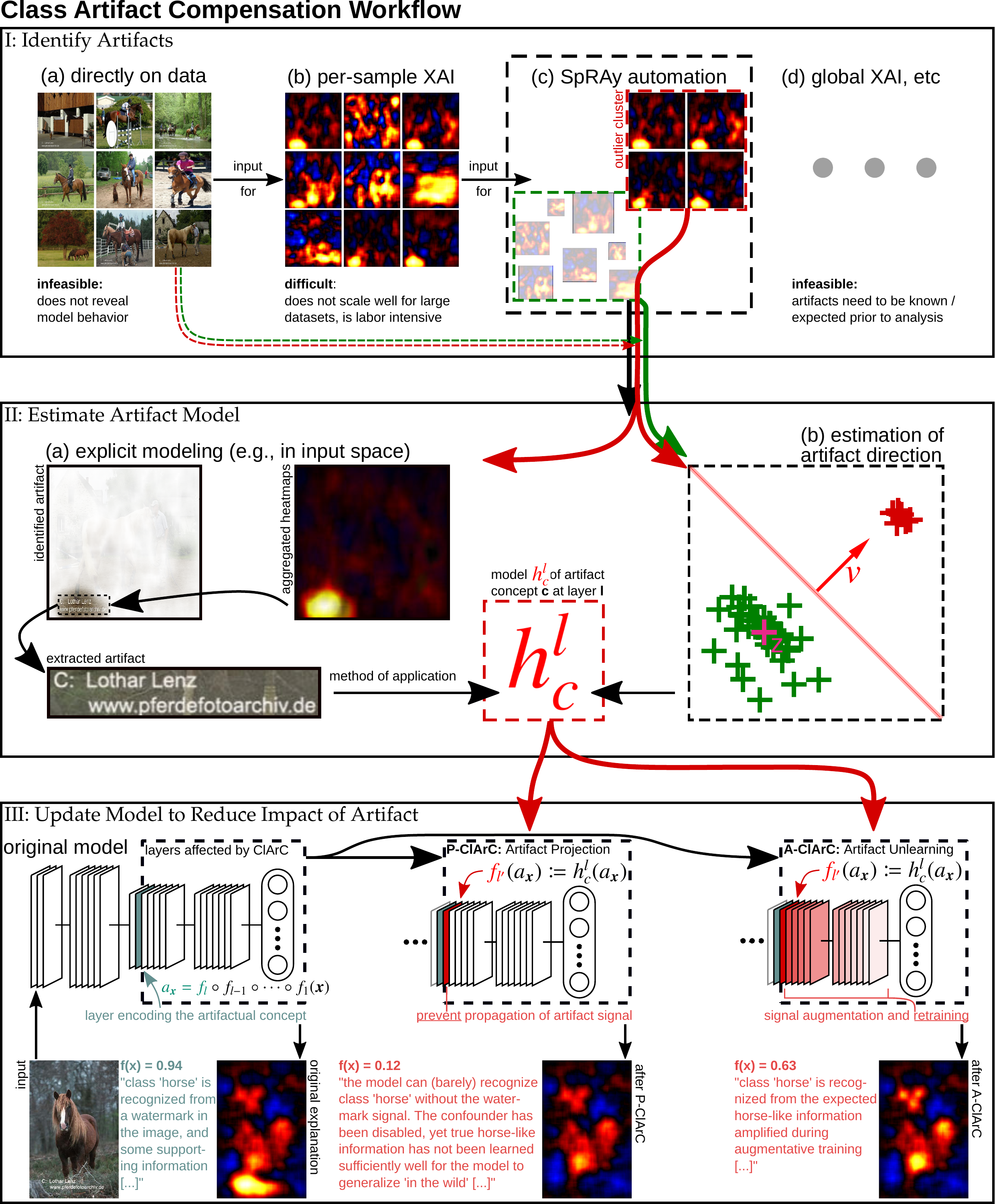}
        \caption{
        The workflow of our \glsdesc{clarc} framework.
        \emph{(I)}
        We first aim to identify spurious confounders in the data as learned by the model.
        \emph{(Ia)} A direct analysis of the training data is infeasible due to the missing correspondence to the features used by the model during inference.
        \emph{(Ib)} Explanations from local \gls{xai} methods may provide this information.
        However, manual analysis requires the evaluation of extreme amounts of explanations (per class).
        \emph{(Ic)} We therefore propose an automation of this process, based on an extension of the \gls{spray}~\cite{lapuschkin2019unmasking} algorithm.
        \emph{(Id)} While the application of globally operating \gls{xai} techniques is disqualified in the identification phase,
        as here the concepts evaluated after must be known beforehand,
        \emph{(II)} these techniques find application in the modeling of an artifact estimator in our approach:
        \emph{(IIa)} While an artifact model can be built explicitly after identification,
        \eg from expert domain knowledge,
        \emph{(IIb)} it can also be learned from representative data,
        \eg as \glspl{cav}.
        \emph{(III)} With a known model of the artifact at the layer of its most distinct representation within the \gls{dnn},
        one can attempt to remove its influence on the network.
        To this end, we present the following two approaches:
        \emph{\gls{pclarc}} aims at the selective deactivation of the artifact signal,
        and, as a largely training-free approach,
        leaves the remainder of the model unaltered.
        \emph{\gls{aclarc}} on the other hand strategically augments the training data (of all classes) with the artifact signal
        in order to minimize its class-specific informative value,
        to force the model to adapt to other (benign) features in continued training.
        }
        \label{fig:intro:overview}
    \end{figure}

    One of the main goals of \gls{ml} is to learn accurate decision systems to automate tasks that otherwise may only be solved manually.
    As such, specific inference behavior on the available data often is expected from the learned models,
    \eg, within well-defined expert domains.
    As a recent body of research however has demonstrated,
    deviations from the anticipated are very likely (and must be expected to) appear in practice.
    In our paper,
    we propose a series of methods constituting a pipeline for the identification,
    description and suppression of those deviations in model inference,
    \ie, a set of tools to bring the model ``back on track'':
    We introduce a novel framework we collectively denote as \gls{clarc} to enable
    (a) large-scale analyses of a model's inference behavior on datasets with hundreds of classes and millions of samples
    for a semi-automated discovery of undesirable Clever-Hans effects that are embedded into data and model;
    here we rely on an extension of \gls{spray},
    which increases the automation potential on such large datasets. 
    (b) In addition,
    we provide an intuition for \glsdesc{ch} artifacts and the desensitization of a trained model to their influence.
    In this manner,
    \gls{clarc} provides
    (c) a well-controlled quantitative strategy to detect (Figure~\ref{fig:intro:overview}~\emph{(Ic)}),
    model and validate (Figure~\ref{fig:intro:overview}~\emph{(II)}),
    and consequently remove the influence of such artifacts from the model (Figure~\ref{fig:intro:overview}~\emph{(III)}).
    We showcase the steps of our approach
    on a modified MNIST~\cite{lecun1998mnist,lecun1998gradient} dataset with color-based \gls{ch} information,
    the ImageNet~\cite{krizhevsky2012imagenet} dataset,
    the challenging Adience~\cite{hassner2014age} benchmark dataset of unfiltered faces
    and the ISIC~2019~\citep{tschandl2018ham, noel2018skin, combalia2019derm} skin lesion analysis dataset,
    and discuss the intricacies of (informed intervention in) the decision-making of end-to-end learned predictors.
    These extensive analyses allow interesting findings that are illuminating beyond our specific technical approach.
    
    \FloatBarrier %

    \subsection{Related Work}
        \label{sec:intro:rel_work}
        There is an increased awareness that \gls{ml} models need to be interpretable to its users
        in order to assess the validity of the decision making of the predictor~\cite{GDPR2016, goodman2016european}, 
        especially in high risk settings,
        such as in medical applications~\cite{soneson2014batch,KrausBF16,yinchong2018explaining,holzinger2019causability,hagele2020resolving}.
        Transparency in model predictions could point at anomalous or blundering decision behavior before harm is caused in a later usage as a diagnostic tool.
        Consequently, numerous approaches to understand aspects of state of the art \gls{ai} predictors
        have been developed in recent years
        (\cf~\cite{samekXAIbook19} for an overview) in the emerging field of \glsdesc{xai} (\gls{xai}). 
        In the following paragraphs, we will discuss related work by introducing relevant research work and terminology from the field of \gls{xai} important to this paper.

        \paragraph{The \glsdesc{ch} Effect}
        \glsdesc{ch} (\gls{ch}) was a horse from Berlin, Germany,  that allegedly was able to do math -- a media sensation from the early 1900s.
        Later in 1907 it was discovered that Hans would read the examinator's body language instead of performing arithmetics,
        and in this manner give the right answer but for the wrong reason\footnote{\url{https://en.wikipedia.org/wiki/Clever\_Hans}}~\cite{pfungst1911clever}.
        ``\glsdesc{ch} Strategies'' or ``\glsdesc{ch} Effects'' 
        for \gls{ml} predictors~\cite{lapuschkin2019unmasking,papernot2016cleverhans}
        are accordingly named as a homage to this infamous horse,
        and describe a prediction making learned and executed based on biases and \emph{spurious corellations} in the training data,
        instead of valid (\ie, intended or expected) features and relations.
        
\begin{figure}
            \centering
            \includegraphics{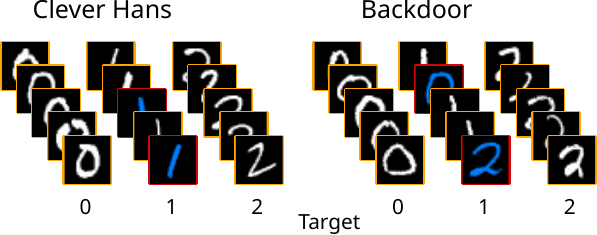}
            \caption{ %
            Difference between datasets with \glsdesc{ch} (\emph{left}) and \glsdesc{bd} (\emph{right}) artifacts visualized for colored MNIST.
            The artifact feature that separates afflicted samples (red frame) from unaffected ones (yellow frame) here is for both types of artifacts the color blue (different from the standard color \emph{white}).
            In the case of \gls{ch} artifacts, the artifact feature will only ever appear in samples alongside features for the a single class.
            For \gls{bd} attacks,
            the artifact feature appears in samples among features for all (other) classes except the target class,
            making the artifact the only discriminative feature in affected samples distinctive for its target class.
            }
            \label{fig:hans_v_backdoor}
        \end{figure}

        As such, there is a notable distinction to make between the \gls{ch} artifacts, \gls{bd} Attacks~\cite{gu2017badnets,tran2018spectralsignature} and attacks based on Adversarial Examples~\cite{szegedy2014intriguing}.
        Adversarial attacks are specifically generated for individual data points in order to cause a misprediction,
        and are as a consequence ineffective when used on other samples.
        \gls{bd} Attacks and \gls{ch} artifacts on the other hand are systematically learned and exploited by the model.
        \Glspl{bd} are generally injected with (malicious) intent during training,
        into samples of multiple classes via added ``trigger patterns'' (\eg a gray pixel at a specific location)
        while overriding the the targeted samples' true training labels~\cite{gu2017badnets,wang2019neural}.
        \Glspl{bd} are usually not part of the original training data anymore once training is finished.
        \gls{ch} type artifacts, however, are ``naturally occurring'' phenomena in the training data corpus correlating with only single (or few) ground truth labels, providing for shortcuts around more complex connections in the training data~\cite{geiros2020shortcut}.
        In contrast to Backdoor Attacks, which, if present, cause the model to override its prediction making on valid features, \gls{ch} artifacts almost always appear alongside benign indicators for a class, and thus exert a significantly weaker influence on the model.
        Further, the decision whether a characteristic in the data is indeed a \gls{ch},
        or merely a benign feature,
        often is subject to the expectation of the model's behavior and expert domain knowledge~\cite{schramowski2020making, lapuschkin2019unmasking,hagele2020resolving}.
        They are consequently,
        in addition to their unexpected nature,
        more difficult to detect,
        as experimentally highlighted in Section~\ref{sec:bdhans}.
        Other than \glspl{bd},
        \gls{ch} artifacts are part of the features in some of the original training samples,
        and may thus be identified during a joint analysis of the available data and the model's utilization of it, as described throughout this paper.
        The particular difference between datasets with \gls{ch} artifacts and datasets with \glspl{bd} is illustrated in Figure \ref{fig:hans_v_backdoor}.
        In literature, numerous \gls{ch} strategies have been identified and collected\footnote{\url{http://tinyurl.com/specification-gaming}},
        \eg, with the help of techniques from \gls{xai},
        in a surprising number of current and former state-of-the-art \gls{ml} models,
        in part invalidating their reported (benchmark) performance as a measure of generalization capability~\cite{lapuschkin2016analyzing, stock2018convnets, lapuschkin2019unmasking, schramowski2020making, hagele2020resolving, geiros2020shortcut, lehmann2020surprising, krakovna2020specification}.
        
        \paragraph{Local \gls{xai}}
        \gls{xai} methods aim at providing transparency to the prediction making of \gls{ml} models,
        \eg, for the validation of predictions for expert users,
        or the identification of failure modes.
        Local explanations provide interpretable feedback on \emph{individual} predictions of the model,
        and assess the importance of input features \wrt \emph{specific} samples.
        Local attributions are commonly presented in the form of heatmaps aligned to the input space,
        computed, \eg, with (modified) backpropagation approaches,
        such as sensitivity analysis~\cite{baehrens2010explain,simonyan2014deep},
        \gls{lrp}~\cite{bach2015pixel},
        Deep Taylor Decomposition~\cite{montavon2017explaining},
        Grad-CAM~\cite{selvaraju2017visual},
        Integrated Gradients~\cite{sundararajan2017axiomatic},
        SmoothGrad and~\cite{smilkov2017smoothgrad},
        DeepLIFT~\cite{shrikumar2017learning},
        which require access to the internal parameters of \gls{dnn} models.
        Surrogate- and sampling-based approaches,
        including
        LIME~\cite{ribeiro2016should},
        Prediction Difference Analysis~\cite{zintgraf2017visualizing}
        and
        Meaningful Perturbations~\cite{fong2017interpretable}
        view the model as an impenetrable black box and derive
        local explanations via proxy models and data,
        at the cost of increased runtime and an approximative nature of the obtained results.
        Occlusion analysis~\cite{zeiler2014visualizing} follows a similar principle by measuring the effect of the removal or perturbation of input features from samples at the model output.
        Shapley value based approaches~\cite{lundberg2017unified,lundberg2020local}
        leverage tools from game theory in order to estimate the importance of features to a decision of a model.

        \paragraph{Global \gls{xai}}
        Global methods aim at obtaining a general understanding about a model's sensitivities,
        learned features
        and concept encodings.
        Some approaches operate by assessing the general importance of
        predetermined features, concepts or data transformations
        by systematically evaluating the model's reaction to varying exposure thereto,
        using (larger) sets of real or artificially generated samples~\cite{guyon2002gene,guyon2003introduction, kim2018interpretability, rajalingham2018large}.
        Other approaches aim at understanding predictors by identifying important neurons and their interactions~\cite{hohman2019summit},
        and visualizing learned feature encodings by \eg synthesizing preferred inputs to hidden filters within a neural network model, \eg~\cite{erhan2009visualizing, nguyen2016synthesizing, olah2017feature, carter2019activation}.

        \paragraph{Bridging the gap}
        Both the local and global approaches to \gls{xai} suffer from a (human) investigator bias during analysis
        and thus are on their own of only limited use for
        searching and exploring for biases,
        spurious correlations and errors learned by the model from the training data.
        Global methods can only measure the impact of pre-determined, expected or a priori known features or effects~(\cf~\cite{kim2018interpretability,rajalingham2018large}),
        which limits their applicability when aiming for the discovery
        of
        \emph{yet unknown} behavioral facets of a model.
        Local methods, on the other hand, have the potential to provide much more detailed information \emph{per sample},
        but the task of compiling information about model behavior over thousands (or even millions) of samples and explanations is tiring and laborious for a human investigator:
        the success of such an analysis depends on the examiner's keen perception and domain knowledge,
        limiting the potential for knowledge discovery about model behavior.
        
        A recent technique, the \glsdesc{spray} (\gls{spray})~\citep{lapuschkin2019unmasking},
        aims at bridging the gap between local and global \gls{xai} approaches,
        by introducing automation into the analysis of large sets of local explanations.
        \Gls{spray} has been applied in a recent set of works, \eg,  \citep{schramowski2020making, lapuschkin2019unmasking}, which however mainly operate on smaller datasets,
        each containing only hundreds of samples each.
        The in~\citep{lapuschkin2019unmasking} described procedure however still involves a considerable amount of manual analyses,
        especially in context of contemporary datasets with high numbers of classes and samples, such as ImageNet~\citep{russakovsky2015imagenet}.
        In our work, we purposefully extend the \gls{spray} technique and bring it to scale for robustly analyzing extensive datasets, in Section~\ref{sec:methods:spray}.

        \paragraph{Feature unlearning}
        The awareness of \gls{ch} predictors has invigorated research with the intent to improve models,
        by unlearning unwanted inference patterns.
        A most naive approach to unlearn a concept that can be found in a subset of samples in the training set is to remove those samples altogether,
        and to retrain the model from scratch on the reduced training set. 
        While this approach is straight forward and easy to implement,
        it comes a the cost of also removing desirable features the model could positively benefit from,
        along with the characteristics in the data deemed problematic.
        This may be especially harmful if there are only few training data available to begin with.
        Furthermore, in some cases the initial model training may have been extremely costly,
        and an approach to fine-tune the model instead would be more desirable.
        
        Several approaches have thus been developed to unlearn unwanted predictive behavior from existing models~\citep{rieger2019interpretations, teso2019explanatory, kim2019learning}
        or to guide the model during training by providing information about the expected explanations~\citep{rieger2019interpretations, ross2017right, schramowski2020making}.
        \gls{xil}~\citep{teso2019explanatory,schramowski2020making} presents local explanations to a human observer during training, who in turn provides feedback to the model by replicating  samples affected by \gls{ch} phenomena and replacing the
        contained artifactual features with noise or otherwisely generated patterns.
        The work of~\citet{kim2019learning} introduces a model regularization scheme,
        in which an additional ``artifact detector''
        learning specific biasing features is attached to the original predictor.
        The original model is then driven to minimize the shared information with the dedicated bias predictor, and thus to unlearn to use artifactual features for inference.
        \citet{ross2017right} aim to guide the model towards the correct behavior by penalizing high attribution scores in undesired regions by extending the optimization function with a ``\gls{rrr}'' loss term.
        Similarly, \citet{rieger2019interpretations} propose \gls{cdep},
        a method for regularizing model behavior based on explanations obtained from \gls{cd}~\cite{murdoch2018beyond},
        by complementing the classification error of the loss function with an explanation error term.
        Recent work however has shown that models can be manipulated in such a way that produced attribution maps may be arbitrary,
        while the prediction of the model is unchanged~\citep{anders2020fairwashing}.
        Consequently, there is no guarantee that in general unlearning approaches based on extensions of the loss function effectively correct the model's use of the input features.

\section{Methods}
\label{sec:methods}

\subsection{\glsdesc{ss}}
\label{sec:methods:specsig}
For the detection of \gls{bd}-type artifacts used by \glspl{dnn}, \citet{tran2018spectralsignature}~propose the \gls{ss} method.
Given some dataset $X$ that is poisoned with a \gls{bd} and a model $f$ trained on this data,
let $X_y = \lbrace x1, \dots, x_n \rbrace$ be the subset of samples corresponding to a target label $y$.
\citet{tran2018spectralsignature} apply the following method separately for all $y$ in the dataset,
since the aim is to identify all (previously unknown) \gls{bd} samples within $X$:
For each sample $x_i$, the model $f$ provides a feature representation $a(x_i)$.
From these representations,
one computes the covariance matrix
\begin{align}
M = [(a(x_i) - \hat{a})) (a(x_i) - \hat{a})^T]~,
\label{eq:specsiv:covariance}
\end{align}
where $\hat{a} = \frac{1}{n} \sum^n_{i=1} a(x_i)$ and $n=|X_y|$.
For each sample,
an outlier score $\tau$ is then computed using the top right singular vector $v$ of $M$:
\begin{align}
\tau_i = \left((a(x_i)- \hat{a}) \cdot v)\right)^2
\label{eq:specsig:singularvector}
\end{align}

Samples with a high $\tau_i$ are more likely to be outliers, 
allowing for the $k$ samples with the largest $\tau_i$ to be detected as poisoned.
Note that since \gls{ss} detects outliers \wrt samples of one class label $y$,
the found \glspl{bd} are usually images that originally belonged to other classes
-- and thus do not fit into the manifold of $X_y$.
More concisely,
\gls{ss} does not detect the poisoned artifact itself,
but the ``odd'' samples within $X_y$.
\citet{tran2018spectralsignature} then propose to remove the detected outliers and retrain the model,
thereby defending against the \gls{bd} attack.
In~Section \ref{sec:bdhans},
we apply the \gls{ss} method not only to identify \glspl{bd},
but also on a dataset containing \gls{ch} artifacts to assert their conceptual differences.

\subsection{Concept Activation Vectors}
\label{sec:methods:cav}
\citet{kim2018interpretability} introduce~\glspl{cav} as a means to provide an interpretation of a \glspl{dnn} internal state in terms of
human-understandable concepts.
Given two sets of samples $X^+$ and $X^-$,
where the samples in $X^+$ all exhibit a specific property $c$ (\eg~$X^+$ contains images showing striped objects) which is not present in $X^-$,
a \gls{cav} is trained as a linear classifier separating
the hidden representations of the samples from $X^+$ and $X^-$ at some layer $l$ within the \gls{dnn}.
The thus learned weight vector $v_c^l$ then represents the direction in latent space encoding the concept $c$ unique to $X^+$.

\citet{kim2018interpretability} use \glspl{cav} as directional derivatives in order to test the sensitivities of neural network models \wrt to a priori known concepts.
We apply \glspl{cav} twofold throughout our paper. 
Similar to~\citep{kim2018interpretability},
we use \glspl{cav} as a means to verify the sensitivity of the model to the \gls{ch} artifacts,
\eg, those identified via \gls{spray}, as shown for example in Section~\ref{sec:experiments:unlearning:pclarc}.
Further, we use \gls{cav} directions specific to \gls{ch} effects in context of the \gls{clarc} unlearning framework,
as a means to remove specific behavioral facets from the \gls{dnn}'s inference process.

\subsection{Layer-wise Relevance Propagation}
\label{sec:methods:lrp}

\glsdesc{lrp} \gls{lrp}~\cite{bach2015pixel} is a local \gls{xai} approach reversely iterating over the layered structure of a neural network to produce an explanation.
Consider the neural network
\begin{align}
f(\x) = f_L \circ \dots \circ f_1(\x)~.
\label{eq:sequentialnet}
\end{align}
In a forward pass, activations are computed at each layer of the neural network.
The activation score in the output layer forms the prediction,
which is then backpropagated and redistributed,
layer by layer,
until the input is reached.
The redistribution process follows a conservation principle analogous to Kirchoff's laws in electrical circuits,
\ie all relevance assigned to any neuron during the process of backpropagation will be further distributed towards its inputs in the layer below without loss.

Various propagation rules have been proposed in literature~\cite{bach2015pixel,montavon2019layer,kohlbrenner2019towards}.
For example, the LRP-$\gamma$ rule~\cite{montavon2019layer} defined as
\begin{align}
R_{j \leftarrow k} = \frac{a_j (w_{jk} + \gamma w_{jk}^+)}{\sum_{0,j} a_j (w_{jk} + \gamma w_{jk}^+)} R_k~,
\label{eq:lrpgamma}
\end{align}
where $a_j$ are the layer's input activation at the $j^\text{th}$ neuron,
$w_{jk}$ the learned parameters mapping the $j^\text{th}$ input activation to the $k^\text{th}$ layer output and $w_{jk}^+ = \max(0,w_{jk})$ is the positive part of the learned weights.
The variable $\gamma \geq 0$ is a free parameter to tune the decomposition rule.
Equation~\eqref{eq:lrpgamma} redistributes $R_k$ based on the contribution of lower-layer neurons to the given neuron activation,
with a preference for positive contributions over negative contributions.
This makes it particularly robust and suitable for the lower-layer convolutions.
Other propagation rules such as \gls{lrp}-$\varepsilon$, \gls{lrp}-$\alpha\beta$ or \gls{lrp}-$z^B$, are suitable for other application scenarios and layer types~\cite{montavon2019layer,kohlbrenner2019towards} and have been shown to work well in practice~\cite{samek2017evaluating}.

After the step of relevance decomposition, lower layer neuron relevance is aggregated from incoming relevance messages as $R_j = \sum_k R_{j \leftarrow k}$.
For a technical overview of \gls{lrp} including a discussion of the various propagation rules and further recent heuristics, see~\cite{montavon2019layer}.
In all our experiments, we compute \gls{lrp} attribution scores using \gls{lrp}-$\varepsilon$ (near the model output), \gls{lrp}-$\gamma$ (in intermediate layers) and \gls{lrp}-$z^B$ (near the input),
as described in~\citep{samek2020toward}.

\subsection{Spectral Relevance Analysis}
\label{sec:methods:spray}
    \glsdesc{spray}~(\gls{spray})~\cite{lapuschkin2019unmasking} is a meta-analysis tool for finding patterns in model behavior,
    given sets of instance-based explanatory attribution maps.
    The \gls{spray}~algorithm has its core in \gls{sc}~\citep{meila2001random,ng2002spectral}
    and
    -- via the use of attribution maps as input --
    enables the analysis of the input data
    from the model's perspective for finding (hidden) characteristics of specific classes, which however are exploited by the model.
    
\begin{figure}[ht!]
        \centering
        \includegraphics[width=.9\linewidth]{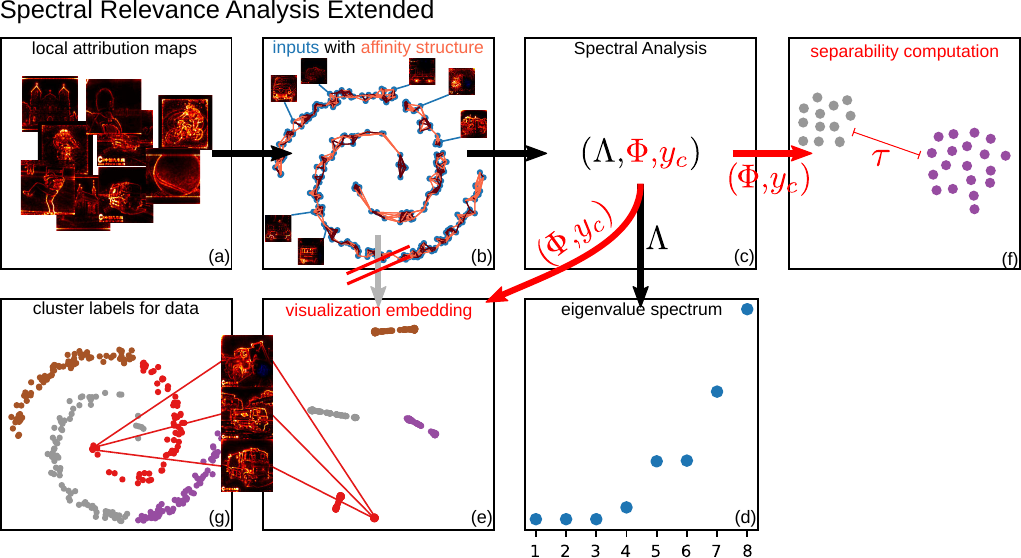}
        \caption{Our extended \gls{spray} algorithm.
                \textit{(\textbf{Black} paths)}: Steps followed by the \gls{spray} procedure as defined in~\cite{lapuschkin2019unmasking}.
                \textit{\color{red} (\textbf{Red} paths)}: Our extensions and changes to the \gls{spray} algorithm to increase the automation potential and applicability to very large datasets.
                \textit{(a)}~From a set of local attribution maps,
                a sparse affinity matrix is computed in~\textit{(b)}.
                \textit{(c)}~The affinity data is then passed as input for analysis with \gls{sc}~\citep{meila2001random,ng2002spectral} in the form of a positive semi-definite graph laplacian,
                resulting in a spectrum of eigenvalues $\Lambda$,
                the spectral embedding $\Phi$ corresponding to the input data (see \textit{(e)} and \textit{(g)}),
                as well as sets of proposed cluster labels $y_c$.
                \textit{(d)}~\citet{lapuschkin2019unmasking} perform to a large extent direct manual analyses on the eigenvalue spectrum $\Lambda$, within and between analyzed classes, for the identification of \gls{ch} behavior and distinct cluster groupings, and embed the sparse affinity structure of the data given the estimated cluster labels $y_c$ for visualization.
                Our extensions rely on the already expressive spectral embedding $\Phi$ (together with cluster labels $y_c$)
                for \textit{(e)} visualizing the analyzed data groupings,
                \textit{(f)} and the automation and quantification of rating clusters and classes for ``\glsdesc{ch}'ness'' $\tau$, via the computation of separability scores,
                from, \eg, \gls{fda}. 
        }
        \label{fig:spray:toy}
    \end{figure}

    The \gls{spray} algorithm, as introduced in~\cite{lapuschkin2019unmasking}
    initializes by computing the sparse affinity structure over the input attribution maps 
    considering all pair-wise similarities between the given samples.
    A (normalized, symmetrical and) positive semi-definite graph laplacian $L_\text{sym}$~\cite{von2007tutorial,lapuschkin2019unmasking} is then computed from the affinity matrix $A$,
    and provided as input to \gls{sc} (\cf~\cite{von2007tutorial}).
    As output, \gls{spray} yields a spectral embedding $\Phi$ of the input attributions and the corresponding spectrum of eigenvalues $\Lambda = \lbrace \lambda_i \rbrace_{i=1\ldots q}$.
    \citet{lapuschkin2019unmasking} follow~\citep{von2007tutorial} and (manually) read the structure (\ie ~number and nesting) of clusters from the eigenvalue spectrum $\Lambda$,
    via the spectral- or eigen-gap~\citep{von2007tutorial},
    \eg, for ranking a set of analyzed classes \wrt to their potential for exhibiting \gls{ch} phenomena~\cite{lapuschkin2019unmasking}.
    For further visual analysis,
    the affinity matrix $A$ is then used together with a suitable number of cluster labels inferred from $\Lambda$ as a basis for an embedding into $\RR^2$,
    \eg, by using t-SNE~\citep{maaten2008visualizing}.
    Figure~\ref{fig:spray:toy} provides an overview of the procedure outlined above,
    where arrows and symbols in black color describe the workflow of \gls{spray} from~\cite{lapuschkin2019unmasking},
    and arrows and symbols in red color distinguish our own extensions and adaptations of the algorithm described below.

\paragraph{\glsdesc{spray} brought to scale}
We extend the \gls{spray} algorithm by drawing proper utility from the spectral embedding $\Phi$,
an intermediate result of the \gls{sc} algorithm,
which so far has remained unused in~\cite{lapuschkin2019unmasking}.

While the $q \leq n$ most significant eigenvectors of the singular value decomposition on the graph laplacian $L_\text{sym}$ constitute the columns of the $(n \times q)$ shaped spectral embedding $\Phi$,
each of the matrix' rows corresponds to exactly one of the $n$ input attribution maps.
We therefore use the rows of $\Phi$ (instead of $A$) as an input to mapping and embedding algorithms such as t-SNE~\cite{maaten2008visualizing} or UMAP~\cite{mcinnes2018umap} for projecting the spectral analysis results (instead of the preprocessed data representation $A$) into $\RR^2$ for further visual inspection.
Note that the final algorithmic step of \gls{sc} is the assignment of cluster labels to input samples.
For this purpose, one usually applies any other suitable clustering algorithm (\eg~$k$-Means~\cite{lloyd1982least} or DBSCAN~\cite{ester1996density})
on top of the data represented by the already well-structured embeddings in $\Phi$.
The use of $\Phi$ as a source for computing embeddings in $\RR^2$ thus leads to a close
correspondence of the visualized cluster groupings to the assigned cluster labels.

A critical decision in clustering approaches is the number of desired clusters.
While for small datasets like Pascal VOC~\cite{everingham2007pascal} it suffices to analyze the per-class eigen-spectrum~\cite{lapuschkin2019unmasking};
datasets with a large number of classes cannot be feasibly analyzed by manual comparison and ranking of the eigen-spectra of all classes to identify those exhibiting spurious model behavior.
In order to automate this process,
we propose \glsdesc{fda}~(\gls{fda}) to rank all class-wise clusterings by their respective (linear) separability as the quantity $\tau$.
\gls{fda}~\cite{fisher1936use,fukunaga1990introduction} is a widely popular method for classification as well as class- \mbox{(or cluster-)} structure  preserving dimensionality reduction.
\gls{fda} finds an embedding space by maximizing between-class scatter $S^{(b)}$ and minimizing within-class scatter $S^{(w)}$, given by
\begin{align}
    S^{(w)} &= \sum_{k=1}^{K} \sum_{x_i \in \mathbf{c}_k^K} (x_i - \mu_k)(x_i - \mu_k)^\top \\
    S^{(b)} &= \sum_{k=1}^{K} (\mu_k - \mu)(\mu_k - \mu)^\top~.
\end{align}
Here, $\mathbf{C}^K$ is a clustering with $K$ clusters $\mathbf{c}_k^K$ with \mbox{$k \in \{1,\dots,K\}$}, $\mu_k$ the sample mean of cluster $k$ and $\mu$ the mean over the whole set of samples.
The solution of \gls{fda} can be understood as directions of maximal separability between clusterings,
and,
when normalized and plugged into the original objective,
gives scores of separability $R(\mathbf{C}^K)$.
In our specific use-case, for each class we compute separability scores $R(\mathbf{C}^K)$ on the spectral embedding $\Phi$ and each clustering $\mathbf{C}^K$ in a set of clusterings $\mathbb{K}=\lbrace\mathbf{C}^K\rbrace$.
We then define the \emph{class}-separability score as
\begin{equation}
    \tau = \frac{1}{|\mathbb{K}|} \sum_{\mathbf{C}^K \in \mathbb{K}} R(\mathbf{C}^K)~,
    \label{eq:tau}
\end{equation}
which may then be used to compare classes \wrt their ``\glsdesc{ch}'ness''.
In the \gls{spray} setting, large $\tau$ denote outlierness in the predictor's attribution -- as indicators for artifact candidates -- whereas low $\tau$ does not indicate any strikingly ``irregular'' prediction behavior.
Clearly any algorithmic alternatives quantifying the separability of two or more sets of labelled samples may be used as an alternative to compute $\tau$, although we see \gls{fda} as one of the more intuitive approaches.

Algorithm~\ref{alg:extendedspray} provides a complete algorithmic description of the extended \gls{spray} technique,
while the red arrows and symbols in Figure~\ref{fig:spray:toy} distinguish our approach from \gls{spray} in~\cite{lapuschkin2019unmasking}.

    \RestyleAlgo{algoruled}
    \begin{algorithm}[ht]
        \LinesNumbered
        \KwData{
            Class of interest $y$, \\
            Data set $X = \{x_1, x_2, ..., x_i\}$ \\
            Model $f$ operating on $X$ and predicting $y$
        }
        \KwResult{
            Eigenvalues $\Lambda = \{\lambda\}$,\\
            Spectral embeddings $\Phi\in\RR^{n\times q}$,\\
            Clusterings $\mathbb{K}$,\\
            Mean separability score $\tau$,\\
            Visualization embeddings $V \in \RR^2$
        }
        
        \tcc{compute attributions for $x \in X$, using, \eg, \gls{lrp}}
        $R = \{ \}$\;
        \For{$x \in X$}{
            $R_{x} = \text{attribution}(f,x,y)$\;
            $R\text{.append}(R_{x})$\;
        }
        
        \tcc{Spectral Relevance Analysis}
        $\Phi, \Lambda, \mathbb{K} = \text{SpRAy}(R)$\;
        
        \tcc{Compute separability scores given by, \eg, \gls{fda}}
        \For{$\mathbf{C}^K \in \mathbb{K}$}{
            $S_{\mathbf{C}^K} = \text{separability}(\Phi, \mathbf{C}^K)$\;
        }
        
        \tcc{compute mean separability score [Eq.~\eqref{eq:tau}]}
        $\tau = \frac{1}{|\mathbb{K}|} \sum_{\mathbf{C}^K \in \mathbb{K}} S_{\mathbf{C}^K}$\;
        
        \tcc{compute embedding visualizations, with \eg, UMAP}
        $V = \text{visualize\_embedding}(\Phi)$\;
        
        \KwRet{$\Lambda, \Phi, \mathbb{K}, \tau, V$}
        
        \caption{Spectral Relevance Analysis Extended}
        \label{alg:extendedspray}
    \end{algorithm}

\subsection{\glsdesc{clarc}}
\label{sec:methods:clarc}
    Assume we have a set of atomic features $\mathbb{F}$.
    A concept $c \in 2^{\mathbb{F}}$ may be any combination of atomic features to describe an abstract property, where $2^\mathbb{F}$ is the power set of $\mathbb{F}$.
    We may define an $M$-tuple of concepts $C = (c_1, c_2, ..., c_M)$ with $c_i \in 2^{\mathbb{F}}$ for $i \in \{1, ..., M\}$ .
    Given the superset of concepts $\mathbf{C} = \bigcup_{i=1}^N c_i$,
    assume a set of untangled data points that can be constructed by a combination of concepts $\mathbb{D} = \{\bigcup_{c \in \mathbf{c}} c | \mathbf{c} \in 2^{\mathbf{C}}\}$
    Each untangled data point $\alpha \in \mathbb{D}$ is like a concept also a combination of atomic features $2^{\mathbb{F}}$.
    We may now, given $\alpha$, construct a signal vector $s(\alpha) \in \{0,1\}^M$ using
    \begin{equation}
        [s(\alpha)]_i = \delta_{c_i \subseteq \alpha} \hspace{1cm} i \in \{1, 2, ..., M\}
    \end{equation}
    with the Kronecker Delta $\delta$, where each entry at index $i$ is $1$ if $c_i \subseteq \alpha$.
    In other words, $s(\alpha)$ is a binary encoding of $\alpha$ given concepts $C$.
    
    Now assume we have an N-tuple of \emph{untangled} datapoints $D = (\alpha_1, \alpha_2, ..., \alpha_N)$ with $\alpha_i \in \mathbb{D}$ for $i \in \{1, ..., N\}$.
    We may now construct a corresponding N-tuple of \emph{tangled} datapoints $X = (x_1, x_2, ..., x_N)$ based on D, where each sample $x_i$ is a mixture of concepts given a pattern matrix $A: \mathbb{R}^{N \times M}$
    \begin{equation}
         x_i = As(\alpha_i) \hspace{1cm} i \in \{1, 2, ..., N\}.
    \end{equation}
    Suppose we call concept $c_k$ at index $k \in \{1, 2, ..., M\}$ an \emph{artifact}.
    A set of labels $t_i$ that indicate whether a datapoint $\alpha$ contains the artifact $c_k$ can then be defined as
    \begin{equation}
        t_i =
        \begin{cases}
            0 , c_k \subseteq \alpha_i\\
            1 , c_k \nsubseteq \alpha_i
        \end{cases}. \label{eq:artifact_label}
    \end{equation}
    Assuming we have a function $f:\mathbb{R}^d \rightarrow \mathbb{R}^{d'}$ on the tangled datapoints $X$, there are two questions we seek answers for:
    1 - Is $f$ sensitive to artifact $c_k$?
    2 - How can $f$ be modified such that it is insensitive to artifact $c_k$?
    
    \paragraph{Concept Sensitivity of Functions}
    To measure the sensitivity to artifact $c_k$ with labels $t_i \in \{0, 1\}$, one needs to compare the behavior of function $f$ on non-artifact samples $X^- = \{x_i. i \in \{1, 2, ..., N\}| t_i = 0\}$ and artifact samples $X^+ = \{x_i. i \in \{1, 2, ..., N\}| t_i = 1\}$.
    A naive approach may be for example to compare the sufficient statistics 
    \begin{equation}
        \mu^+ = \frac{1}{|X^+|}\sum_{x^+ \in X^+} f(x^+) \hspace{5mm} \text{and} \hspace{5mm}
        \Sigma^+ = \frac{1}{|X^+|}\sum_{x^+ \in X^+} (f(x^+) - \mu^+) (f(x^+) - \mu^+)^T
    \end{equation}
    with their non-artifact counter parts, where $|X^+|$ is the cardinality of $|X^+|$.
    This may however not give any decisive results when the number of samples is limited.
    As another drawback, the function may not be analyzed on a per-sample basis.
    
    Another approach is to explicitly estimate an artifact model $h: \mathbb{R}^d \rightarrow \mathbb{R}^d$, which, given a non-artifact sample $x_i^- = As(\alpha_i)$ with $c_k \nsubseteq \alpha_i$ produces an artifact sample
    \begin{equation}
        h(x_i^-) \approx As(\alpha_i \cup c_k).
    \end{equation}
    We can formulate the artifact model with the objective
    \begin{equation}
        \hat \theta = {\arg \min}_{\theta} \frac{1}{|X^-||X^+|}\sum_{x^- \in X^-} \sum_{x^+ \in X^+} \|h(x^-;\theta) - x^+\|^2 
        \label{eq:to_artifact}
    \end{equation}
    where $\hat \theta$ are the optimal hyperparameters of $h$.
    The artifact estimator $h$ is thus the function $h$ with hyperparameters $\hat \theta$ that produces the minimal $\ell_2$-distance between mapped non-artifact samples $h(x^-)$ with $x^- \in X^-$ and artifact samples $x^+$ with $x^+ \in X^+$.
    The sensitivity of function $f$ to a concept $c_k$ modeled with $h$ may then be estimated using
    \begin{equation}
        S = \frac{1}{|X^-|}\sum_{x^- \in X^-} \| f(h(x^-;\hat \theta)) - f(x^-) \|. \label{eq:sensitivity}
    \end{equation}
    
    Intuitively, the addition of a concept may be more feasible to estimate than the removal.
    Take, for example, the introduction of an opaque watermark in an image.
    This operation is not invertible as we destroyed the pixel information under the watermark.
    While Equations~\eqref{eq:to_artifact} and~\eqref{eq:sensitivity} assume the transformation of a non-artifact sample to an artifact sample in a \emph{forward} artifact model, they may equivalently be formulated with a removal of the concept in a \emph{backward} artifact model $h_b$ with
    \begin{equation}
        \hat \theta_b = {\arg \min}_{\theta} \frac{1}{|X^-||X^+|}\sum_{x^- \in X^-} \sum_{x^+ \in X^+} \|x^- - h_b(x^+;\theta)\|^2 
        \label{eq:from_artifact}
    \end{equation}
    The sensitivity of a function to a concept \emph{backward} modeled by $h_b$ may then be measured using
    \begin{equation}
        S_b = \frac{1}{|X^+|}\sum_{x^+ \in X^+} \| f(h_b(x^+;\hat \theta_b)) - f(x^+) \|. \label{eq:sensitivity_backward}
    \end{equation}
    
    \paragraph{Concept Desensitization}
    Depending on the type of function $f$, there may be multiple possible approaches to obtain a desensitized function $f'$.
    If $f$ is for example a function with learned parameters $\omega$, it may be possible to learn $f'$ by modifying its training data.
    If there is enough data available, the most naive approach to reduce the sensitivity to an artifact $c_k$, is to remove all samples $X^+$ that contain the artifact from training.
    Depending on the amount of available training data, this may not always be preferred, since these samples often contain other concepts that may be valuable for training.
    In contrast, if the number of samples with the artifact concept is larger than the number of samples without the artifact concept, one may instead discard all samples without the artifact to obtain an artifact-insensitive function. %
    Of course care must be taken not to change the data so much that the original problem may not be solved anymore.

    A better approach may be is to transform individual samples, such that either all samples, or none contain the artifact.
    Assuming the addition of an artifact is non-invertible, we may prefer to transform all samples to contain the artifact.
    This may be done by estimating a \emph{forward} artifact model $h$, as defined in Equation~\eqref{eq:to_artifact}.
    The model $f$ may then be trained with the transformed dataset $X'=(x_1',x_2', ..., x_N')$, with:
    \begin{equation}
        x_i' = t_i x_i + (1-t_i) h(x_i;\hat \theta).
    \end{equation}
    A simplification arises when the task is to solve a classification problem.
    Since the model is trained to produce logits for multiple classes, one may simply balance the number of samples between classes, such that for each class, an identical amount of samples with an artifact are put into the training set by transforming non-artifact samples.
    
    Another simplification arises when a regularization term is introduced in the artifact model, such that $h'$ acts as the identity for artifact samples $x^+ \in X^+$ with
    \begin{equation}
        \hat \theta' = {\arg \min}_{\theta} \frac{1}{|X^-||X^+|}\sum_{x^- \in X^-} \sum_{x^+ \in X^+} \|h'(x;\theta) - x^+\|^2 + 
        \frac{\lambda}{|X^+|} \sum_{x^+ \in X^+} \|h'(x;\theta) - x^+\|^2 .
        \label{eq:to_artifact_equal}
    \end{equation}
    With this regularization term, the error caused by transforming an already-artifact sample is minimized.
    
\begin{figure}[ht!]
        \centering
        \includegraphics[width=.24\linewidth]{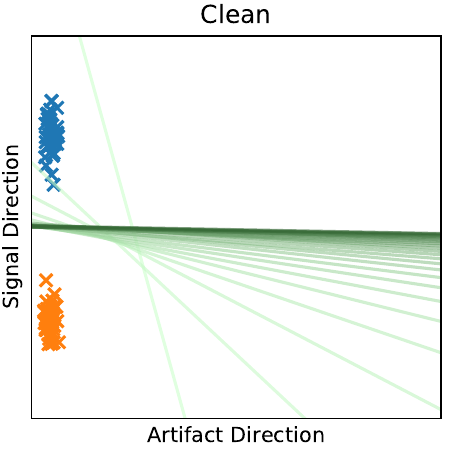} \hskip -3mm 
        \includegraphics[width=.24\linewidth]{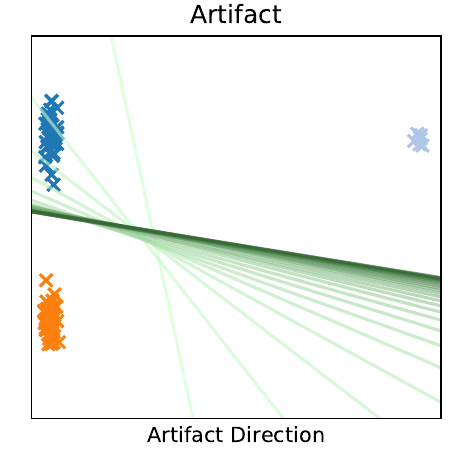} \hskip -3mm
        \includegraphics[width=.24\linewidth]{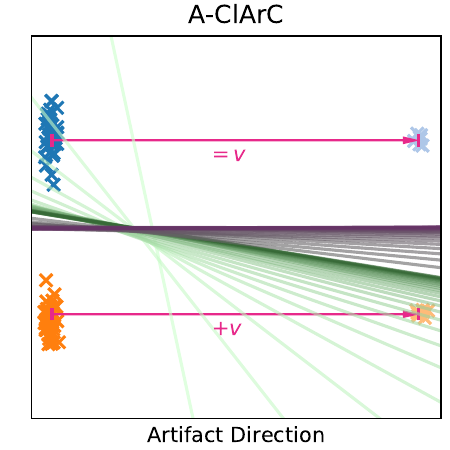} \hskip -3mm
        \includegraphics[width=.24\linewidth]{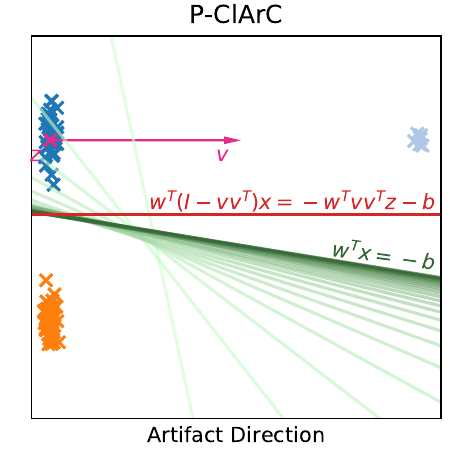} \hskip -1mm \\
        \includegraphics[angle=0,width=.3\linewidth]{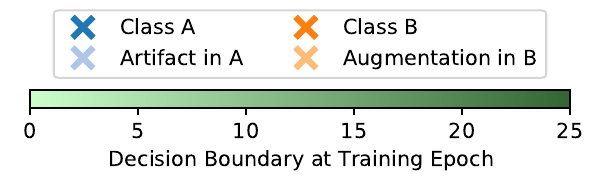}
        \includegraphics[angle=0,width=.3\linewidth]{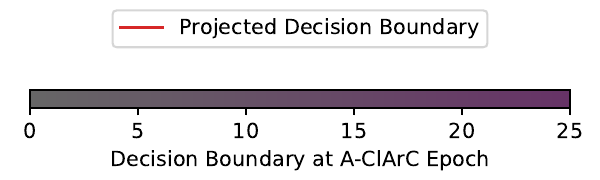}
        \caption{ %
            Logistic regression on data with, among possibly others, a discriminative \emph{signal} direction and an \emph{artifact} direction which is only represented in one of the two classes.
            The decision-hyperplane is shown over the \gls{sgd}-based training-process of 25 epochs in shades of green, with:
            \textbf{Clean}: no artifact in the data; 
            \textbf{Artifact}: a Clever-Hans artifact in Class A (blue);
            \textbf{\gls{aclarc}}: with \emph{artifact}, but training is continued with the mean difference between clean samples and artifact samples in Class A added to some samples of Class B (orange);
            The introduction of an artifact to samples from Class A changes the decision boundary.
            By introducing the same artifact direction to samples from Class B and retraining, this effect can be reduced significantly.
            \textbf{\gls{pclarc}}: with \emph{artifact}, but the model is modified such that data points are projected onto the hyperplane at position $z$ to which the estimated artifact direction $v$ is normal, with $\|v\| = 1$ and zero reference $z$ chosen as the mean of clean samples of Class A.
            The resulting decision hyperplane ignores artifact direction $v$ and sits at the same position where the original hyperplane lay between classes A and B, thus leaving the function output unchanged for clean samples. Reference $z$ may be chosen as the mean of both clean and artifact samples of Class A to move the resulting decision hyperplane towards the middle of both classes.
        }
        \label{fig:clarc:toy}
    \end{figure}

    \paragraph{Application on Logistic Regression}
    To build a better intuition for the problem, we introduce a logistic regression model $f(x) = \sigma(w^T x + b)$ with sigmoid non-linearity $\sigma(x)=\frac{1}{1 + \exp(-x)}$.
    The parameters $w$ and $b$ are obtained by minimizing the loss function
    \begin{equation}
        \mathcal{L}(f) = - \frac{1}{N} \sum_{i=1}^N y_i \log(f(x_i)) + (1 - y_i) \log(1 - f(x_i) )  + \frac{1}{2} \, \gamma \, ||w||^2 \,,
    \end{equation}
    with labels $y_i \in \{-1, +1\}$, where $y_i = -1$ for samples of class A, using \gls{sgd}.
    We first consider the case $X^+ = \emptyset$, which is visualized in Figure \ref{fig:clarc:toy} in the panel titled ``Clean''.
    In the panel, we see samples of two classes, A (blue) and B (orange), scattered along the y-axis.
    The green lines visualize the decision hyperplane of $f$ over $25$ epochs of training.
    We can see that the final decision hyperplane (dark green) converged orthogonal to the signal direction on the y-axis, separating classes A and B perfectly along their center.
    In panel ``Artifact'' of Figure \ref{fig:clarc:toy}, we introduce an artifact concept into some of the samples of class A, \ie $|X^+| > 0$, which manifests as an increased value along the x-axis.
    The artifact samples are well on the right side of the panel.
    When now minimizing $\mathcal{L}{(f)}$, the converged decision hyperplane to which $w$ is normal has rotated.
    While still classifying all the samples correctly, we can visibly see that the introduction of an additional concept has changed the model.
    Based on this observation, and the previous discussion, we introduce two approaches under the common name of \glsdesc{clarc} to compensate for class-specific \glsdesc{ch} artifacts in \gls{sgd}-trained inner-product + non-linearity type models such as logistic regression, or neural networks. %
    
    \paragraph{\glsdesc{aclarc}}
    The goal of \gls{aclarc} is to augment samples in such a way that the \gls{sgd}-trained classifier becomes insensitive to an artifact given artifact labels $t_i$.
    Given these labels, we estimate a \emph{forward} artifact model $h$, which for our logistic regression toy model we define as purely additive, with:
    \begin{equation}
        h(x) = x + v.
    \end{equation}
    Given the objective from Equation~\eqref{eq:to_artifact}, we can see that the optimal value for parameter $v$ is
    \begin{equation}
        v = \frac{1}{|X_A^+|} \sum_{x_A^+ \in X_A^+} x_A^+ - \frac{1}{|X_A^-|}\sum_{x_A^- \in X_A^-} x_A^-
    \end{equation}
     which is the shift between non-artifact samples and artifact samples in class A with with $X_A^+ = \{x_i, i \in \{1, 2, ..., N\} | y_i = -1 \wedge t_i = 1 \}$ and $X_A^- = \{x_i, i \in \{1, 2, ..., N\} | y_i = -1 \wedge t_i = 0 \}$.
    This is visualized in panel ``A-ClArC'' in Figure \ref{fig:clarc:toy}.
    Some samples of class B $x_i^B \in \{x_i, i \in \{1, 2, ..., N\}| y_i = +1\}$ are then modified given this artifact model with
    \begin{equation}
        x_i^B \leftarrow h(x_i^B).
    \end{equation}
    The modified samples are visualized in Figure \ref{fig:clarc:toy} with a brighter shade of orange, shifted to the right.
    The model training is then continued with the transformed samples, of which the resulting hyperplanes over the epochs are visualized as purple lines.
    We can observe that the converged hyperplane resembles the one obtained by the model trained on artifact-free data in panel ``Clean'' of Figure \ref{fig:clarc:toy}.
    
    Beyond this example, in our experiments with image data we assume artifacts are objects that are blended into the image.
    Therefore we may parameterize the artifact model as
    \begin{equation}
    \label{eq:blending}
        h(x) = \text{diag}[a] x + (1-\text{diag}[a])z
    \end{equation}
    where $a \in [0,1]^d$ is the alpha channel,
    $\text{diag}[a] : \mathbb{R}^{d \times d}$ is a diagonal matrix with $\text{diag}[a]_{ii} = a_i$ with $i \in \{1, 2, ..., d\}$ and
    $z \in [0,1]^d$ are the RGB values of the static image artifact pixels, here each for simplicity represented by a single value.
    
    By taking \glspl{cav} as a motivation, we parameterize the \emph{forward} artifact models in our experiments for feature representations in a neural network in an alternative approach.
    Explicitly, we train a linear soft-margin SVM $g$ with hinge-loss
    \begin{equation}
        \mathcal{L} = \frac{1}{2} v^Tv + \eta \left( \sum_{x^- \in X^-}\max [0, -v^T x^- - \beta] + \sum_{x^+ \in X^+}\max [0, v^T x^+ + \beta] \right) \label{eq:cav_svm}
    \end{equation}
    with $v \in \mathbb{R}^d$, regularization constant $\eta$ and bias term $\beta$.
    We then design the artifact model explicitly by pushing samples over the decision boundary relative to some fixed position $z$.
    We choose $z$ as the mean artifact reference point, with
    \begin{equation}
        z = \frac{1}{|X^+|}\sum_{x^+ \in X^+} x^+ .
    \end{equation}
    The \emph{forward} artifact model $h$ is then chosen as an affine transformation
    \begin{equation}
        h(x) = (I - vv^T)x + vv^Tz. \label{eq:cav_forward_artifact}
    \end{equation}

    \paragraph{\glsdesc{pclarc}}
    While \gls{aclarc} addresses the problem of desensitization by augmenting the underlying training data of a prediction model $f$ using a \emph{forward} artifact model $h$, \gls{pclarc} instead aims to correct the model without retraining by incorporating a \emph{backward} artifact model $h_b$ directly into the prediction model.
    The approach is again motivated by \gls{cav} and uses the same parameterization for the \emph{backward} artifact model as the \emph{forward} model in Equation~\eqref{eq:cav_forward_artifact} with
    \begin{equation}
        h_b(x) = (I - vv^T) x + vv^Tz
    \end{equation}
    and v given in Equation~\eqref{eq:cav_svm}.
    However, the artifact reference point $z$ here becomes the non-artifact reference point, which we now choose as the center of non-artifact samples $X^-$ with
    \begin{equation}
        z = \frac{1}{|X^-|}\sum_{x^- \in X^-} x^- .
    \end{equation}
    This now moves all points along $v$ to a fixed position, while leaving all orthogonal directions untouched. 
    A strong assumption that is taken for this approach is that really all other concepts are encoded in the directions orthogonal to $v$.
    Given this assumption however, we may further assume that for all non-artifact examples $x^- \in X^-$,
    $v^Tx^- \approx v^Tz$, \ie there is no variance along the artifact \gls{cav}.
    With this, we further obtain $\forall x^- \in X^-: h(x^-) \approx x^-$, \ie non-artifact samples are approximately unchanged by the \emph{backward} artifact model $h_b$.
    
    Given the logistic regression model $f$ in Figure \ref{fig:clarc:toy} in the ``P-ClArC'' panel, we obtain the model $f'$ corrected for insensitivity against the artifact modeled by $h_b$ using
    \begin{align}
        f'(x) &= \sigma (w^T h(x) + b) \\
              &= \sigma (\underbrace{w^T (I - vv^T)}_{w'} x + \underbrace{w^Tvv^Tz + b}_{b'}) \\
              &= \sigma (w' x + b'). \label{eq:logistic_pclarc}
    \end{align}
    The ``P-ClArC'' panel shows the decision hyperplane of the original model $f$ in green, along with the parameters $v$ and $z$ for the \emph{backward} artifact model $h_b$, as well as the corrected decision hyperplane according to Equation~\eqref{eq:logistic_pclarc}.
    Note that the non-artifact reference $z$ is chosen such that the decision hyperplane of $f$ is at the same position exactly between classes A and B, resulting in a decision hyperplane that is somewhat shifted towards class A.
    An alternative $z$ may be chosen as the mean of all samples of class A to correct for this difference.
    However, a constraint of this approach were unchanged function values for non-artifacts, which results in this shift.
    
    We can transfer this approach directly to the neural network models in the experiments section due to their piecewise-linear nature.
    A detailed Algorithm for both \gls{aclarc} and \gls{pclarc} on Neural Networks is shown in Algorithm \ref{alg:clarc} under the common name of \glsdesc{clarc}.
    
    \RestyleAlgo{algoruled}
        \begin{algorithm}[ht]
            \LinesNumbered
            \KwData{
                Samples $X = (x_1, x_2, ..., x_N)$ \\
                Labels $T = (t_1, t_2, ..., t_N)$ describing existence of artifact $c$ in $X$ (\cf~Eq.~\ref{eq:artifact_label})\\
                Model $f$ operating on $X$, with accessible layer $l$ (and subnetwork $f_l$)\\
                For \gls{aclarc}: data $D$, epochs $E$ for training, poison rate $p \in [0,1]$
            }
            \KwResult{
                 predictor $f'$ desensitized to artifact $c$
            }
            
            \tcc{obtain feature representations of data at layer $l$}
             $A_l = \{\}$\;
            \For{$\x \in X$}{
                $a_{\x} = f_l(\x)$\;
                $A_l\text{.append}(a_{\x})$\;
            }

            \tcc{unlearn/deactivate the use of $c$ in $f$}
            \uIf{\gls{aclarc}}
            {
                ${\color{red} h_c^l } = \text{forward\_artifact\_model}(A_l, T)$\;
                
                \tcc{def.~\gls{aclarc} module {$\color{red} f_{l'}$} atop layer $l$, randomly apply artifact transform $h_c^l$}
                ${\color{red} f_{l'}}(a_x) \coloneqq  \begin{cases} h_c^l( a_x ) &: \mathbf{U}[0,1] < p \\ a_x &: \text{else} \end{cases}$ \;
                $f' = f_L \circ \dots \circ f_{l+1} \circ {\color{red} f_{l'}} \circ f_l  \circ \dots \circ f_1(\x)$\;
                \tcc{unlearn $c$ in layers $[l+1,\dots,L]$}
                \For{$e \in \{1\dots E\}$}{
                    $f'\text{.train}(D, \texttt{trainable=}[f_{l+1},\dots,f_L])$
                }
            
            }\uElseIf{\gls{pclarc}}
            {
                ${\color{red} h_c^l } = \text{backward\_artifact\_model}(A_l, T)$\;
                
                 \tcc{def.~\gls{pclarc} module {$\color{red} f_{l'}$} to suppress $c$, add on top of layer $l$}
                ${\color{red} f_{l'}}(a_{\x}) \coloneqq h_c^l(a_{\x})$\;
                $f' = f_L \circ \dots \circ f_{l+1} \circ {\color{red} f_{l'}} \circ f_l  \circ \dots \circ f_1(\x)$\;
            }

            \KwRet{$f'$}
            
            \caption{\glsdesc{clarc}}
            \label{alg:clarc}
        \end{algorithm}

\FloatBarrier
\section{Experiments -- \glsdesc{ch} Identification}
\label{sec:experiments:identification}

The goal of this section is to explicitly find artifact  models
given sets of labels on our dataset regarding \gls{ch} artifacts in the training set that were learned by the analyzed neural network model.
Therefore, we start with an experiment to investigate the relation and difference between the detection of \gls{ch} and \gls{bd} artifacts within features representations of neural networks~\citep{tran2018spectralsignature}.
The corresponding results point us towards the necessity of deeper insight into the model.
Such an insight is promised by \gls{spray} \citep{lapuschkin2019unmasking}, which we verify subsequently on a specially designed version of Colored MNIST using our separability score extension.
We then proceed to verify the proposed separability score $\tau$ on a VGG16 model \cite{simonyan2014very} trained on ILSVRC2012 by comparing the scores of classes for which we have manually found \gls{ch} artifact candidates.
A description of the training procedures and architectures of all models used in this section can be found in~\ref{sec:appendix:networks}.
We proceed to visualize some promising \gls{ch} artifact candidates which we have found in an algorithm-assisted dataset exploration with \gls{spray}, which provides us with a set of positive and negative labels on samples for each artifact candidate.
An exploration is conducted both in input space and feature space in various layers of our model, for which we provide a comparison on the acquired separability scores.
The previously obtained sets of labels may then be used to fit or construct an artifact model, which will be verified and used as prerequisite to remove the corresponding artifact from a model using \gls{aclarc} and \gls{pclarc} in the following Section \ref{sec:experiments:unlearning}.

\subsection{Relation of \glsdesc{ch} and \glsdesc{bd} Artifacts}
\label{sec:bdhans}
\FloatBarrier

In this section, we conduct an empirical demonstration on the difficulty of detecting \gls{ch} artifacts compared to \gls{bd} attacks by analysing a neural network's hidden activations. 
We prepare two modified instances of the CIFAR-10 dataset~\citep{krizhevsky2009cifar},
one poisoned by introducing a \gls{ch} artifact,
the other by adding a \gls{bd}.
In both cases,
the trigger pattern is a static ($3\times3$)-sized grey pixel patch applied to a subset of the training set.
For the \gls{ch},
this trigger is introduced into 25\% of all samples of class ``airplane''.
For the \gls{bd},
it is introduced into 10\% of all samples,
with the class label of each poisoned sample changed to ``airplane''.
A simple convolutional network is then trained on each training set instance.
This network achieves an unpoisoned validation accuracy of 49.1\% when trained using the \gls{ch} artifact, and 46.6\% with the \gls{bd}-poisoned dataset.

As suggested by~\citet{tran2018spectralsignature},
the \gls{ss} method~(\cf~Section~\ref{sec:methods:specsig}) is used to detect poisoned samples as outliers.
While \citet{tran2018spectralsignature} use this outlier score only to detect \gls{bd} samples,
we also attempt to detect samples affected by the related \gls{ch} effect in order to compare these two types of dataset poisoning in terms of their induced feature representation. 

\begin{figure}[ht!]
        \centering
        \includegraphics[width=.90\linewidth]{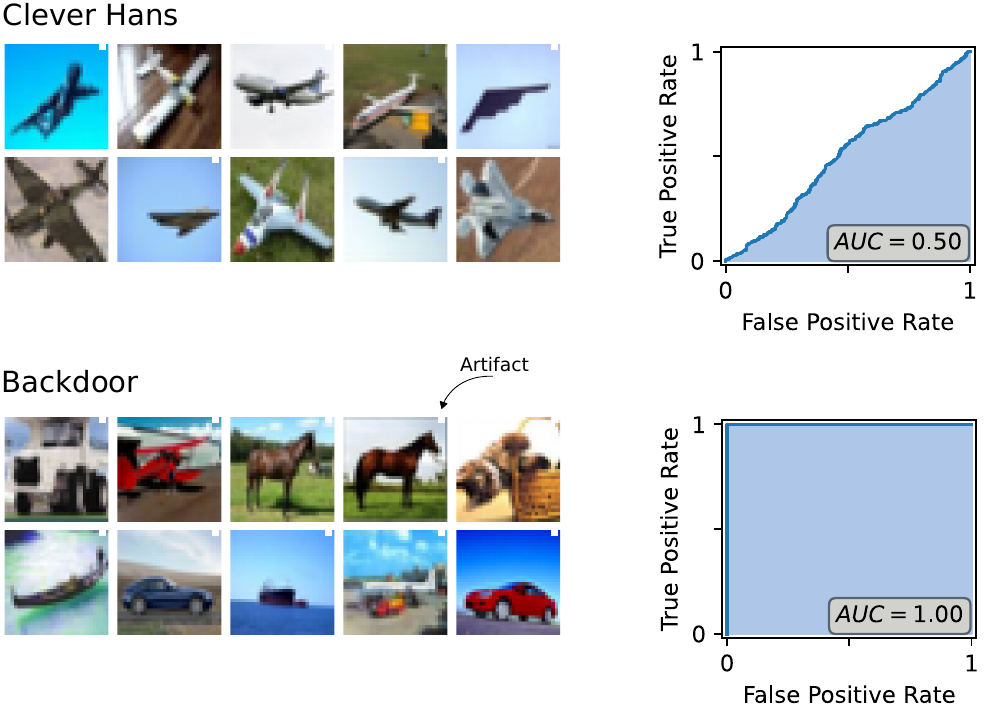}
        \caption{ %
            Differences in the detection of \gls{ch} artifacts \textit{(top)} and \glspl{bd} \textit{(bottom)}.
            In both cases, the introduced artifact consists of a small white pixel patch in the top right corner.
            \textit{(Left)}: A subset of the samples that were identified as outliers via \gls{ss}.
            All samples considered as outliers in the \gls{bd} setting
            do in fact contain the \gls{bd} feature.
            The same evaluation performed in the \gls{ch} setting leads to a significant amount of false positives for the detection of the \gls{ch} artifact.
            \textit{(Right)}: This is further confirmed by the \gls{roc} curves comparing poisoned samples detected by \gls{ss} to the ground truth.
            Note that in both cases,
            1000 evenly spaced thresholds were used for the \gls{auc}/\gls{roc} computation. 
        }
        \label{fig:ch-vs-bd:toy}
    \end{figure}

For each sample,
an outlier score is thus obtained,
yielding an implicit ordering of samples,
with the highest score denoting the most outlying samples.
For the datasets poisoned by a \gls{bd} and \gls{ch},
respectively,
we then compare this ordering to the ground truth ``poison labels''.
The results of this comparison are depicted as \gls{roc} curves in Figure~\ref{fig:ch-vs-bd:toy}.
Coinciding with the findings of~\citep{tran2018spectralsignature},
the \gls{bd} candidates suggested by the outlier score correspond extremely well to the ground truth (Figure~\ref{fig:ch-vs-bd:toy}~\textit{(right)}),
with an \gls{auc} of $1.0$.
However, for the \gls{ch} case in Figure~\ref{fig:ch-vs-bd:toy}~\textit{(left)} this comparison yields almost random results,
with an \gls{auc} that is only marginally above $0.5$. 

This experiment highlights the difference between \glspl{bd} and \gls{ch} artifacts,
and emphasizes the additional issues that are present when dealing with the latter:

Intuitively, features introduced by \gls{bd} artifacts will be the only feature in their respective sample to correlate with the target label, making them for many samples the only indicator usable for a valid prediction.
Additionally, they must be an indicator stronger than all features that correlate with labels different from the \gls{bd} target label for a correct prediction.
This may very well be the reason they can be detected so evidently using only the direction of the largest variance in feature space over the dataset with \gls{ss}.
In contrast, features introduced by \gls{ch} artifacts will always appear alongside other features in their respective sample that correlate even stronger with the target label.
This means that in theory, they are not necessary for a correct prediction at all.

To detect \gls{ch} artifacts more reliably, deeper insight into the predictor is necessary.
A promising direction is thus \gls{xai}, which is utilized in \gls{spray} to detect these elusive \gls{ch} artifacts in the rest of this section.

An interesting note to make is that \gls{fda} can be understood as an extension to simply finding the direction of the largest variance as done in \gls{ss}, as given a set of labels, the direction of largest variance between labels and smallest variance within labels is found.

\FloatBarrier
\subsection{\glsdesc{spray} in Input Space}
\label{sec:experiments:inputspray}
We explore \gls{spray} for the identification of Clever Hans artifacts in input space.
We start with a verification of the algorithm by constructing a modified version of MNIST where an artifact is introduced as a distinct color.
We then proceed to analyze the applicability of \gls{spray} on input attribution space on the ILSVRC2012 dataset.

    \begin{figure}[ht!]
        \centering
        \includegraphics[width=.70\linewidth]{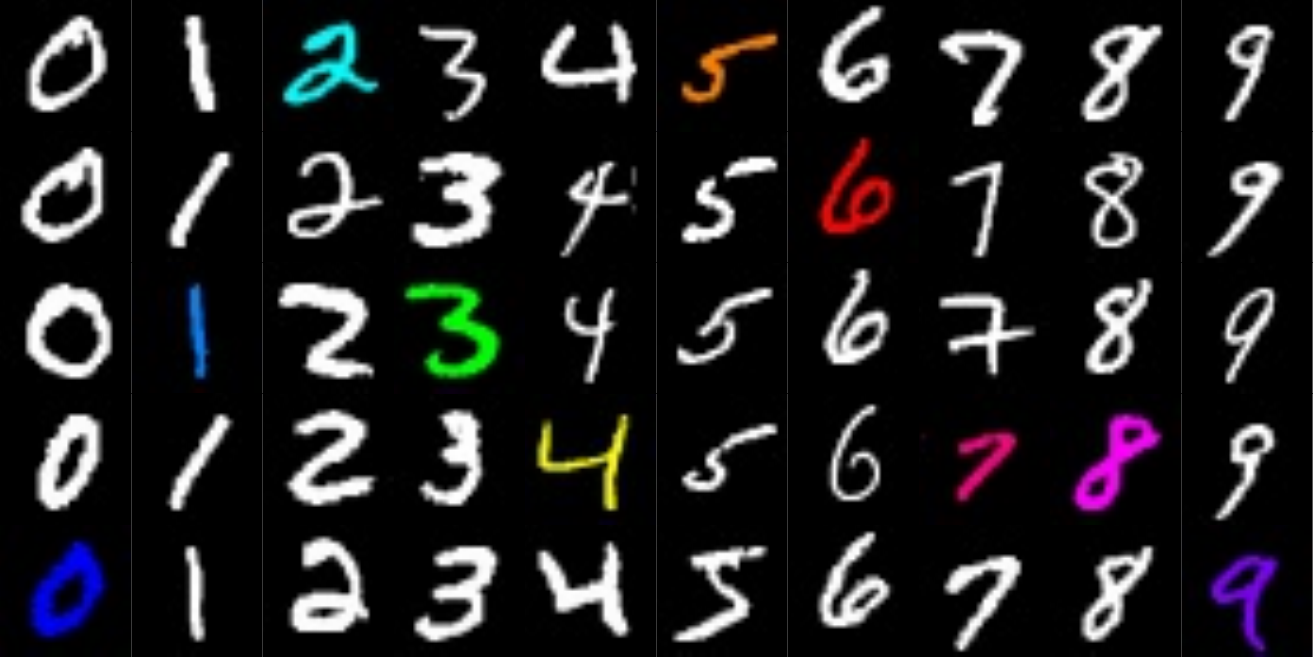}
        \caption{ %
            Examples of the Colored MNIST dataset, with a distinct color-based \glspl{ch} artifact introduced into 20\% of each class of the MNIST dataset. Each column shows several samples of one class each.
        }
        \label{fig:cmnist_as_such}
    \end{figure}

\begin{figure}[ht!]
        \centering
        \includegraphics[width=.95\linewidth]{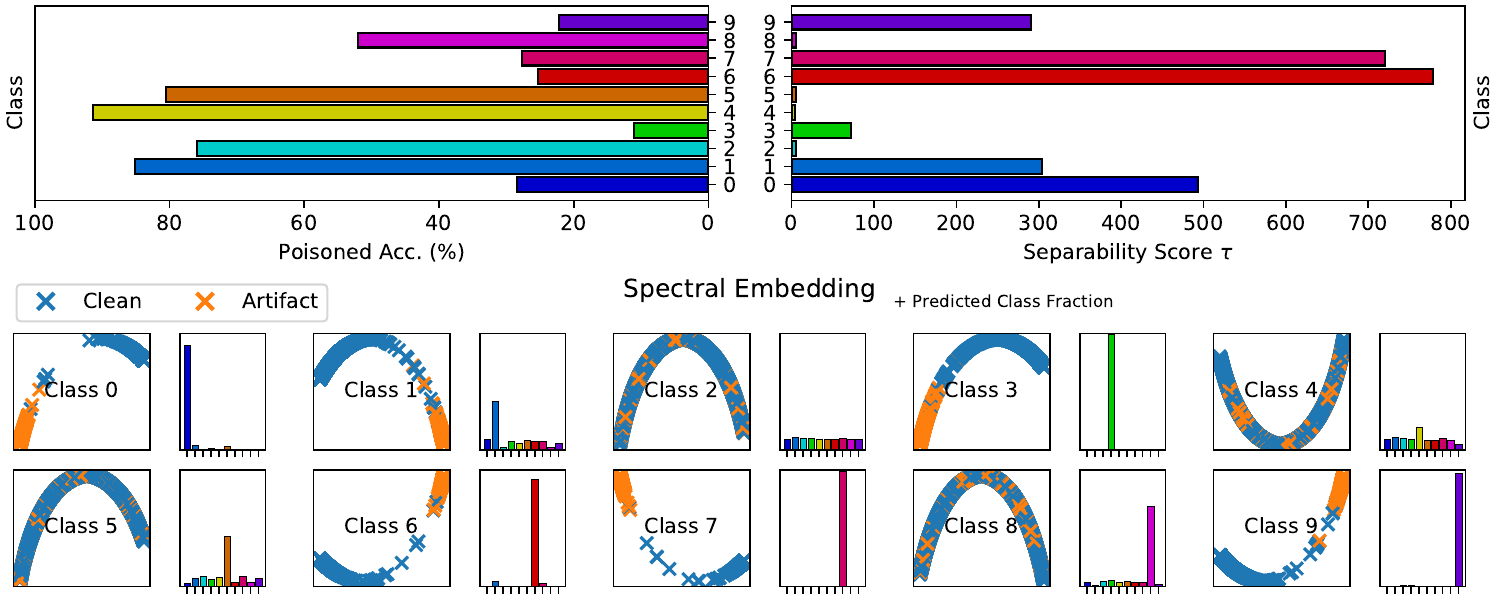}
        \caption{ %
            \textbf{Top:} Accuracy on poisoned dataset (left) and separability score $\tau$ of 2-cluster agglomerative clustering where the class included a clever hans.
            \textbf{Bottom:} Spectral Embedding (left) with 4 neighbors and 2 eigenvalues of each individual class on its corresponding dataset, where orange crosses are colored artifact samples and blue crosses are uncolored clean samples. The predicted class fractions are shown for each class to the right of its Spectral Embedding.
            For each of the 10 classes, a modified MNIST dataset was prepared where 20 percent of the samples in that particular class were colored to act as artifact samples.
            One model was trained on each of these datasets.
            The poisoned accuracy is the accuracy of each of these models on the validation set with each sample colored in the same artifactual color. The used colors were the same as the ones shown in \ref{fig:cmnist_as_such}.
            Models with a high poisoned accuracy and a low separability score indicate that the model has not learned the artifact.
            Models with a low poisoned accuracy and a high separability score indicate that the artifact was learned.
            The spectral embeddings show a clear split for models where the artifact was learned.
        }
        \label{fig:cmnist_spray}
    \end{figure}

    \paragraph{\glsdesc{spray} on Colored MNIST}
    The \gls{spray} framework is applied on colored MNIST setup as following.
    For each of the 10 MNIST classes, we create a dataset where for the corresponding class, samples are colored with a probability of 20 percent in a distinct color as shown in Figure \ref{fig:cmnist_as_such}.
    The rest of the samples are left in their original white color.
    On each of these datasets, a simple feed-forward convolutional neural network is trained (\cf~\ref{sec:appendix:networks:coloredmnist}).
    We can then verify for each model how much it has learned the color to be a distinct feature for the corresponding class, by evaluating the model accuracy and the fractions of the predicted classes on a validation set which has been completely colored in the color of the artifact.
    Subsequently, we do a \glsdesc{spray} by using 4 neighbors to build an affinity graph of the attributions to compute the spectral embeddings reduced to the dimensions corresponding to the 2 smallest eigenvalues.
    Note that we did not sum over the color channels of the attributions, as is often done for visualization purposes, since the color plays an important role in this experiment.
    We do a simple agglomerative clustering with 2 clusters on the spectral embedding, and compute its separability score $\tau$.
    The aforementioned results are visualized in Figure \ref{fig:cmnist_spray}.
    The spectral embeddings at the bottom of Figure \ref{fig:cmnist_spray} form a crescent-like shape for all classes.
    When the attribution can be well separated, clean and artifact samples move towards the opposite ends of the the crescent.
    This is visible for classes 0, 1, 3, 6, 7 and 9.
    These classes also show a high separability score $\tau$ when compared to the scores of the other classes.
    With the exception of class 1, all of these classes also show a low performance on the poisoned validation set, with an accuracy below 30 percent.
    In contrast, again with the exception of class 1, all classes with an accuracy above 50 percent show a separability score close to 0.
    The predicted class fraction for classes with a high separability score show a high tendency of the model to predict poisoned samples as the artifact class, especially for classes 0, 3, 6, 7 and 9.
    On classes 2 and 4, the model seems to not, or barely have picked up the artifact as a class-relevant feature.
    Class 8 shows a high confusion even though the separability score is close to 0.
    In contrast, class 1 shows comparatively low confusion even though its separability score is high.
    It is worth to note that all models show a reduced accuracy on the poisoned validation set compared to the accuracy on a clean validation set of 98 to 99 percent, even for class 2 where the confusion does not seem to focus on the artifact class.
    This means that even though we may confuse models by coloring all samples of the whole validation set, we cannot detect the artifact in some of these models using \gls{spray}.
    Only part of the reason for this seems to be that the model has not picked up the artifact during training, since for example class 8 shows a relatively high tendency to confuse colored samples for their corresponding artifact class, yet the \gls{spray} does not give any indication of an artifact in the class.
    
    Concluding this experiment, the assigned importance of an artifact may vary greatly between models and classes, and even though we may not find the artifact in all instances where the model has in fact picked up an artifact as an important feature, \gls{spray} pointed out most artifacts in this setup.
    
\begin{figure}[ht]
        \centering
        \includegraphics[width=0.6\linewidth]{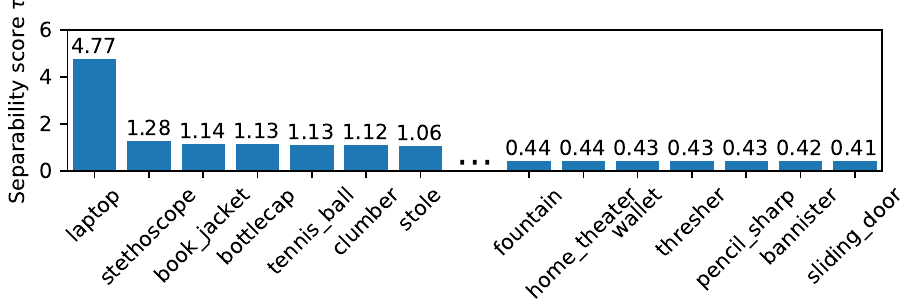}
        \caption{%
            Mean separability score $\tau$ of spectral embedding of attributions based on \gls{fda}.
            A high $\tau$ means there are significantly different decision strategies being used, potentially of \gls{ch} type.
        }
        \label{fig:fdaval}
    \end{figure}

\begin{figure}[t]
        \centering
        \includegraphics[width=.49\linewidth]{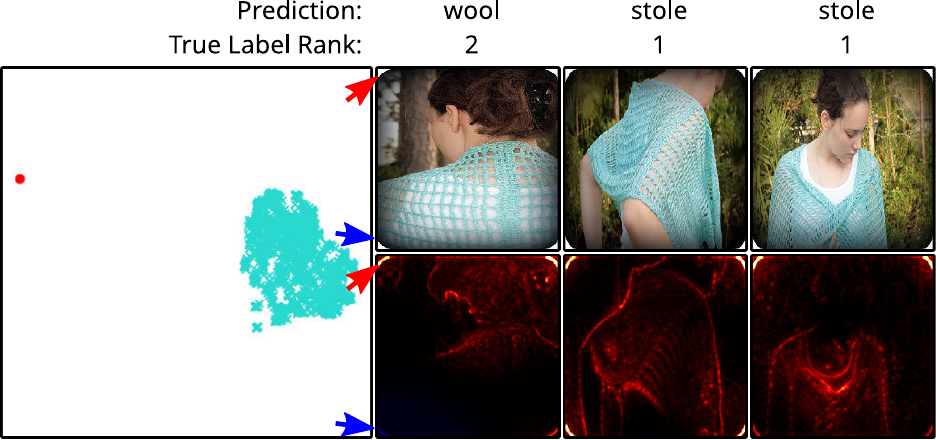}
        \includegraphics[width=.49\linewidth]{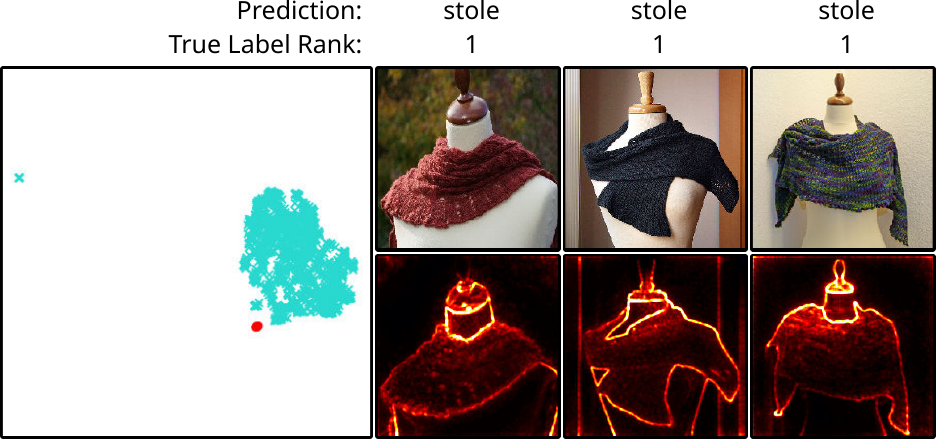}\\
        \vskip 1.5mm
        \includegraphics[width=.49\linewidth]{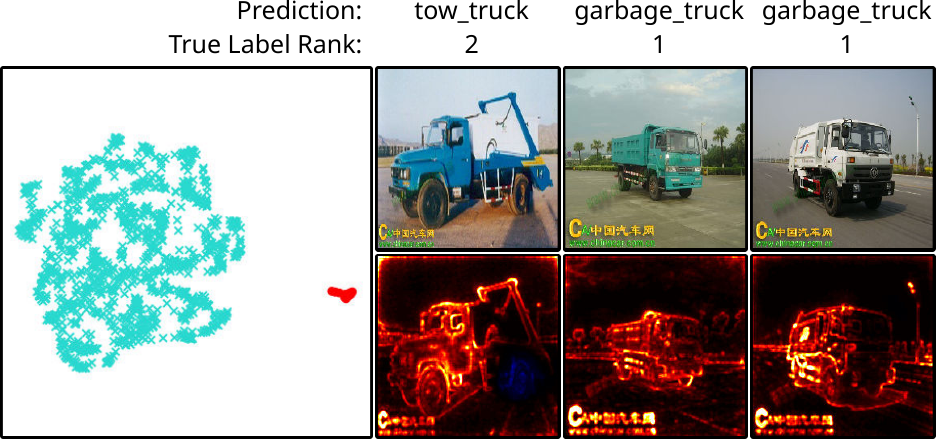}
        \includegraphics[width=.49\linewidth]{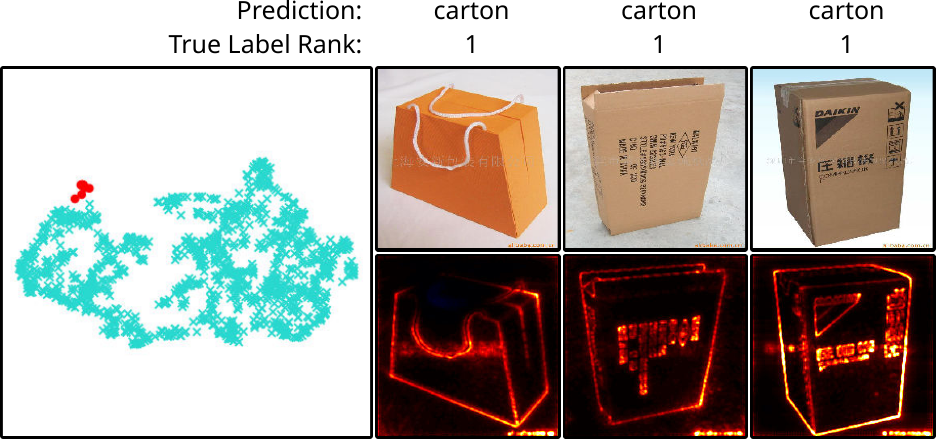}\\
        \vskip 1.5mm
        \includegraphics[width=.49\linewidth]{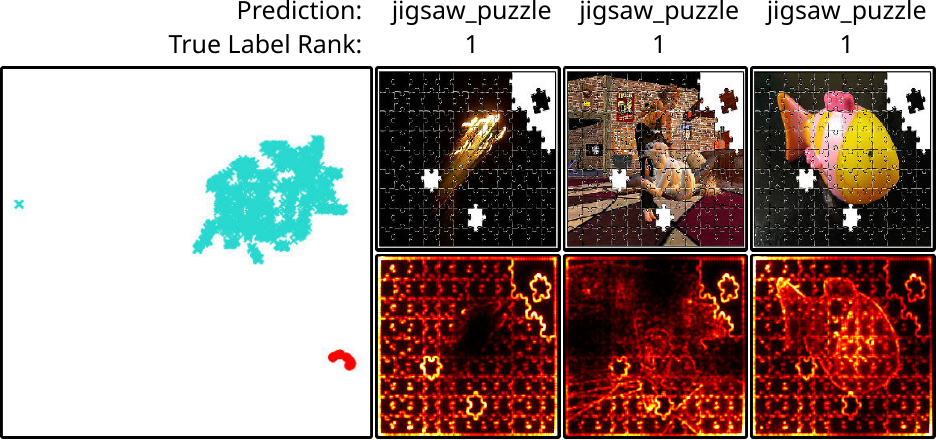}
        \includegraphics[width=.49\linewidth]{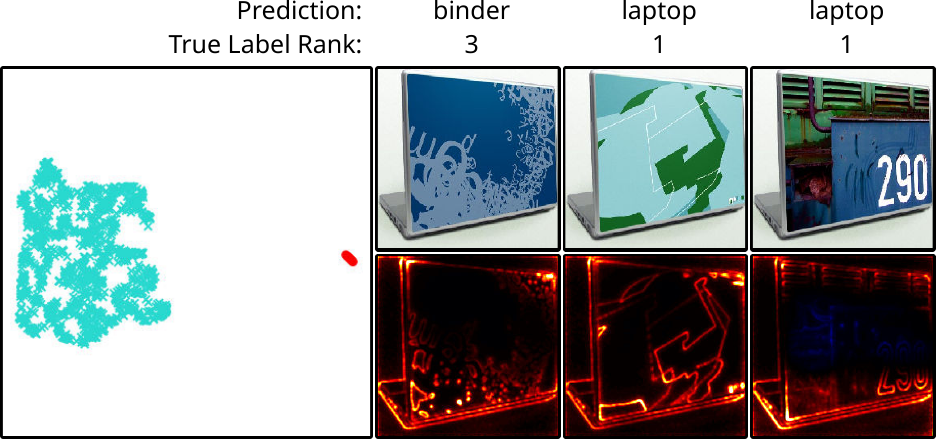}\\
        \caption{
            Each panel shows the UMAP (left) with samples and heatmaps (right) of significant clusters, highly separated from the rest of the samples.
            For each cluster, example images (top) and their respective attributions (from the identified cluster are shown.
            The location of the samples in the UMAP are highlighted in red.
            Attribution maps show relevant image regions \emph{supporting} the classifier decision in yellow-red, 
            irrelevant regions in black color and relevant regions \emph{contradicting} the final prediction
            in blue-cyan.
            Above the sample images the classifier's \mbox{top-1} predicted class and the prediction rank of the true label are shown.
        }
        \label{fig:experiments:spray:imagenet}
    \end{figure}

\paragraph{Quantifying \glsdesc{ch} Candidates on ImageNet}
    We examine ILSVRC2012 for \gls{ch} candidates by applying \gls{spray} with various clustering approaches for which we compute cluster separability scores $\tau$ (Eq. \eqref{eq:tau}) for each class.
    Figure~\ref{fig:fdaval} lists a ranking of the ImageNet classes with the highest and lowest $\tau$ values with a striking result for class laptop, due to a large cluster with copies of almost the same image (see UMAP of its spectral embedding in Figure~\ref{fig:experiments:spray:imagenet} (bottom right)).
    
    \begin{figure}[ht]
        \centering
        \includegraphics[width=.2\linewidth]{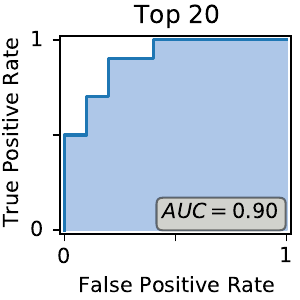}
        \includegraphics[width=.2\linewidth]{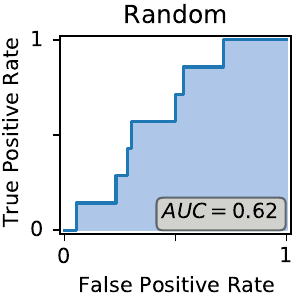}
        \includegraphics[width=.2\linewidth]{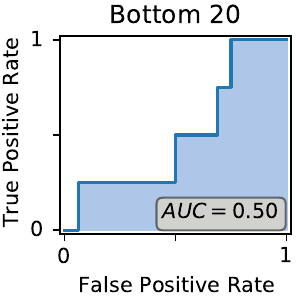}
        \caption{%
            \gls{roc} curves for artifact-existence versus \gls{fda}-Ranking.
            Left: Top 20 classes with highest values of $\tau$.
            Mid: 63 random classes with any values of $\tau$.
            Right: Bottom 20 classes with lowest values of $\tau$.
        }
        \label{fig:fdaroc}
    \end{figure}
    We inspect the validity of the class ranking for \gls{ch} candidates generated by \gls{fda} in a small experiment, by screening
    a subset of all 1000 ImageNet classes, namely
    (1) those with the 20 highest $\tau$ scores,
    (2) those with the 20 lowest $\tau$ scores and
    (3) 63 randomly picked classes.
    In all three cases, we assume a positive \gls{ch} ``prediction'' per class due to a large value of $\tau$.
    We then produce ``ground truth'' labels via manual assessment of the existence of a \gls{ch} candidate.
    We would like to remark that this ``ground truth'' has been established based on the class label description
    in the taxonomy of the ImageNet dataset and our subjective human understanding of the image content.
    Using this information we produce \gls{roc} curves and corresponding \gls{auc} values.
    
    The results show a clear picture 
    validating that a high $\tau$ score is indeed a strong indicator for the presence of \gls{ch} candidates
    (Figure~\ref{fig:fdaroc}~(left), high \gls{auc}).
    Both randomly selected or bottom 20 classes (Figure~\ref{fig:fdaroc}~(mid, right))
    yield essentially random \gls{auc} scores due to only sporadically encountered \gls{ch}.
    However, the $\text{\gls{auc}}\gg0$ here also show that even a $\tau$ rating in the lowest 2-percentile does not guarantee a class to be free of \gls{ch} behavior. 
    Summarizing, large $\tau$ is an excellent indicator for \gls{ch} behavior, but small $\tau$ is no ultimate guarantee for their absence, so further research will be needed here to ideally bring forward indicators that can provide a theoretical bound for absence of \gls{ch} behavior.
    
\paragraph{Inspecting and Isolating \glsdesc{ch} Candidates}

    Based on the ordering by \gls{fda} and $\tau$ established in the previous section,
    we manually investigate whether the \gls{ch} candidate classes show prominent \gls{ch} artifacts to be expected.
    The \gls{spray} framework provides as a side effect (through its spectral embedding space $\Phi$) also a basis for visualizing clusters of heatmaps, here we use UMAP.
    Promising clusters are often located far away from the rest of datapoints in the UMAP embedding, see \eg~Figure~\ref{fig:experiments:spray:imagenet} center left the UMAP scatter-plot of class ``garbage truck''.
    There, the red cluster-members all show examples of images of the same watermark with high attribution in LRP.
    Another intriguing example is the top middle UMAP plot of class ``stole'':
    while not as separated as for other examples, we find a cluster of mannequins wearing stoles, with high attribution scores on the mannequin's ``head''.
    For class ``carton'', we can see even two artifacts at the same time: watermark written with Hanzi in the center of the image, as well as a watermark in latin characters in the bottom right.
    The bottom right watermark is in fact not only present in the carton class.
    
    Based on the clustering labels provided by \gls{spray}, for each artifact, we may extract a set of labels that indicate whether a sample is affected by the artifact candidate.
    Using these labels along with the corresponding samples, we may estimate an artifact map according to Section \ref{sec:methods:clarc}, which given a clean sample creates a poisoned version of the sample with the artifact present.
    This may be done for example by training a generative model conditioned on the presence or absence of an artifact, manually extracting a watermark from an affected image using an image manipulation framework, or something as simple as fitting a linear regression model.
    For \gls{aclarc} in input space, we manually extract the artifact from samples labeled as poisoned, such that we can apply it to samples by a simple affine transformation $h(x) = (I - \text{diag} (\alpha))x + \text{diag} (\alpha)z$ where $z$ is a vector with the pixel values of the watermark, and $\alpha$ an alpha channel the same size as the number of pixels, which is zero for all pixels except the ones where the watermark is present.
    For \gls{pclarc}, we instead use the labels to train a linear classifier $f(x) = v^T x + b$ with $\|v\| = 1$, which is used to instead estimate an inverse artifact map as an affine transformation $h(x) = (I - vv^T) x - vv^T z$, where z is chosen as the mean over all clean samples of the class, as highlighted in Section \ref{sec:methods:clarc}.

\FloatBarrier
\subsection{\glsdesc{spray} on ImageNet in Feature Space}
\label{sec:experiments:featurespray}
\FloatBarrier

Until now we have based our \gls{spray} solely on model attributions in input space.
While this has not been explored by \citet{lapuschkin2019unmasking}, we attempt base the analysis on model attributions in feature space for additional insight and compare the obtained separability scores over the various intermediate representations at different model depths.
The motivation behind using intermediate representations is that the model must encode increasingly invariant representations of concepts towards its classification task in higher levels,
which may not be detectable with the contribution scores in input space.
We investigate which clusters of samples contribute the most towards the separability score of a given class.
To this end, we compute the score $\tau$ as many times as there are clusters,
with samples from one cluster withheld in each iteration.
In this setting the cluster group with the lowest separability score will have left out the cluster of samples with the highest contribution to the outlierness of the class. 
Complete class separability scores, along with samples of clusters with the highest outlier score, are reported in Figure \ref{fig:experiments:spray:imagenet}.

 \begin{figure}[ht!]
        \centering
        \includegraphics[width=.75\linewidth]{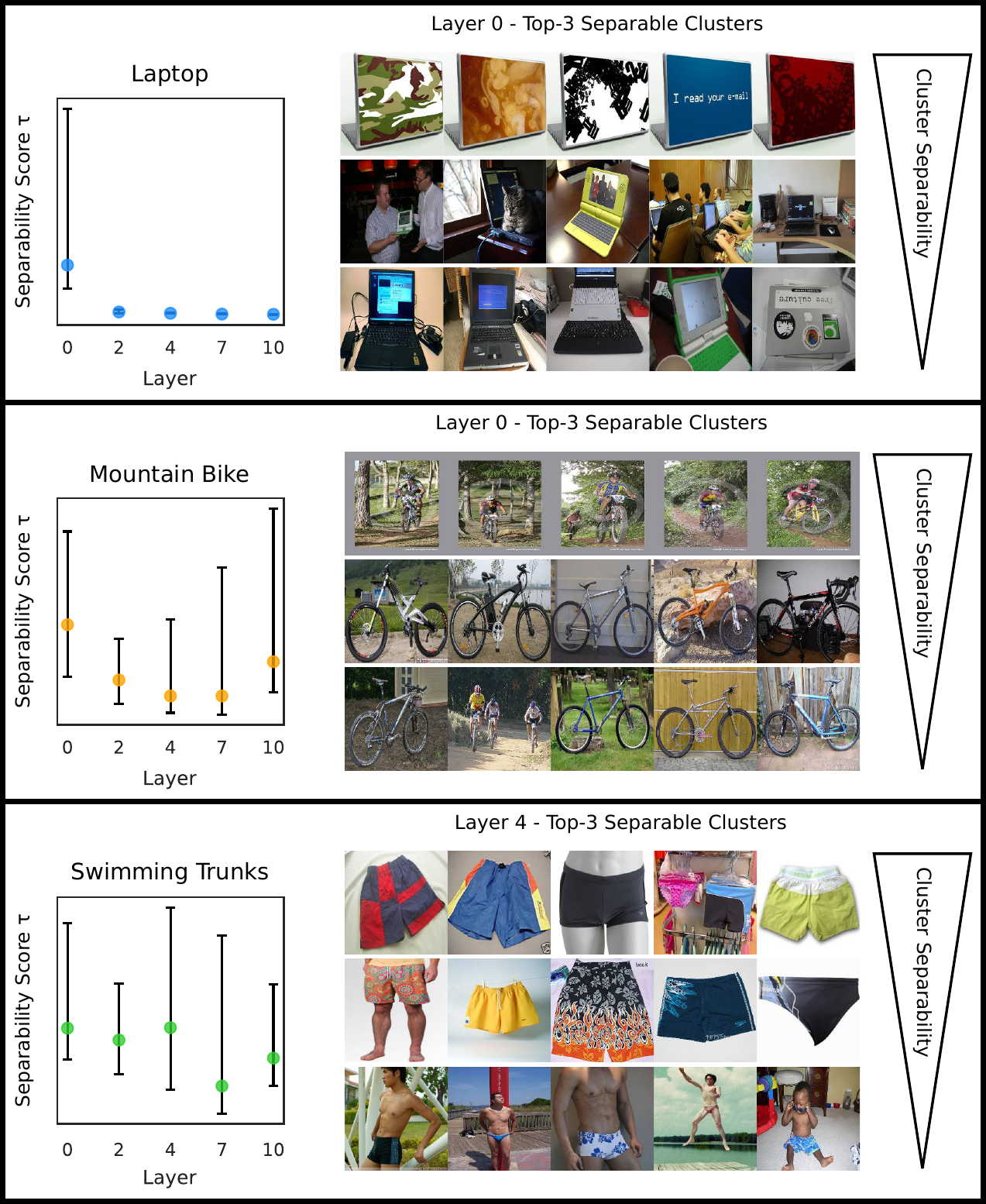}
        \caption{ %
            Separability of various clusterings of spectral embeddings $\Phi$ in multiple layers of VGG16. The measurements of the \gls{fda} scores \textit{(left)} at each layer are varied over the number of clusters chosen for \gls{spray}.
            The shown ImageNet classes are laptop \textit{(top)}, mountain bike \textit{(middle)} and swimming trunks \textit{(bottom)}. Note that the measured absolute magnitude of the separabilit score $\tau$ might be different between the three classes, so that only relative within-class comparisons can be inferred here. The scores $\tau$ vary strongly over various layers for different classes. \Eg, the \gls{fda} score for ``laptop'' is comparatively large at the input layer, but then decreases with increasing depth of the layer. ``Swimming Trunks'', on the other hand, seems to separate best at layer 4.  For this layer of maximum separation score, examples of the top three separating clusters are shown to the \textit{(right)}, revealing possible \gls{ch} artifacts.
        }
        \label{fig:featurespray}
    \end{figure}

In this figure,
the above analysis is shown for the three example classes ``laptop'', 
``mountain bike'',
and ``swimming trunks'' \textit{(top to bottom)}.
Within each panel,
a relative comparison of the separability score $\tau$ over layers,
\ie,
the input layer (layer 0) and various intermediate layers obtained from the model's convolutional feature extractor (layers 2-10) can be found to the \textit{(left)}.
At each layer,
measurements vary over the chosen number of clusters $K \in \{2, \cdots, 32\}$,
with the respective mean shown as a colored dot.
However,
a high $\tau$ does not necessarily \emph{only} occur due to the presence of a \gls{ch},
although \emph{if} a \gls{ch} is present and well represented, 
a high separability score is likely.
Thus,
correspondingly on the \textit{(right)} side of each panel,
for $K = 32$ and the layer with the \emph{highest} mean $\tau$ in Figure~\ref{fig:featurespray}~\textit{(left)}, 
samples of the top three clusters in terms of contribution to separability
(\ie, 
the separability score decreased the most when this cluster was left out) are visualized.
The most contributing cluster is shown in the \textit{(top row)},
decreasing towards the \textit{(bottom row)}.

We find that the separability scores vary significantly with the layers:
for the ``laptop'' class, 
the clearly highest separability score appears at the input layer. 
Here,
a cluster showing laptop lids has the largest separability contribution,
showing the same laptop (albeit with different patterns printed on its lid),
digitally rendered from the same angle in each sample in front of a white background.
Thus,
this cluster seems to describe a \gls{ch} artifact.
Results for the ``mountain bike'' class behave in a similar manner.
Again, 
the highest separability score is found at the input layer, 
and,
correspondingly,
the cluster with the highest separability contribution there seems to contain a \gls{ch} in the form of a distinctive gray border and a watermark.

In contrast to the first two examples,
the largest mean $\tau$ value for the class ``swimming trunks'' occurs not at the input layer,
but at intermediate layer 4 of the model instead.
Again,
the top contributing cluster consists of relatively similar samples,
however,
they are all perfectly \emph{valid} examples of ``swimming trunks'',
with no distinguishable artifact between them.
The same seems to be the case for the second most contributing cluster.
Interestingly,
the third most separable cluster is extremely dissimilar to the first two, 
with every sample containing male upper bodies -- a feature that, 
while often appearing alongside ``swimming trunks'' should not indicate this class in any way. 
In other words, 
a \gls{ch}.

This last example demonstrates why it may be difficult to automate the process of \gls{ch} identification:
While a \gls{ch} is in fact present in the class,
it is not the \emph{top} separating cluster,
but has the third highest contribution (of 32 total clusters) to the $\tau$ score instead.
More concisely,
the most separable cluster is not necessarily a \gls{ch},
and a high separability score does not guarantee the presence of a \gls{ch}. 
Thus,
\gls{spray} offers an indication of which clusters in which classes are \gls{ch} \emph{candidates},
but -- in accordance with the property of \gls{ch} artifacts of requiring expert domain knowledge to detect (Section \ref{sec:intro:rel_work}) -- human judgement is still required for a final decision.
We further note that the \glspl{ch} found in the first two examples are relatively simple features.
They can, 
in fact, 
be expressed as an affine transformation in input space.
Correspondingly, 
the highest separability score for these classes occurs in input space.
In contrast,
the third presented example,
where the ``upper body'' \gls{ch} was identified,
is far more complex,
but the highest $\tau$ score is also found at a deeper intermediate layer.
Thus,
there seems to be a correlation between the \emph{complexity} of an artifact, 
and the \emph{depth} of the layer at which it separates best from the rest of the class.

\FloatBarrier

\section{Experiments -- Concept Desensitization}
\label{sec:experiments:unlearning}

In the previous section, we obtained cluster labels for (potential) \gls{ch} artifacts,
and correspondingly are able to estimate artifact models for \gls{ch} candidates in ILSVRC2012 according to Section~\ref{sec:methods:clarc}.
The goal of this section is to verify the impact of these artifacts candidates on our classification model and at the same time reduce their impact by using either \gls{aclarc} or \gls{pclarc}.
We first verify \gls{aclarc} empirically by introducing a controlled setup based on a variation of MNIST, where artifacts are introduced as colors.
With the established verification, we proceed an attempt to unlearn \gls{ch} artifacts candidates using \gls{aclarc} given the artifact estimators modeled after Section~\ref{sec:experiments:identification}, first in input space, then in feature space, and at the same time measure their respective impact on the classification model.
We then proceed to verify \gls{pclarc} empirically on a setup similar to the previous one on a variation of colored MNIST.
Subsequently, an extensive analysis using \gls{pclarc} on ILSVRC2012 is presented,
followed by an analysis on the ISIC~2019 dataset.
Finally, we report results on the Adience dataset using \gls{pclarc}, touching upon the issues of fairness and robustness in machine learning.

    \subsection{Unlearning Concepts with \glsdesc{aclarc}}
    \label{sec:experiments:unlearning:aclarc}
    
        After identifying several \gls{ch} artifacts of the ILSVRC2012 in Sections \ref{sec:experiments:inputspray} and \ref{sec:experiments:featurespray},
        we aim to desensitize models to them in the following experiments, 
        firstly by employing the proposed \gls{aclarc} method.
        \gls{ch} artifacts appear -- by definition -- alongside desired features of a class. 
        Furthermore,
        each \gls{ch} only natively occurs within one class and helps a model predict this class correctly.
        As such, 
        if unlearning is successful,
        a \emph{decrease} in the measured accuracy (as opposed to the \emph{true} generalization accuracy) is to be expected, 
        making it difficult to distinguish from simply confusing the network.
        Due to these unique properties of \gls{ch} artifacts,
        our method for evaluating the experiments is two-fold: 
        A quantitative evaluation of whether \gls{aclarc} leads to a desensitization against a concept representation,
        combined with a qualitative assertion of whether this representation corresponds to the target concept and leads to an unlearning thereof.
        
        \paragraph{ \glsdesc{aclarc} on Colored MNIST}
            As an empirical verification of the method,
            \gls{aclarc} is applied on a simple convolutional feed-forward type network (\cf~\ref{sec:appendix:networks:coloredmnist}) on the previously described MNIST dataset with color artifacts.
            Here, we train the three variants of the model:
            (1) For the first model, of the 10 different classes, the samples of one class are colored with a probability of 20 percent during training.
            We call this the \emph{native model}.
            (2) Another model is trained, but in addition to coloring the same single class as before, we also color samples of all other classes with a probability of 20 percent.
            We call this model \emph{a priori ClArC}.
            (3) For the third model, we continue training from the learned \emph{native model}, but also color according to the \emph{a priori ClArC} samples of all classes with a probability of 20 percent.
            This model we call \emph{a posteriori ClArC}.
            
            To evaluate the influence of the color-based \gls{ch}, we introduce two test modes.
            One test mode describes the performance of the models on the real dataset, where samples of the \gls{ch} class are colored with a probability of 20 percent.
            The second test mode describes the performance of the models on a maximally poisoned dataset, where every sample is colored. 
            By comparing these two performances, we get a measure of the error caused by the \gls{ch}.
            Note that in this toy setting we can actually measure the performance of the model on the clean, \gls{ch} free dataset, which would normally not be available.
            The performance on the realistic dataset is as one would expect marginally better (around 0.02 percent) than the performance on the clean dataset for the \emph{native model}.
            However, when comparing these quantities to the fully poisoned dataset, they do not differ very much, and thus we compare the realistic setting to the fully poisoned setting.
            
                \begin{figure}[ht!]
                    \centering
                    \includegraphics[width=.3\textwidth]{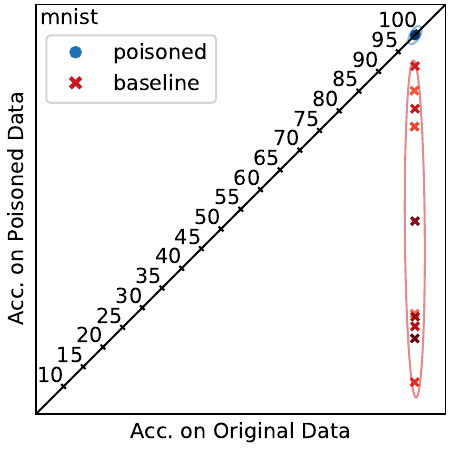}
                    \caption{ %
                        Accuracy on a realistic test set (x-axis) vs. accuracy on a fully poisoned test set (y-axis) on colored MNIST.
                        Red crosses describe the baseline, a \emph{native model} which has seen a \gls{ch} artifact during training for 20 percent of the samples of one class.
                        Blue dots describe a fine-tuned version according to \gls{clarc}, the \emph{a posteriori ClArC}, of the aforementioned models.
                        The red and blue ellipses describe the confidence of the points.
                        For visualization purposes, the ellipses are drawn with $40 \sigma$ in x- and y-direction for \emph{a posteriori ClArC} (poisoned) and $40 \sigma$ in x-direction and $1.4\sigma$ in y-direction for the \emph{native models} (baseline), where $\sigma$ is the standard deviation of the accuracies in the respective direction.
                    }
                    \label{fig:cmnist_clarc_scatter}
                \end{figure}
            
            Figure \ref{fig:cmnist_clarc_scatter} shows the accuracy on the realistic test set on the x-axis and the accuracy of a maximally poisoned dataset (all samples colored) on the y-axis.
            The \emph{native models} (baseline) are represented by red crosses, while the \emph{a posteriori clarc} (poisoned) are represented by blue dots.
            All models achieve an accuracy of about $99$ percent on the realistic test set.
            As one would expect, the \emph{native models} perform considerably worse on the fully poisoned test set.
            Some models are only slightly impacted by the poisoning, which means they do not pay as much attention on the \gls{ch} artifact (color).
            Other models however perform as bad as only $10$ percent accuracy, which has the model predict the class only based on the \gls{ch} artifact.
            Fine-tuning the model according to \gls{clarc}, as done for the \emph{a posteriori ClArC} model, results in all models now performing very closely to how they perform on the realistic dataset.
            Therefore, the models have successfully been fine-tuned to ignore the \gls{ch} artifact.
            We have therefore empirically shown the effect of \gls{ch} artifacts on the model, as well as shown the effectiveness of \gls{clarc}.

\begin{figure}[ht!]
            \centering
            \includegraphics[width=.4\textwidth]{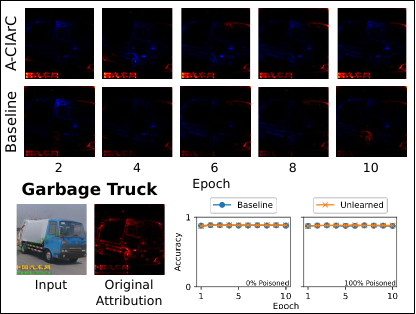}
            \includegraphics[width=.4\textwidth]{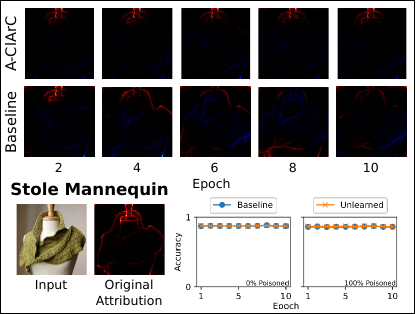}
            \includegraphics[width=.4\textwidth]{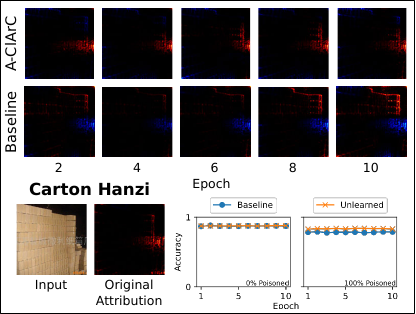}
            \includegraphics[width=.4\textwidth]{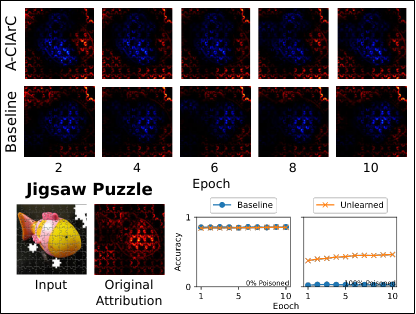}
            \caption{ %
                In each panel: an input example with a \gls{ch} artifact alongside its attribution of the original model (bottom left).
                The pixel-wise difference between the original attribution and attributions for an \gls{aclarc} fine-tuning at every 2 epochs (top), with attributions of a baseline model below which was fine-tuned with the same hyperparameters, but without modifying the training set.
                Red means the orignal model assigns more relevance to the highlighted part, while blue describes the opposite.
                The model performance (bottom right) is shown on both the unchanged validation set (0\% poisoned) and the poisoned validation set (100\% poisoned) over the epochs for both the baseline model and the A-ClArC model (unlearned).
            }
            \label{fig:experiments:aclarc:imagenet:garbage}
        \end{figure}

\begin{figure}[b]
            \centering
            \includegraphics[width=.30\linewidth]{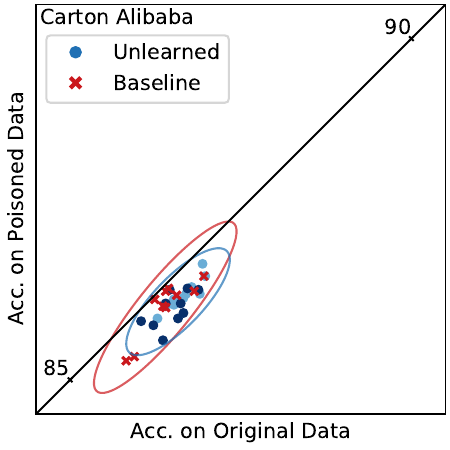}
            \includegraphics[width=.30\linewidth]{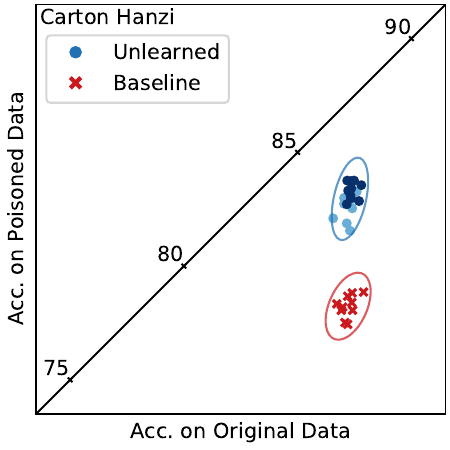}
            \includegraphics[width=.30\linewidth]{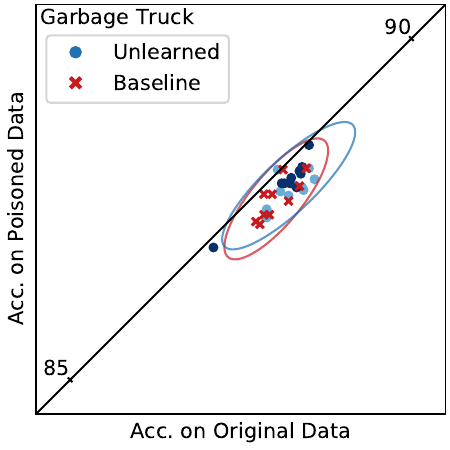}

            \includegraphics[width=.30\linewidth]{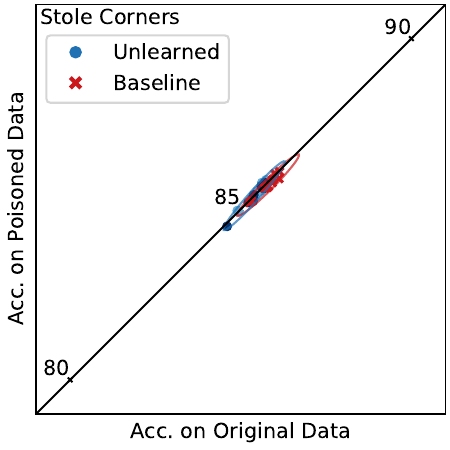}
            \includegraphics[width=.30\linewidth]{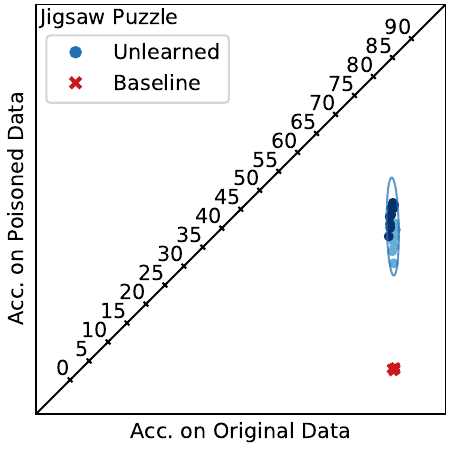}
            \includegraphics[width=.30\linewidth]{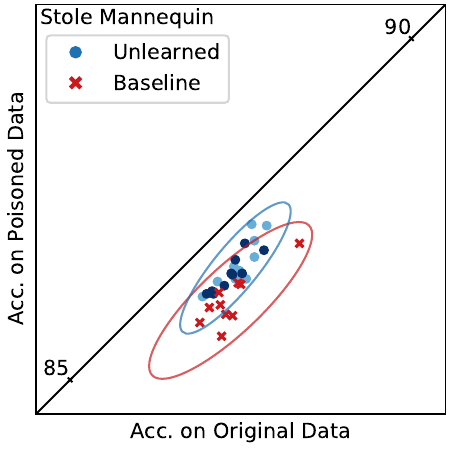}

            \caption{
                Original (x-axis) vs. poisoned (y-axis) validation set accuracy of baseline model (red) and A-ClArC models (blue) over training poison rate (bright 20\%, dark 50\%).
                All points are below the line of equal accuracy on poisoned and original data, which means they consistently perform better on the clean dataset.
                For ''carton hanzi`` and ''jigsaw puzzle``, the unlearned models perform significantly better on the poisoned validation set than the baseline.
                This can also be seen less significantly for ''stole mannequin``.
                In all cases, the accuracy of the unlearned model does not visibly decline compared to the baseline model.
                In the case of ''jigsaw puzzle``, the artifact is for many samples the only class-defining feature, which therefore extremely confuses the model on the poisoned validation set.
            }
            \label{fig:experiments:aclarc:imagenet:scatter}
        \end{figure}

        \paragraph{\glsdesc{aclarc} on ImageNet}
        We conduct a similar setup to the one used on Colored MNIST on ILSVRC2012.
        Due to the size of the dataset, we only use the previously described \emph{native model}, which is the model trained on the original training set with all natural artifacts included, and the \emph{a posteriori \gls{aclarc} model}, which is a fine-tuned version of the \emph{native model}.
        Additionally, we introduce a \emph{baseline model}, which is fine-tuned with the same hyperparameters as the \emph{a posteriori \gls{aclarc} model} but trained on the unmodified training set.
        Furthermore, we reduce our training to a subset of $100$ classes of the original ILSVRC2012. 
        One \emph{a posteriori \gls{aclarc}} and one baseline model is trained for each artifact candidate model we have identified in Section \ref{sec:experiments:identification}.
        We fine-tune on the original model for a total of $10$ epochs, and report the model accuracies using the two previously introduced test modes, where we use the original validation set (0\% poisoned) as well as the original validation set with the artifact introduced into all samples (100\% poisoned) in Figure \ref{fig:experiments:aclarc:imagenet:garbage}.
        The model performances are also compared for these two test modes in scatter plots in Figure \ref{fig:experiments:aclarc:imagenet:scatter}.
        As an additional approach to evaluate whether the importance of artifact was reduced for the prediction of each model, we visualize the difference between the attribution of the original model, and either the \emph{a posteriori \gls{aclarc}} or the \emph{baseline} model in each epoch in Figure \ref{fig:experiments:aclarc:imagenet:garbage}.
        
        The performance for both the \gls{aclarc} and the baseline model do not seem to change considerably for artifacts ``stole'' and ``garbage truck'' when looking at the performances in Figure \ref{fig:experiments:aclarc:imagenet:garbage}.
        This can be seen by very similar confidence ellipses in Figure \ref{fig:experiments:aclarc:imagenet:scatter} over all epochs and an additional poisoning of the training data at 50\% for class ``garbage truck''.
        Class ``stole mannequin'' in Figure \ref{fig:experiments:aclarc:imagenet:garbage} which corresponds to ``stole mannequin head'' in Figure \ref{fig:experiments:aclarc:imagenet:scatter} shows however a slight improvement in the poisoned validation mode in the latter.
        Class ``carton Hanzi'' in Figure \ref{fig:experiments:aclarc:imagenet:garbage} which corresponds to ``carton Hanzi'' in Figure \ref{fig:experiments:aclarc:imagenet:scatter} shows a clear improvement over the baseline model on the poisoned validation set.
        We can see a strong collapse in the performance on the poisoned dataset for ``jigsaw puzzle'' for the baseline model, likely caused by the fact that the artifact is a very clear indicator for the class.
        However, the \gls{aclarc} returns to about 50 \% of accuracy on the poisoned validation set.
        For none of the artifacts in Figure \ref{fig:experiments:aclarc:imagenet:garbage} we see the \gls{aclarc} model perform worse than the baseline model.
        By investigating the heatmap differences of the \gls{aclarc} and baseline model to the original model, we can see that the \gls{aclarc} model consistently decreases the amount of relevance assigned to the artifact location in the input image.
        For ``garbage truck'', there is a watermark in the bottom left corner of the image.
        The \gls{aclarc} attribution subtracted by the original attribution shows strong positive values on the watermark location, indicated by the red color.
        The baseline attribution difference to the original model seems to decrease and increase the relevant pixels more generally focused on edges in the image.
        Even though the baseline model also seems to weakly reduce the relevance on the watermark in the second epoch, this is not as targeted and consistent as for the \gls{aclarc} model.
        A similar behavior can be seen for the class ``stole mannequin'', where the \gls{aclarc} model  consistently reduces the relevance on the mannequin, while the baseline partly even reduces the relevance on the stole itself.
        For the ``carton Hanzi'' artifact, which here corresponds to the Hanzi in the center of the image, the \gls{aclarc} model also reduces the relevance on the characters, mostly concentrated at the higher contrast area at the right hand side of the image.
        The baseline model even increases the relevance on the location of the watermark and decreases the relevance on the actual cartons compared to the original model.
        While somewhat harder to see, the \gls{aclarc} model seems to reduce the jigsaw pattern away from the object of interest that the baseline model for ``jigsaw puzzle''.
        
        Similarly, Figure \ref{fig:experiments:aclarc:imagenet:scatter} gives similar insights for ``carton chinese watermark'' and ``jigsaw puzzle cutting pattern'', where all \gls{aclarc} models (poisoned) perform comparably on the original dataset, but outperform the baseline significantly on the poisoned validation set.
        With ``stole mannequin head'', \gls{aclarc} outperforms the baseline slightly.
        ``carton alibaba watermark'' seems to only weakly affect both models, with no visible improvement for \gls{aclarc}.
        The alibaba watermark is found not only in class ``carton'', but in many classes of ILSVRC2012, and is a rather small artifact in the bottom right of the image, possibly cropped most of the time during training, which is why it may not be a very strong artifact for class ``carton'' alone.
        ``stole rounded edges'' is also a very small artifact at the corner of only a few samples in class stole.
        Presumably for this reason, we do not see the either model particularly impacted by poisoning the validation set.
        The ``garbage truck'' artifact result is somewhat surprising, both models seem to be only slightly affected poisoned dataset, with only at best a very slight improvement of the \gls{aclarc} model over the baseline model.
        
        Therefore, we may conclude that \gls{aclarc} in input space does seem to work for some artifacts that are very significant in input space, but may not show any significant effect otherwise.
            
        \FloatBarrier

        \paragraph{ \glsdesc{aclarc} in Feature Space}
        Equivalently to the previous section, we may instead choose to do a fine-tuning with \gls{aclarc} using artifact representations that we have found in the feature space of any layer of a neural network in Section \ref{sec:experiments:featurespray}.
        There, 
        we noted that these intermediate representations of artifacts differ significantly in how well they can be separated via \gls{spray}. 
        For each artifact, 
        a different layer depth seems to allow for an optimal separability,
        and this depth seems to correlate to the complexity of the respective artifact.
        
        Building on those observations, 
        we conduct another experiment,
        similar to the one in \textit{Augmentative Class Artifact Compensation on ImageNet},
        applying \gls{aclarc} using feature space representations of the target \gls{ch} concept.
        In contrast to the input space variant of \gls{aclarc},
        here,
        the target \gls{ch} concept is represented via \glspl{cav}.
        
        We again compare an \textit{unlearned model} (corresponding to the \emph{a-posteriori \gls{aclarc}} introduced previously) that is fine-tuned for 10 epochs on a subset of ILSVRC2012 consisting of 100 classes and employs \gls{aclarc} to a \textit{baseline model} that is fine-tuned in the same manner, 
        but without \gls{aclarc}.
        Both models are initialized from the \textit{native model}, 
        which is the same VGG-16 as for \textit{Augmentative Class Artifact Compensation on ImageNet}.
        For \gls{aclarc},
        the target \gls{ch}-artifact -- described by a \gls{cav},
        \ie,
        a direction in feature space -- is added during fine-tuning to the activations at the respective layer $l$,
        with a probability $p$ of 50\% and a contribution of 50\%. 
        The contribution denotes in which ratio the original activations and the added \gls{cav} are mixed. 
        This method of introducing the \gls{cav} to the activations corresponds to $a_i = 0.5 $ $\forall i \in \{1, 2, \cdots, d\}$ w.r.t. Equation \ref{eq:blending}), 
        and is used multiple times over Section \ref{sec:experiments:unlearning:aclarc} (for concept desensitization and evaluation) and \ref{sec:experiments:unlearning:pclarc} (only for evaluation).
        As described in Section \ref{sec:methods:clarc},
        the parameters of layers $\{1, \cdots, l\}$ are not altered during training,
        to keep the feature representation of the target concept static.
        Again,
        we employ the two test modes described previously,
        reporting accuracies on the original (0\% poisoned) and a poisoned validation set (100\% poisoned).
        The poisoning process,
        however,
        is executed in feature space for this experiment,
        using the computed \gls{cav} to poison the activations at layer $l$ instead of introducing the artifact in input space.
        This experiment is repeated for feature extractor layers $l \in \{0, 4, 10\}$,
        with layer 0 denoting the input layer.
        
\begin{figure}[ht!]
            \centering
            \includegraphics[width=.90\linewidth]{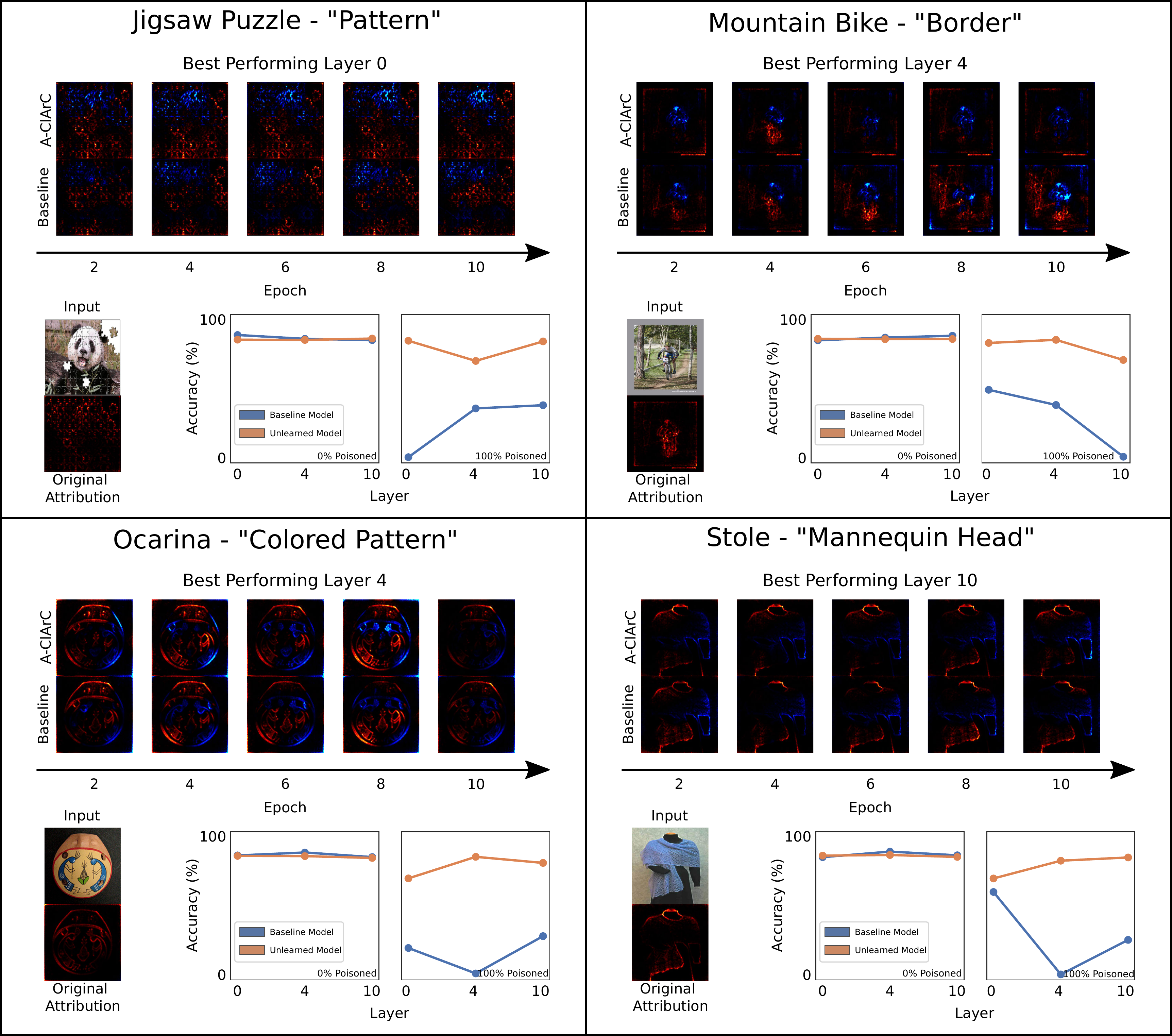}
            \caption{ %
                \gls{aclarc} in feature space for four example \glspl{ch}.  \textit{(Bottom)} of each panel: Accuracy of the baseline that was fine-tuned without \gls{aclarc} and the unlearned model that employed \gls{aclarc} on progressively more poisoned data. The poisoning is performed by adding the \gls{cav} computed for \gls{aclarc} to the activations after the respective layer. The unlearned model vastly outperforms the baseline with increasing poison rate, showing that it has grown relatively invariant to the target \gls{ch} concept. \textit{(Top)} of each panel: Example image and corresponding \gls{lrp} relevance differences to the original model (i.e., initial starting model before training) that used \gls{aclarc} are shown over training for the respective best performing layer, i.e., the layer where the accuracy of the unlearned model in a poisoned setting is largest. Here, the \textit{red} areas of the attribution maps are used \textit{less} than the original model for prediction; the \textit{blue} areas \textit{more}.
            }
            \label{fig:featurespace_aclarc}
        \end{figure}

        The results of this experiment are summarized in Figure \ref{fig:featurespace_aclarc}. 
        The four \glspl{ch} shown there (``pattern'', ``border'', ``colored pattern'', ``mannequin head'') are chosen to range from relatively simple to quite complex concepts. 
        
        At the  \textit{(bottom right)} of each panel of this figure,
        the validation accuracy after the final epoch is visualized for the three investigated layers.
        Results for both previously discussed test modes are shown:
        To the \textit{(left)}, 
        \ie,
        for the 0\% poisoning setting,
        we note that the unlearned model performs equally well as the baseline.
        The application of \gls{aclarc} did thus not affect the model's accuracy on unpoisoned data in a negative manner,
        indicating that it does not confuse a model unnecessarily or introduce any unfair biases.
        To the \textit{(right)},
        the 100\% poisoning setting is reported.
        Here,
        the  unlearned model \emph{vastly} outperforms the baseline model for every single \gls{ch} example.
        However,
        its accuracy varies w.r.t the layer at which \gls{aclarc} is applied -- as could be expected based on the results from Section \ref{sec:experiments:featurespray},
        where we found the separability score $\tau$ of samples containing a \gls{ch} artifact from clean samples to be quite dependent on the layer where \gls{spray} is applied. 
        Moreover, 
        the layer where performance is best seems to correlate to the perceived complexity of the \gls{ch}.
        \Eg,
        for the ``pattern'' artifact in class ``jigsaw puzzle'', 
        which can easily be represented via an affine transformation in input space,
        thus being a relatively simple \gls{ch},
        the unlearned model performs best at the input layer.
        In contrast,
        for the far more complex ``mannequin head'' from the ``stole'' class,
        the highest accuracy is retained for intermediate layer 10.
        In fact,
        for all closer investigated \glspl{ch},
        the performance of the unlearned model at the ``optimal'' layer in the \textit{poisoned} setting is almost on par with the performance of both unlearned and baseline models on the \emph{unpoisoned} setting,
        demonstrating a significant gain in invariance against the concept described by the \gls{cav} after applying \gls{aclarc}.
        
        However,
        since we employ the computed \gls{cav} to poison the validation data,
        the above evaluation only shows that invariance against the target concept is gained,
        if the \gls{cav} represents that concept correctly.
        Thus,
        to ascertain whether in fact the \emph{target} concept is unlearned via \gls{aclarc},
        we interpret Figure \ref{fig:featurespace_aclarc} (\textit{(top)} of each panel),
        where (\gls{lrp}) attribution difference maps of one example image per class are shown over epochs for the respectively best performing layer.
        For reference, 
        the original image of each of these examples and the corresponding \gls{lrp} attribution map of the native model can be found to the \textit{(bottom left)} of each panel.
        In the attribution difference maps,
        \textit{(red)} colors indicate areas that lost relevance compared to the native model,
        while \textit{(blue)} colors show a gain in relevance.
        Each \textit{(top)} row of attribution difference maps emphasizes these changes for the unlearned model that employs \gls{aclarc},
        while each \textit{(bottom)} row does the same for the baseline instead.
        
        For class ``jigsaw'',
        at the input layer,
        we note that the model employing \gls{aclarc} is able to successfully reduce the relevance of exactly the target artifact,
        consisting mainly of three distinct puzzle piece shapes in the upper right as well as the bottom left of the image,
        while gaining relevance on the panda's head.
        The baseline model,
        in contrast,
        slightly reduces the relevance of the upper-right puzzle piece (although not as significantly as the unlearning model does),
        however,
        it barely has an effect on the lower-left puzzle pieces.
        
        For the ``border'' artifact in the ``mountain bike'' class,
        we find a similar behavior,
        with the unlearned model precisely reducing the relevance of the ``border'',
        while simultaneously putting more emphasis on the desired features of the mountain bike and its driver.
        However,
        here we additionally observe another interesting effect:
        The unlearned model is relatively stable in terms of which features receive more or less relevance over the course of fine-tuning,
        instead only varying in \textit{intensity},
        not \textit{locality},
        pointing to a goal-oriented behavior.
        The same is not true for the baseline model, 
        on the other hand,
        which seems to vary w.r.t. both.
        
        While this observation with regards to training stability is also confirmed for the ``ocarina'' class,
        neither the \gls{aclarc} model nor the baseline manage to correctly decrease relevance for the full ``colored pattern'' artifact. 
        It seems that the computed \gls{cav} representation for that artifact may not sufficiently capture the artifact direction in this instance.
        This could be either due to the high variability in terms of how this artifact appears for different samples,
        or because of the examples offered for computing the \gls{cav} vector not describing the target \gls{ch} precisely enough.
        
        For the ``mannequin head'' concept,
        however,
        the correct concept seems to be unlearned by the \gls{aclarc} model, 
        and with high stability.
        On a first glance, 
        the baseline model seems to reduce relevance of similar features as the unlearned model does.
        But,
        when inspecting this more closely,
        we find that the \gls{aclarc} model actually reduces the relevance of the ``mannequin head'' with higher precision and more completely -- and simultaneously loses less relevance on actually desirable features,
        \ie,
        the lower part of the blue stole.
        
        Although there are cases where the unlearning in featurespace via \gls{aclarc} is not successful -- for instance due to the computed \gls{cav} not representing the correct concept --,
        generally,
        it performs extremely well,
        gaining significant invariance against a target concept.
        Moreover,
        the method performs in an extremely stable manner,
        showing improvements in comparison to a baseline model both quantitatively and qualitatively.
        We were further able to confirm our findings from Section \ref{sec:experiments:featurespray} again,
        demonstrating a connection between artifact complexity and the layer at which it can be unlearned with the best results.
        
        However,
        the application of \gls{aclarc} still requires tedious and time-consuming fine-tuning.
        In contrast,
        the second proposed method for concept removal -- \gls{pclarc} -- is far more efficient in that respect. 
        Keep in mind, 
        though,
        that -- in contrast to \gls{aclarc} -- \gls{pclarc} does not perform true \textit{unlearning} in that sense,
        since it does not allow the network an opportunity to adapt its weights,
        and instead rather \textit{suppresses} artifacts.
        Due to its promising properties with regards to efficiency,
        the following experiments will be dedicated to evaluating the \gls{pclarc} method -- and whether it can keep up with \gls{aclarc} in performance.

    \FloatBarrier
    \subsection{Unlearning Concepts with \glsdesc{pclarc}}
    \label{sec:experiments:unlearning:pclarc}
         After the identification several \gls{ch} type artifacts used by %
         models trained on the ILSVRC2012 dataset (see Sections \ref{sec:experiments:inputspray} and \ref{sec:experiments:featurespray}), 
         we have successfully demonstrated the removal of their influence on the model in the previous paragraphs, using \gls{aclarc}. 
         However, as \gls{aclarc} requires the model to be fine tuned, 
         it is not very efficient and might even become tedious in an iterative artifact identification and removal process. 
         The \gls{pclarc}-method proposed in \ref{sec:methods:clarc}, 
         on the other hand, 
         does not require any further training after the modeling of the artifact, 
         but conversely does not allow the model to adapt its weights and strictly \textit{unlearn} -- as \gls{aclarc} does. 
         Instead, 
         it acts as a filter and removes a concept's contribution to the output. 
         Whether the concept suppression of \gls{pclarc} is successful and comparable to \gls{aclarc} is evaluated experimentally in the following paragraphs. 
         
         First, we measure the performance of \gls{pclarc} in a toy setting on ColoredMNIST, 
         before proceeding to the more complex ILSVRC2012 domain. 
         Finally, we  touch upon the subjects of fairness and reliability in machine learning
         by showing that \gls{pclarc} is able to increase the robustness in the prediction of biased real-world datasets, 
         \ie, the ISIC 2019 dataset in a skin lesion classification setting,
         and the Adience face classification dataset with a \gls{dnn} trained to predict biological gender. 
        
        \paragraph{\glsdesc{pclarc} on Colored MNIST}
            \FloatBarrier
            To assess the validity of the proposed \gls{pclarc} method, we first apply the method in a toy setting with relatively simple (\gls{ch} type) concepts in the dataset.
            More concisely,
            as described in the eariler Section~\ref{sec:experiments:inputspray},
            we add color-based \gls{ch} artifacts to the MNIST dataset~\cite{lecun1998mnist,lecun1998gradient} by distinctly changing the tint of 20\% of the samples per class.
            While simple,
            the resulting concept is complex enough as to not have a pixel-wisely localizable representation in input space.
            We train a simple convolutional network as described in~\ref{sec:appendix:networks}.

\begin{figure}[ht!]
                    \centering
                    \includegraphics[width=.6\linewidth]{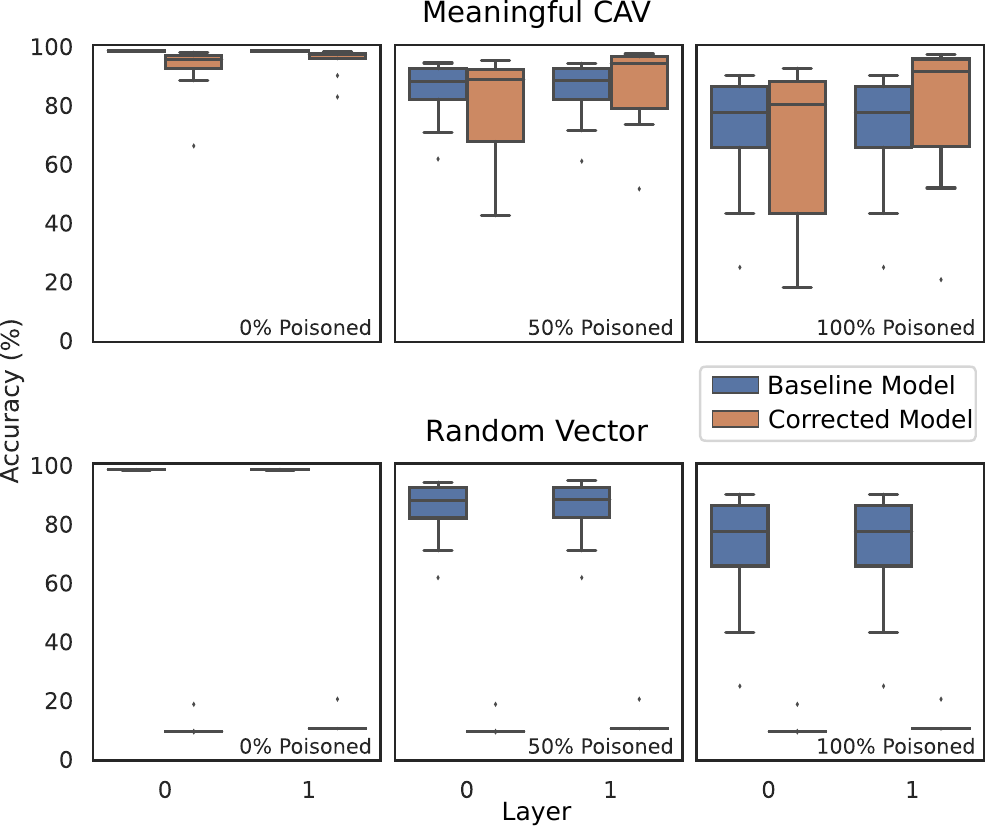}
                    \caption{ %
                        \gls{pclarc} on Colored MNIST with 0 \textit{(left)}, 50 \textit{(middle)} and 100 \textit{(right)} percent validation set poisoning. For the evaluation, the \gls{ch} Artifact is added in the input space.
                        The accuracy of a \textit{baseline} model \textit{(blue)} and an \textit{corrected} model employing \gls{pclarc} \textit{(orange)} on these datasets is compared for \glspl{cav} obtained after the input layer ($0^\text{th}$ layer) and the first convolutional layer ($1^\text{st}$ layer). Measurements are taken from the separate unlearning of all \gls{ch} artifacts in the Colored MNIST dataset. In the \textit{(top)} row, the corrected model uses meaningful \glspl{cav} that are computed from two distinct sets of data samples, as described in Section \ref{sec:methods:clarc}. In contrast, a random vector is utilized instead for the \textit{(bottom)} row. While the meaningful \gls{cav} leads to an improvement of the corrected model over the baseline for the poisoned datasets, the random vector has an extremely detrimental effect on model performance in every case. 
                    }
                    \label{fig:capcmnist_input}
                \end{figure}

            For one color concept and intermediate layer $l$ at a time,
            we ``unlearn'' the target concept without re-training by using \gls{pclarc},
            and evaluate the success of this unlearning procedure using an altered (or poisoned) test set:
            here, 
            the target concept color is applied to a certain percentage of samples from the (whole) test set.
            We then evaluate and compare the performance of the original model to the performance of the model desensitized to the color concept via \gls{pclarc} on this poisoned test data,
            as shown in Figure~\ref{fig:capcmnist_input} \textit{(top)}.
            The accuracy \textit{(y-axis)} of the original model \textit{blue} and the corrected model \textit{orange} is compared for the poison rates $0\%$ (uncolored MNIST),
            $50\%$,
            and $100\%$ \textit{(left to right)},
            averaged over all ten classes.
            This comparison is visualized for the input layer and the first convolutional layer of the feature extractor \textit{(x-axis)}.
            
            With increasing dataset poisoning,
            the model to which \gls{pclarc} is applied outperforms the baseline model.
            However,
            the accuracy of both models on average decreases slightly with higher poison rates,
            showing that while \gls{pclarc} makes the model more robust against the \gls{ch}  artifact specifically,
            the model is not completely unaffected otherwise. 
            Note however,
            that since the \gls{cav} is only computed from samples within one class,
            due to the class-specific properties of \gls{ch} artifacts,
            this evaluation may suffer from generalization issues of that \gls{cav} vector,
            when applied to other classes,
            explaining the high variance of the performance after applying \gls{pclarc}.
            As a sanity check,
            we further perform the same evaluation using randomly generated \glspl{cav},
            as shown in Figure~\ref{fig:capcmnist_input} \textit{(bottom)}.
            As expected,
            the model to which \gls{pclarc} was applied does not outperform the baseline model in this instance, and instead only achieves a considerably lower accuracy due to the arbitrary and not data-specific projection of the features.
            In combination with Figure \ref{fig:capcmnist_input} \textit{(top)},
            this shows not only that the computed \glspl{cav} describe the targeted color concept in a meaningful manner,
            but also that the proposed \gls{pclarc} method is able to exploit the \gls{cav} representation successfully in order to make a model more robust against the target concept.

            The above method of evaluation,
            however,
            again requires the addition of concepts in input space (since the colors are introduced to the test samples in input space) and may thus not be suitable for arbitrary (especially more complex) concepts.
            Especially ``naturally occurring'' artifacts known (and in this paper discovered) to appear in various popular datasets,
            \eg, 
            \gls{ch} artifacts like the mannequin head in ILSVRC2012~\citep{russakovsky2015imagenet},
            colored band-aids in ISIC~2019~\citep{tschandl2018ham, noel2018skin, combalia2019derm},
            or shirt collars in Adience \citep{hassner2014age} do often not have a singular, pixel-wise representation in input space,
            and,
            as such,
            the performance of \gls{pclarc} on these artifacts would be difficult to assert using the above method of poisoning data in input space.
            Thus, we propose the following alternative:
            as previously established,
            \glspl{cav} offer a representation of a concept in feature space.
            Instead of altering test samples in input space,
            we can thus poison the test data by adding the \gls{cav} corresponding to a target concept to latent activations of a certain percentage samples at layer $l$ during inference,
            and again compare the predictions of the model before and after applying \gls{pclarc}.
            As such,
            this evaluation is not restricted to the input space.
            Its validity is,
            however,
            dependent on whether the obtained \gls{cav} actually denotes the correct concept.
            Therefore,
            we also aim to validate whether the \gls{cav} correctly describes the targeted concept:
            To this end, we discard all network layers after layer $l$,
            and model the network output with the \gls{cav} classifier receiveing its inputs from layer $l$.
            We thus obtain a network that classifies for a given input sample,
            whether it contains the concept described by the \gls{cav}, or not.
            In the following,
            this network is called \textit{\gls{cav}-predictor}.
            After applying \gls{lrp} to this \gls{cav} classification network,
            the resulting relevance maps can be evaluated in terms of whether they correspond to the expected target concept. 
            
\begin{figure}[ht!]
                    \centering
                    \includegraphics[width=.99\linewidth]{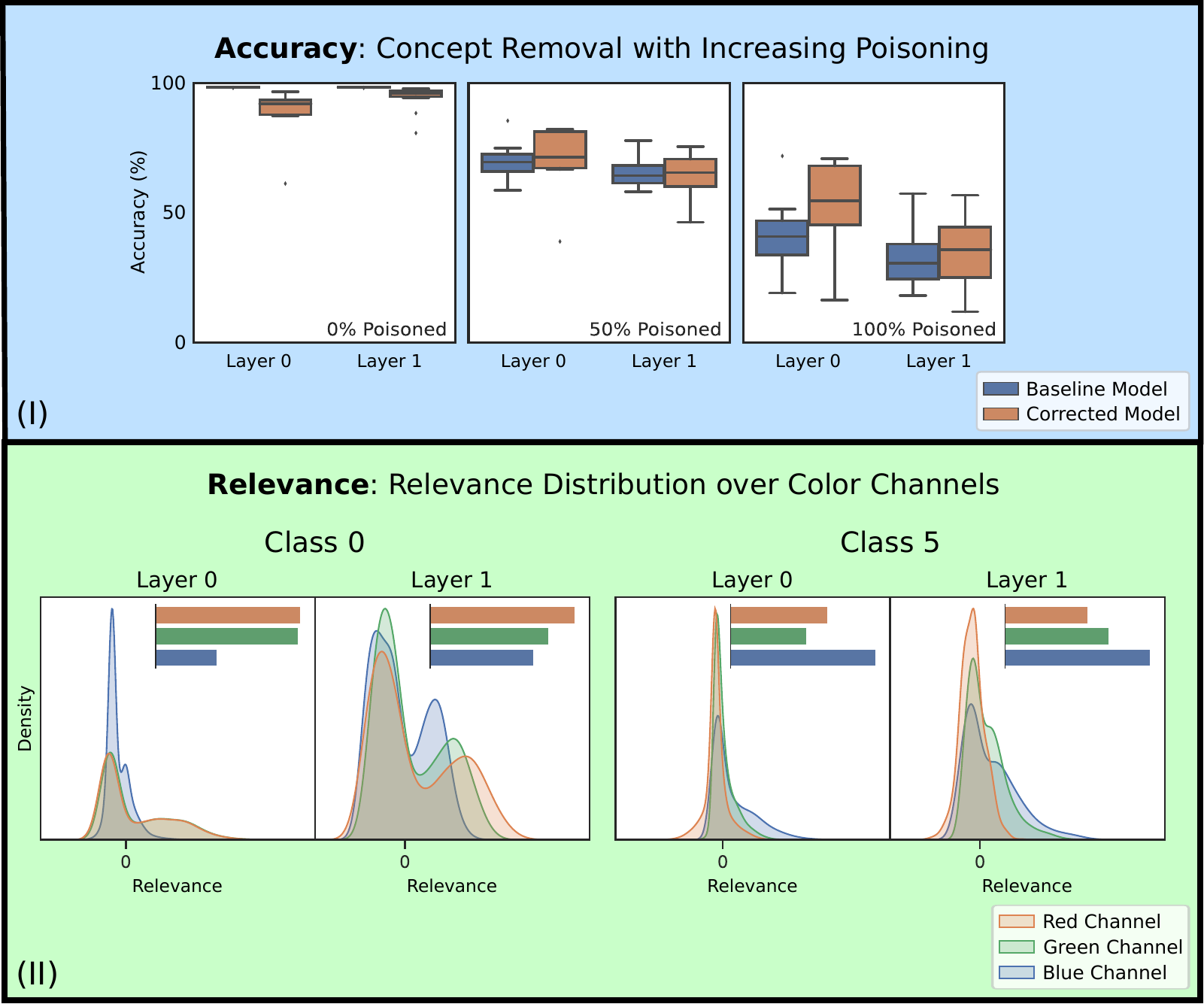}
                    \caption{ %
                        \gls{pclarc} on Colored MNIST with the artifact added in feature space as a \gls{cav} during evaluation (as opposed to Figure \ref{fig:capcmnist_input}, where the artifact is added in input space).  
                        \textit{(I):} Evaluation of performance with 0 \textit{(left)}, 50 \textit{(middle)} and 100 \textit{(right)} percent validation set poisoning. The accuracy of a \textit{baseline} model \textit{(blue)} and a \textit{corrected} model using \gls{pclarc} is compared over all classes for layers 0 and 1.
                        The corrected model outperforms the baseline model.
                        \textit{(II):} Validation of the concept that is described by the computed \gls{cav}. 
                        The distribution of \gls{cav}-predictor (\gls{lrp}-) relevance over color channels is shown for the classes 0 \textit{(left}) and 5 \textit{(right)}, with the introduced \gls{ch} concepts \textcolor{cmnist_0}{blue} and \textcolor{cmnist_5}{orange}.
                        The bar plots above show the sum of (unsigned) relevances across color channels.
                        The \gls{cav}-predictor assigns most relevance to the color channels that differentiate the poisoned samples from the from the original samples (\eg, red and green for the \textcolor{cmnist_0}{blue} artifact).
                    }
                    \label{fig:capcmnist_feat}
                \end{figure}

            With the second proposed method of evaluation shown in Figure~\ref{fig:capcmnist_feat}~\textit{(I)},
            both models decrease in accuracy relatively, especially for higher rates of dataset poisoning and in comparison to Figure~\ref{fig:capcmnist_input}.
            However, the \gls{pclarc}-corrected model significantly outperforms the baseline on the poisoned validation dataset, and more so when the artifact has been modeled after latent feature representations.
            Furthermore,
            the \glspl{cav} seem to describe their respective color artifact with high precision:
            In Figure \ref{fig:capcmnist_feat}~\textit{(II)},
            the distribution of \gls{lrp}-relevances for the \gls{cav}-predictor is visualized across the three color channels,
            for classes ``0'' \textit{(left)}
            and ``5'' \textit{(right)},
            and,
            respectively,
            colors \textcolor{cmnist_0}{blue}
            and \textcolor{cmnist_5}{orange}.
            Higher relevance is mostly attributed to the color channels that describe the target color concept.
            \Eg,
            for the ``blue'' artifact,
            high relevance is attributed equally to the green and red channels,
            and less to the blue channel:
            Due to the additive rgb color system,
            red and green are the altered channels when introducing a blue artifact.
            In addition,
            Figure~\ref{fig:capcmnist_feat} \textit{(II)} shows the absolute amount of relevance attributed to each color channel,
            confirming that the \gls{cav} indeed describes the target \gls{ch}.
            However,
            as indicated in both parts of Figure~\ref{fig:capcmnist_feat},
            the disentaglement of benign and artifactual features
            seems to work better for layer 0 than for layer 1,
            implying that the \gls{cav} encodes the coloring more precisely there.
            Apparently the coloring is in fact a relatively simple (\ie static \wrt its embedding into the input dimensions) \gls{ch},
            that is still most accurately represented in input space.

        \paragraph{\glsdesc{pclarc} on ImageNet}
        
        With the above toy example showing promising results, 
        we further apply and evaluate \gls{pclarc} in the more complex setting of ILSVRC2012,
        where various \gls{ch}-type artifacts were identified using \gls{spray}, 
        as described in Sections \ref{sec:experiments:inputspray} to \ref{sec:experiments:featurespray}.
        
        Equivalently to the corresponding experiments with \gls{aclarc} in Section \ref{sec:experiments:unlearning:aclarc}, 
        we use the VGG-16 model with the pretrained weights obtained from the Pytorch model zoo. 
        \gls{pclarc} is performed at layers 0, 4, and 10 of the model's convolutional feature extractor in separate experiments. 
        We evaluate on a subset of 100 (randomly chosen) ILSVRC2012 classes that include the class where a \gls{ch} occurs in the data (called ``target class'' in the following).
        Again we compare a \textit{corrected} model that employs \gls{pclarc} to a \textit{baseline} model that does not. 
        For this purpose, we use an unpoisoned and a poisoned validation dataset, 
        with the latter being augmented by adding the \gls{cav} that encodes the target \gls{ch} to the activations of all samples at the respective layer (100\% poisoning). 
        To assert how well \gls{pclarc} suppresses the target \gls{ch} concept, 
        we again employ the previously established twofold evaluation method that does not rely on the introduction of artifacts in input space,
        combining a quantitative comparison between the two models' outputs with a qualitative analysis of the difference in attributed relevances.
        The \gls{ch} artifacts that are inspected more closely were identified using \gls{spray} and range from simple artifacts with static placement in pixel space (\eg, laptop - ``lid'') to relatively complex conceptual and non-static concepts (\eg, swimming trunks - ``upper body'').  
        
        Since dataset poisoning for the purpose of evaluation is achieved by adding the computed \gls{cav} to activations at the respective intermediate layer, 
        it is not sufficient to show that \gls{pclarc} successfully counteracts this, 
        since the same \gls{cav} is used in its projection step. 
        Rather, 
        we first need to establish that the \gls{cav} actually encodes a meaningful feature of the target class.
        Furthermore, 
        to be valid,
        \gls{pclarc} should be concept-specific,
        and thus optimally not have any effect on the network's inference for samples that do not contain the target artifact.
        
\begin{figure}[ht!]
                \centering
                \includegraphics[width=\linewidth]{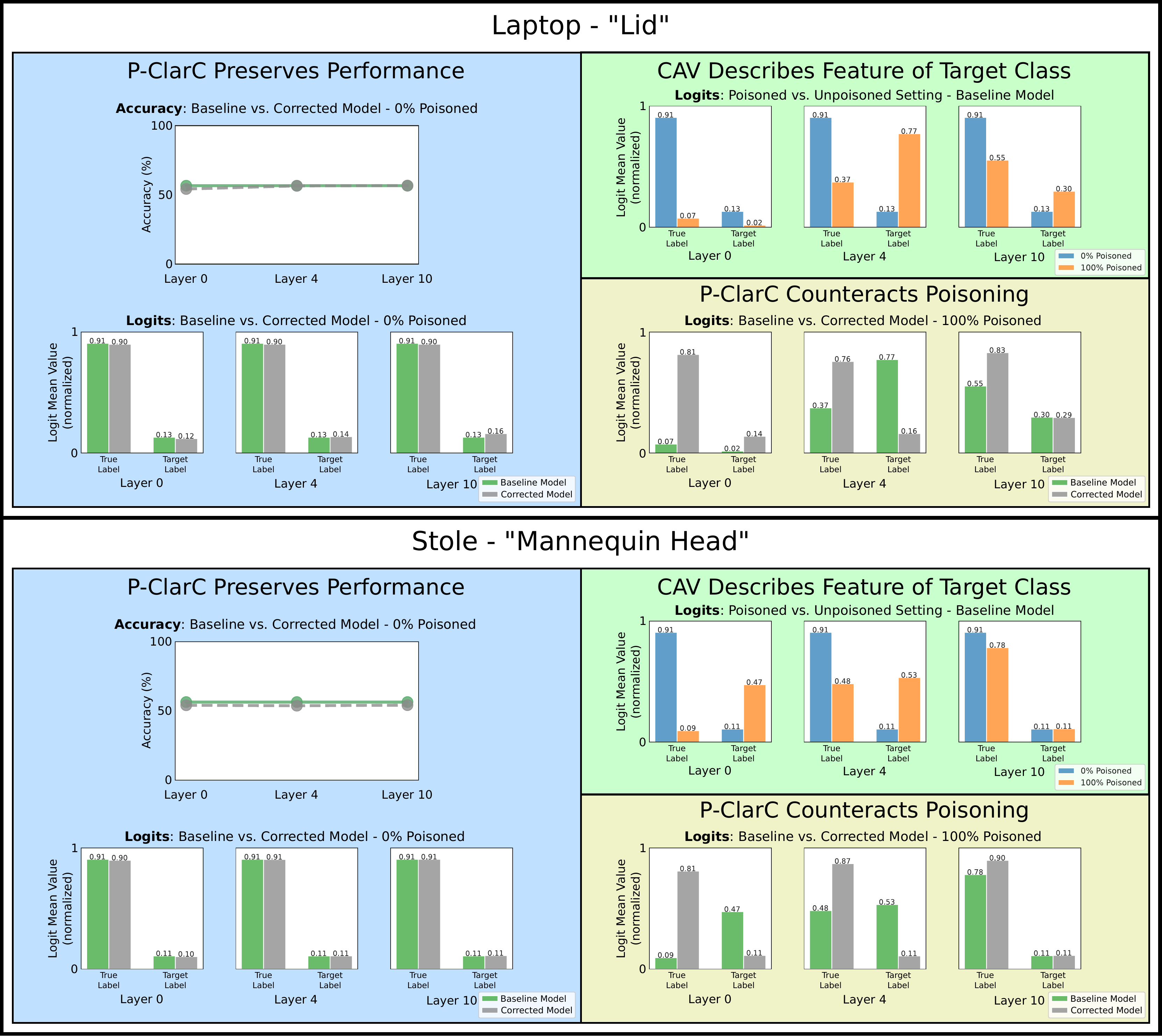}
                \caption{ 
                    Quantitative Evaluation of \gls{pclarc}. As examples for a simple \gls{ch} and a complex \gls{ch}, laptop - ``lid'' \textit{(top panel)} and stole - ``mannequin head'' \textit{(bottom panel)} are shown, respectively. The mean logit values in this figure were normalized sample-wise by dividing by the largest absolute logit value and thus constrained to the interval $[-1, 1]$ for a relative evalation.
                    Results are depicted for layers 0, 4, and 10 of the convolutional feature extractor of the model.
                    The evaluation is threefold:
                    1. Performance is preserved when applying \gls{pclarc} to unpoisoned data (\textit{(left)} of each panel), as both the accuracy nor the logits of the true and target class barely vary between baseline and corrected model. 
                    2. When using the computed \gls{cav} to poison activations additively, the ratio between the true class target class logits diminishes (\textit{(top right)} of each panel), indicating that the feature encoded by the computed \gls{cav} is specific to the target class.
                    3. \gls{pclarc} successfully removes a target concept, encoded by the computed \gls{cav}, since it moves the logits closer to the ground truth, increasing values for the true label, while reducing them for the target label (\textit{(bottom right)} of each panel).
                }
                \label{fig:capimage}
            \end{figure}

        The results of a quantitative analysis of these three properties is shown in Figure \ref{fig:capimage} for the classes ``laptop'' and ``stole'',
        with the \glspl{ch} ``lid'' and ``mannequin head'',
        as examples for a relatively simple and a more complex \gls{ch},
        respectively. 
        
        Note that in this figure, 
        class-wise (normalized) logits are visualized as opposed to the final softmax probabilities,
        since slight changes may not be easily registered in the latter,
        due to the high number (1000) of classes in the ILSVRC2012 dataset and thus the model's output.
        However, 
        as the model is originally trained to optimize softmax probabilities, 
        it is sufficient to only compare the relative relationship between classes outputs due to the shift invariance of the softmax function.
        As shown on the \textit{(left)} side of Figure~\ref{fig:capimage}, 
        \gls{pclarc} preserves a model's performance 
        if applied to unpoisoned data. 
        Note that since the 100 validation classes also contain the target class, 
        a very slight change in performance can be found for,
        \eg,
        class ``laptop'' at Layer 0 or the class ``stole''.
        However, 
        the mean logit values never vary by more than $0.03$ between the baseline and corrected model,
        with the ratio of true label logits and target label logits barely changing. 
        
\begin{figure}[ht!]
                \centering
                \includegraphics[width=.9\linewidth]{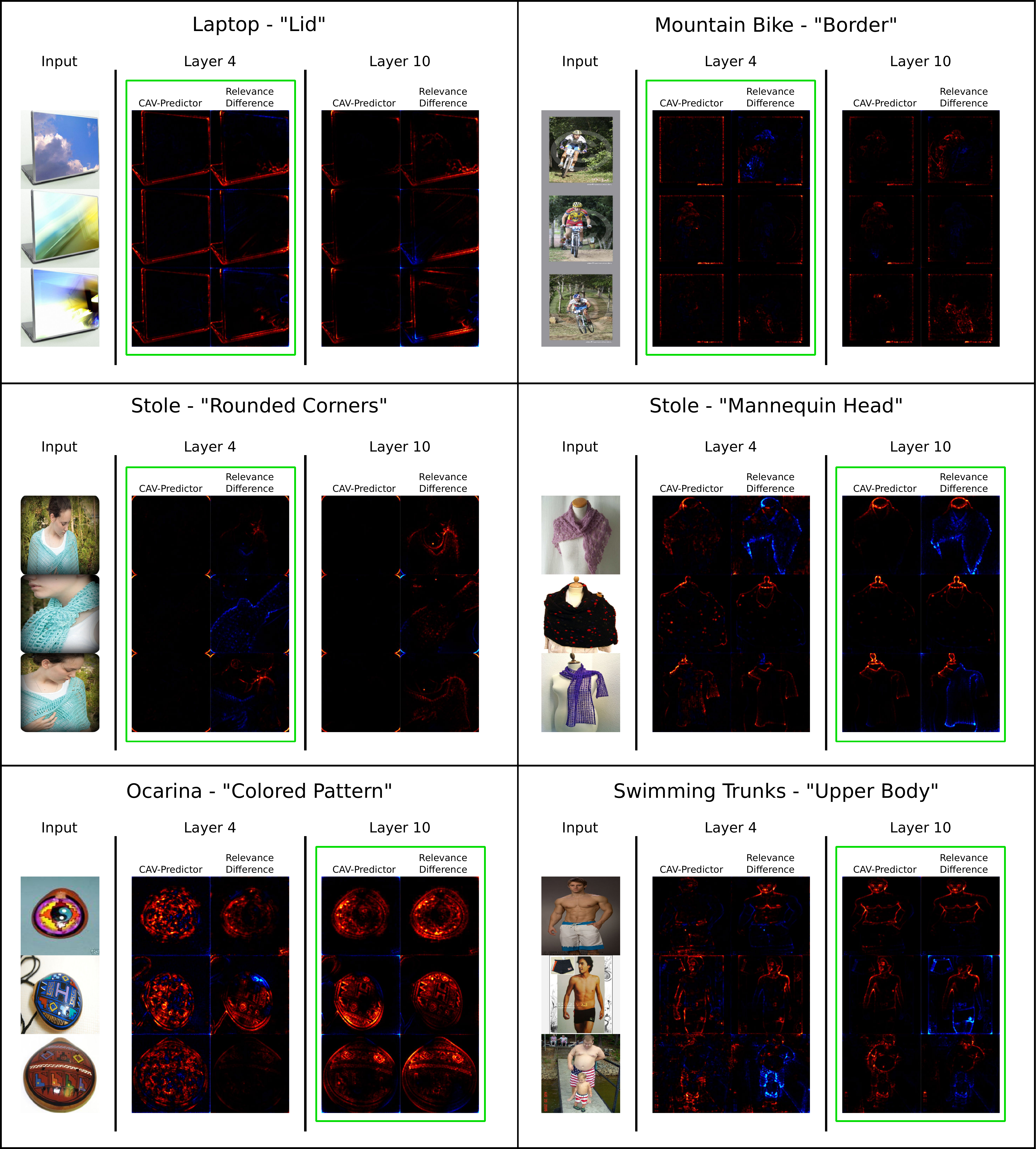}
                \caption{ %
                    Effects of \gls{pclarc} on ILSVRC2012. In every panel, \gls{pclarc} was applied after layers 4 and 10. For each of these, the \gls{lrp} relevances of the \gls{cav}-predictor and the relevance difference between the \textit{baseline} and the \textit{corrected} model is visualized. In the relevance difference images, the corrected model focused less on the areas highlighted in \textit{red} compared to the baseline model, but more on the parts highlighted \textit{blue}. While the first three \glspl{ch} (``lid'', ``border'', ``rounded corners'') occupy the same pixels between samples, the last three \glspl{ch} (``mannequin head'', ``colored pattern'', ``upper body'') consist of more complex features. In line with this complexity, \gls{pclarc} seems to perform better on earlier layers of the feature extractor for the first group, with the heatmaps corresponding more to the target concept, as indicated by the \textit{green} border. The opposite seems to be the case for the second group, concurring with the separability scores in Figure \ref{fig:featurespray}.
                }
                \label{fig:capimage_heatmap}
            \end{figure}

        Since the projection step of \gls{pclarc} relies on the computed \gls{cav} precisely representing the targeted \gls{ch} concept,
        we next assert whether the \gls{cav} is meaningful \wrt the target class,
        \ie,
        whether it describes a feature specific to the target class.
        We do this by observing how the mean logit values of the true and target class labels change when the model's inference process is poisoned by adding the computed \gls{cav} to the activations of the respective layer,
        thereby shifting these activations in the \gls{ch} direction -- as it is described by the \gls{cav}.
        The corresponding results are demonstrated in the top right of each panel of Figure \ref{fig:capimage}.
        Here,
        we note a significant decrease in the mean logit values of the true class labels when poisoning the validation data. 
        At the same time,  
        the mean logit values of the target labels mostly increase,
        \eg,
        class for ``stole'' in layer 4,
        where the true label logit mean value diminishes from $0.91$ to $0.48$ due to poisoning,
        while for the target label it increases from $0.11$ to $0.53$ at the same time. 
        An exception is the class ``laptop'' at layer 0,
        where values decrease for both (sets of) classes.
        However, 
        the \textit{ratio} between true label and target label logits always changes in favor of the target label with poisoning.
        We thus deduce that since adding the computed \gls{cav} to the activations at the respective layer relatively increases the model's confidence of the target class over the true class, 
        the \gls{cav} encodes for a feature that is specific to the target class. 
        Note that this does not necessarily imply that the \gls{cav} describes the exact target \gls{ch} concept, 
        which is an observation that we investigate further in Figure \ref{fig:capimage_heatmap} a few paragraphs further.
        
        In the poisoned setting,
        the baseline model consistently assigns a larger logit value to the target class than to the true class, in contrast to the unpoisoned setting.
        Observed exceptions to this rule are layer 0 for class ``laptop'' and layer 10 for class ``stole'' (\cf the bottom right parts of the panels in Figure~\ref{fig:capimage}).
        This can be explained by the relative complexity of the respective artifacts and their (attempted) point of encoding in the network.
        Artifacts best expressed statically in pixel space (here, the laptop's lid) are more readily encoded by a \gls{cav} trained here, compared to later layers, where the model has developed invariances against pixel-specific encodings.
        Conversely, more complex and semantic concepts such as the mannequin head,
        which as a feature appear in multiple locations and poses over the dataset
        are more readily encoded in invariant latend representations later in the model.

        When employing \gls{pclarc},
        the model manages to correct this skewed distribution successfully,
        and assigns larger logit values to the true class than to the target class.
        \eg,
        for layer 4 of class ``stole'', 
        the baseline model infers a normalized logit mean value of $0.48$ for the true class, but $0.53$ for the target class.
        The corrected model,
        however,
        shifts this distribution in favor of the true class by projecting the activations beyond the \gls{cav}-predictor's hyperplane,
        assigning a normalized logit mean value of $0.87$ to the true class and $0.11$ to the target class.
        Note that in this example,
        the previous \textit{unpoisoned} mean logit values (\textit{(top right)} of each panel, \textit{blue}) are almost perfectly restored.
        Thereby, 
        it successfully counteracts the introduced poisoning.

        \gls{pclarc} projects samples beyond the hyperplane separating samples within a class that contain a target \gls{ch} concept and samples that do not. 
        Thus, 
        its performance is entirely dependent on how well the learned \gls{cav}, 
        \ie, 
        the vector orthogonal to that hyperplane, 
        describes the target \gls{ch}. 
        The quantitative analysis of Figure \ref{fig:capimage},
        however, 
        is only able to assert that the \gls{cav} describes \textit{some} feature specific to the target class of which the influence on the model's inference can be increased by adding it to the activation in feature space (and consecutively decreased again applying \gls{pclarc}). 
        Up to this point, 
        however, 
        we have not yet shown that the computed \gls{cav} describes \textit{exactly} the target \gls{ch} feature
        or that \gls{pclarc} is able to successfully remove a \gls{ch} concept that is not added artificially, 
        but naturally occurs in the data.
        
        For these two purposes,
        relevance maps computed via \gls{lrp} are shown in Figure~\ref{fig:capimage_heatmap}, 
        with each panel dedicated to one specific \gls{ch} concept. 
        These \glspl{ch} range (in the order of \textit{left} to \textit{right}, and \textit{top} to \textit{bottom}) from simple artifacts that are present in roughly the same pixels of each affected sample (e.g., laptop - ``lid'') to far more complex features (\eg, swimming trunks - ``upper body'') that present differently in each sample and can thus not be described in input space in a uniform manner. 
        For three example images of the each \glspl{ch}, 
        and for the models's intermediate layers 4 and 10, 
        (1) relevance maps of the \gls{cav}-predictor \textit{(left)} and (2) the difference in relevance attribution between the baseline model and the \gls{pclarc}-employing corrected model are shown.

        The former (1) visualizes which features are important to the decision hyperplane of the linear \gls{cav}-predictor in classifying wheter a sample contains a \gls{ch} or not, 
        with \textit{red} highlighting positive relevance, and \textit{blue} negative relevance, 
        and thereby offers an estimation of how well the computed \gls{cav} encodes the correct concept.
        The latter (2), 
        on the other hand, 
        shows how the importance of features for a certain decision changes when employing \gls{pclarc}, 
        with features that are more relevant to the after applying \gls{pclarc} in \textit{blue}, 
        and features whose relevance is reduced in \textit{red}, 
        and thus indicates how successfully the target concept's influence on the model's prediction is removed by \gls{pclarc}.
        
        For all \gls{ch} examples, 
        the \gls{cav} seems to correctly encode the targeted artifact, 
        as the correct features are used to identify them as containing the \gls{ch}. 
        However, 
        there are notable variations in the precision of the \gls{cav}-predictor on the correct features between \glspl{ch} and -- for the same \gls{ch} -- between layers. 
        \Eg, 
        for layer 4 of the laptop - ``lid'' artifact, 
        the outline of a laptop backside, 
        digitally rendered from a specific angle by the image creator,
        is clearly visible.
        However, 
        for layer 10, 
        only the corners of the same outline seem to be relevant, 
        indicating that the artifact is encoded by the \gls{cav} less precise and complete.
        A similar trend can be observed for other relatively simple \glspl{ch},
        \ie, 
        mountain bike - ``border''. 
        For stole - ``rounded corners'' the \gls{cav} seems to be on point for both layers.
        In contrast, 
        the \gls{cav}-predictor for the more complex stole - ``mannequin head'',
        seems to focus on the mannequin head artifact as well as the correct class features of the stole itself in layer 4,
        however, 
        in layer 10,
        it seems to single out the mannequin head artifact almost exclusively.
        A similar effect occurs with swimming trunks - ``upper body'', 
        where the layer 4 heatmaps are relatively diffuse, 
        while for layer 10 only the human upper bodies are assigned a large positive relevance.
        
        In a similar manner, 
        the relevance difference heatmaps show that the artifact is less impactful on the model's decision-making after applying \gls{pclarc}. 
        Again, 
        the success of this artifact removal varies with the specific \gls{ch} and layer, 
        and this variation seems to behave in the same way as for the \gls{cav}-predictor heatmaps, as described above, 
        although some differences exist.
        \Eg,
        for layer 4 of the laptop - ``lid'' artifact, 
        where the \gls{cav}-predictor seems to be most precisely learned,
        the corrected model assigns far less relevance to the outline of the laptop's lid.
        In contrast, 
        at layer 10,
        suddenly also parts of the image imprinted on the lid are removed.
        Mountain bike - ``border'' behaves in a similar manner,
        however,
        for stole - ``rounded corners'',
        while both \gls{cav}-predictor and relevance difference heatmaps somewhat coincide at layer 4,
        with the corners being removed correctly,
        at layer 10 mainly the blue stole itself receives less relevance,
        and relevance on the rounded corners actually increases.
        This makes sense, as the rounded corners are an extremely simple artifact -- thereby being removed more successfully in earlier layers, 
        in accordance with our previous findings. 
        However, 
        it also seems that just because the \gls{cav} seems to describe the artifact correctly,
        the unlearning result does not always exactly correspond to that.
        Note, 
        however, 
        that since these heatmaps are normalized \wrt to the largest absolute relevance value, 
        the rounded corners may only be assigned an extremely large relevance,
        and the some smaller relevance value.
        In fact,
        this example further showcases another interesting problem:
        The samples of the class stole that contain the ``rounded corners'' artifact also always contain the same person and the same blue stole. 
        The \gls{ch} is thus ill defined here, 
        since the ``rounded corners'' cannot be described by only using example images,
        making \glspl{cav} apparently not the ideal choice of representation for this specific artifact.
        Since the resulting \gls{cav} would encode both, in a way, 
        this is thus both a simple and a complex \gls{ch},
        with \gls{pclarc} removing the simple part (``rounded corners'') at the earlier layer,
        and the more complex ``blue stole'' at the later layer.
        
        Matching these interpretations, 
        the complex ``mannequin head'', 
        ``colored pattern'', 
        and ``upper body'' artifacts are removed far more successfully at layer 10.
        Note especially the class ``swimming trunks'', 
        where not only the relevance of the upper body decreases, 
        but also relevance on the swimming trunks themselves is increased.
        The same effect is also visible for the ``mannequin head'' artifact.
        
        To summarize,
        there seems to be an intermediate layer where the computed \glspl{cav} not only encode the correct and intended \gls{ch} concept  -- although this layer differs for each respective artifact.
        The \gls{ch} correction is also more precise at the same layer, 
        not only leading to a lessened impact of the targeted artifact on the model's prediction,
        but also often an \textit{increase} of the correct non-\gls{ch} class features.
        In fact, this layer largely coincides with the complexity of the targeted artifact, 
        confirming expectations and our findings from Section~\ref{sec:experiments:featurespray}.
        Although,
        we observe that in comparison to Section~\ref{sec:experiments:featurespray},
        the best performing layers are shifted backwards in the network,
        \eg,
        for the ``lid'' \gls{ch},
        this optimum seems to be at layer 4 instead of layer 0 when applying \gls{pclarc},
        possibly due to exploitation of the model's feature space representation at later layers being more invariant.
        Further taking the results of Figure~\ref{fig:capimage} into account,
        where we showed how \gls{pclarc} not only counteracts poisoning and shifts the prediction towards the true class, 
        but also does not affect performance on unpoisoned data in a significant manner,
        we thus surmise that \gls{pclarc} is an efficient but powerful tool for concept removal.
        Note,
        however,
        that \gls{pclarc} will not lead to an increased generalization performance, 
        since the model never has a chance to adapt its weights for learning other features and thus correct its faulty prediction reasoning.
        Nevertheless, 
        its strengths lie in its ability to offer a fairer estimation of a model's generalization performance, 
        untainted by features that should not contribute to the decision-making.
        
        \gls{pclarc} is able to successfully reduce the impact of \gls{ch} artifacts on a model's prediction, 
        and employing it on ILSVRC2012 is able to demonstrate that fact.
        Although,
        this dataset my not be sufficient for showcasing how powerful \gls{pclarc} can be towards the solution of some pressing problems hindering the application of \gls{ml}-methods in real-world scenarios.
        For this reason, the following paragraphs will offer two examples, 
        where \gls{pclarc} is employed to avoid predictions for the wrong reasons with dangerous consequences, 
        and to increase classification fairness on biased data.

        \FloatBarrier

        \paragraph{Unlearning with \glsdesc{pclarc} on ISIC 2019}
        
            In the previous sections,
            we have confirmed the success of \gls{pclarc} applications on toy examples and more complex settings on real photographic images,
            \ie the ILSVRC2012 dataset.
            
            In this (and the following) section, 
            we will apply \gls{pclarc} to more domain specific datasets in order to solve practically relevant issues.
            Here, we demonstrate that \gls{pclarc} can be used to increase the trustworthiness of models trained for skin lesion classification on the ISIC~2019 dataset.
            As it common practice, 
            we fine-tune a neural network (here a VGG-16 model) pretrained on ILSVRC2012 on the ISIC 2019  \cite{tschandl2018ham, noel2018skin, combalia2019derm} skin lesion classification dataset for 100 epochs, 
            using the weights from the Pytorch model zoo for initialization.
            Due to ISIC 2019 not having a pre-defined labeled test set, 10\% of the original training set were split off instead to evaluate its performance.
            Our model achieves a final test accuracy of 82.15\%.
            
            It is known,
            however, 
            that the ISIC~2019 dataset
            contains several issues and confounders.
            First and foremost,
            a significant data artifact, that only occurs in the largest class, 
            \ie
            colorful band-aids next to the photographed skin alteration.
            Since this artifact is again limited to one class,
            it constitutes a \gls{ch}-type artifact.
            For the purpose of skin lesion classification, 
            aimed to be applied in the medical field to assist medical personnel or allow mobile diagnoses \cite{esteva2017skincancer},
            \glspl{ch} like these can have serious consequences, 
            as they may easily lead to a misclassification,
            affecting the resulting diagnosis,
            and,
            as such,
            the life of a patient.
            Especially, 
            since the affected class, 
            ``melanocytic nevus'', 
            is a benign form of skin alteration,
            possibly leading to fatal false negatives in terms of skin cancer diagnosis.
            
\begin{figure}[ht!]
            \centering
            \includegraphics[width=.65\linewidth]{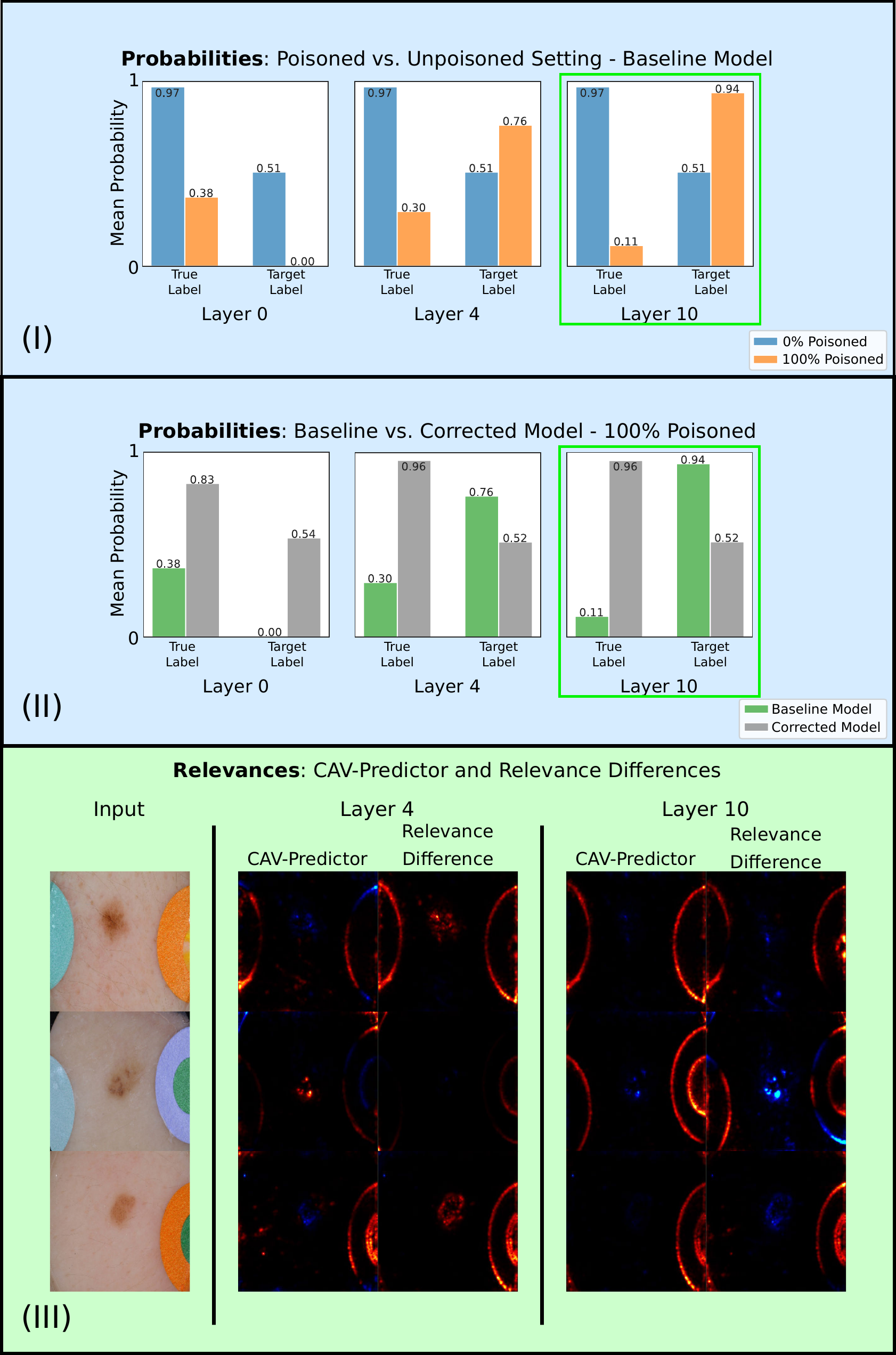}
            \caption{ %
                Employing \gls{pclarc} on ISIC 2019 to suppress the ``colored band-aid'' \gls{ch} within the ``melanocytic nevus'' class. \textit{(I)}: Effect of adding the computed \gls{cav} to activations at the respective layer. The mean softmax probability of the ``melanocytic nevus'' class increases, while true class probability decreases for layers 4 and 10, indicating that the \gls{cav} encodes a feature specific to the ``melanocytic nevus'' class. \textit{(II)}: Success of the concept suppression of \gls{pclarc}. The poisoning of \textit{(I)} can be mitigated using \gls{pclarc}, restoring confidence in the true class. For \textit{(I)} and  \textit{(II)}, the layer where \gls{pclarc} seems to perform best is marked by a green border. \textit{(III)}: Example images and corresponding \gls{cav}-predictor- and \gls{lrp} relevance difference heatmaps for the layers 4 and 10, where the quantifications of \textit{(I)} and \textit{(II)} yielded positive results.
                In \gls{cav} predictor heatmaps, \textit{red} areas indicate high relevance, \ie, highlight features indicative for the \gls{cav} direction.
                In the difference heatmaps, \textit{red} areas were attributed less relevance, and thus used less by the model, after applying \gls{pclarc}. The focus of the \gls{cav}-predictor seems to be relatively diffuse in layer 4, and only partly located on the targeted band-aids, supported by the only partial success of the concept mitigation, where sometimes even desired features are diminished. In contrast, both heatmap types are extremely precise at layer 10, and not only is the relevance of the \gls{ch} reduced, but the melanoma itself becomes more important for the model's decision.
            }
            \label{fig:capimage_isisc2019}
        \end{figure}

            With this in mind,
            we aim to mitigate the effect that the ``colorful band-aids'' \gls{ch} has on the model's prediction by employing \gls{pclarc}. 
            For this purpose, 
            we again compare the model which \gls{pclarc} is applied
            and the original model in terms of predictions and \gls{lrp} relevance maps. Results are shown in Figure~\ref{fig:capimage_isisc2019}.  %
            Here, 
            as opposed to the corresponding evaluations for ILSVRC2012 (Figure~\ref{fig:capimage}) where normalized mean \textit{logit} values were considered (due to the high number of classes), 
            we measure the more stabilized mean \textit{softmax} probabilities, 
            since ISIC 2019 only contains 9 distinct classes, whereas ILSVRC2012 contains 1000. 
            Due to the missing test set labels, 
            the (whole) training set is used for the quantitative evaluations in panels \textit{(I)} and \textit{(II)} of this figure.
            However,
            since the application of \gls{pclarc} does not contain any further training,
            the model never has the opportunity to adapt to the performed alterations in any way,
            \eg by shifting its inference strategy to features which prior to \gls{ch} removal
            had a merely supporting function.

            In Figure~\ref{fig:capimage_isisc2019}~\textit{(I)},
            for layers 0 (\ie, the input layer), 4, and 10,
            the effect of adding the \gls{cav} computed (for later usage during \gls{pclarc}) to the activations at the respective layer is measured. 
            If the \gls{cav} encodes a feature that is specific to the target class ``melanocytic nevus'', 
            one would expect the softmax probability of that class to increase when poisoning the samples in that way, 
            while confidence in the actual true class label would decrease simultaneously.
            Note that due to the true class changing from sample to sample, 
            the sum of (mean) true class and target class probabilities may exceed 1 in this figure.
            For layer 0,
            we observe a decrease both for true and target labels,
            indicating a generally confusing effect of the \gls{cav}-poisoning on the model,
            as could be expected to some degree:
            In input space,
            the encoding of \glspl{ch} via \glspl{cav} may not be feasible, 
            because the data is too complex in its raw form,
            and no invariant representation learned by the model has been applied yet.
            In contrast, 
            for layer 4 and even more so for layer 10,
            the softmax probabilities exhibit precisely the expected effect.
            \Eg, 
            for layer 10,
            they change from $0.97$ to $0.11$ for the true class, 
            but rise from $0.51$ to $0.94$ for class ``melanocytic nevus''.
            As indicated by the \textit{green border},
            this effect is most prominent in layer 10.
            We thus infer that the computed \gls{cav} indeed denotes a concept specific to the target class -- at least for layers 4 and 10.
            
            Building on that assertion, 
            the next step is to validate whether the \gls{pclarc} method is able to counteract said poisoning.
            Figure~\ref{fig:capimage_isisc2019}~\textit{(III)} shows the corresponding results.
            The inference results of the \textit{baseline} model and the model employing \gls{pclarc} are compared in the form of \textit{mean softmax probabilities}.
            With the data poisoned in the same manner as in Figure~\ref{fig:capimage_isisc2019}~\textit{(I)},
            not only should confidence in the target class decrease with a successful removal of an artifact,
            but also confidence of the true class should increase,
            restoring the predicted probabilities of an unpoisoned setting as closely as possible.
            As visible throughout Figure~\ref{fig:capimage_isisc2019}~\textit{(I)} to~\textit{(II)}, 
            this is barely the case for layer 0,
            partly due to the probabilities already decreasing both for the target and the true class because of the poisoning.
            Even so,
            \gls{pclarc} manages to almost restore the original confidences,
            with the true label probability growing from $0.38$ to $0.83$ (unpoisoned $0.97$) and target label probability from $0.00$ to $0.54$ (unpoisoned $0.51$). 
            Although the \gls{cav} at layer 0 is not meaningful, 
            \gls{pclarc} can still mitigate the poisoning,
            showcasing again the need for our two-part quantitative evaluation,
            validating that not only the concept suppression is successful,
            but also that the \gls{cav} encodes a target-class-specific concept.
            For layers 4 and 10,
            \gls{pclarc} restores predictions even more closely to the original values shown in Figure~\ref{fig:capimage_isisc2019}~\textit{(I)},
            increasing confidence in the true class,
            while decreasing confidence for ``melanocytic nevus'' simultaneously.
            \Eg,
            for layer 4,
            the former rises from from $0.30$ to $0.96$ (unpoisoned $0.97$),
            the latter from $0.94$ to $0.52$ (unpoisoned $0.51$).
            In fact, 
            the same result is obtained for layer 10:
            But since the poisoned probabilities deviate more extremely from an evaluation on unpoisoned data,
            which is also why we find an application \gls{pclarc} at layer 10 to be even more successful in counteracting poisoning (see \textit{green border}).
            
            In Figure~\ref{fig:capimage_isisc2019}~\textit{(III)}, 
            we aim to confirm above assertions for layers 4 and 10 by means of three sample images of the class ``melanocytic nevus'' that contain the targeted ``colored band-aid'' \gls{ch}. 
            The results here are obtained from the unperturbed data of the ISIC~2019 dataset,
            as opposed to the artificially poisoned setting of Figure~\ref{fig:capimage_isisc2019}~\textit{(I)} and~\textit{(II)}.
            For each sample and layer,
            a heatmap computed for the 
            \gls{cav} predictor is shown,
            highlighting areas which speak for the presence of the concept described by the \gls{cav} in \textit{red} color, 
            and areas speaking against it in \textit{blue} color.
            Furthermore, 
            to the \textit{right} of the \gls{cav}-predictor heatmaps,
            the difference in relevances between the model to which \gls{pclarc} is applied and the original model is visualized,
            with \textit{red} areas denoting a \textit{decreased} relevance after the application of \gls{pclarc}.
            Conversely, \textit{blue} areas identify features which are \textit{increasingly} used by the model.
            For layer 4, 
            the computed \gls{cav} seems to encode the ``colored band-aid'' concept only relatively diffusely,
            with some portion of the positive relevance being attributed to the nevus (\ie, the desired feature) itself, as can be seen for the middle example.
            Similarly, the heatmap for the \gls{cav} predictor also attributes negative relevance to the
            \gls{ch} features.
            The then following \gls{ch} correction results suffer from similar issues: 
            While relevance is decreased on the ``colored band-aids'' themselves,
            often also the nevus receives less relevance,
            \eg, as observable with the first and second examples.
    
            In contrast,
            the \gls{cav}-predictor heatmaps are far more precisely marking the confounding features in layer 10,
            with not only the \gls{ch} being extremely relevant,
            but the desired features also beeing a seemingly neutral (black color in heatmap) or an even negative indicator (blue color in heatmap) for the presence of the encoded concept.
            The accompanying difference maps show a strong decrease in relevance for the \gls{ch} areas,
            and a simultaneous increase in the relevance of the desired features,
            showing not only that \gls{pclarc} in layer 10 successfully corrects the faulty usage of the ``colored band-aid'' as an important feature for the model to decide for the ``melanocytic nevus'' class, but also further shifts the model's focus to the actually desired features,
            \ie, 
            the nevi themselves.
            
            Since the computed \gls{cav} is not only meaningful \wrt the target class,
            but also exactly describes the targeted \gls{ch} artifact (at least for layers 4 and 10),
            and since \gls{pclarc} is able to unlearn that concept,
            we can thus surmise the -- albeit layer-dependent -- success of the \gls{pclarc} method on the ISIC~2019 skin lesion classification dataset for mitigating the effects of training data containing \gls{ch} artifacts.
            Due to the corrected model using desired features \textit{preferably} to the \gls{ch} features, 
            its trustworthiness increases, 
            reducing the risk of costly misclassifications caused by the \gls{ch}.

        \paragraph{Unlearning with \glsdesc{pclarc} on the Adience dataset of unfiltered faces}
        
        As opposed to the medical setting of ISIC 2019, 
        we now apply \gls{pclarc} to a gender classification task with the Adience dataset~\cite{hassner2014age}.
        This dataset has various known problems,
        \eg,
        a relatively high class imbalance,
        as well as a multitude of biases within the data models tend to quickly overfit on,
        as in part identified in \cite{lapuschkin2017understanding} via \gls{lrp}.
        
        In the gender classification setting,
        one of these bias-concepts is the presence of shirt collars in the class of male faces.
        Samples labelled as ``male'' with a  shirt collar are a common occurence within the dataset,
        and samples labelled as ``female'' wearing a showing a shirt collar are quite rare.
        Thus,
        models trained on the Adience dataset often use use this confounding feature as a \gls{ch} for the class defining the appearance of male faces,
        thereby short-cutting (the learning of) more complex features.
        This is also the case for the VGG-16 model we trained for gender classification.
        Using the pretrained ILSVRC2012 weights provided by Pytorch for initialization,
        the model was trained over 100 epochs on folds 1-4,
        keeping fold 0 for testing.
        The final accuracy achieved by this model was 94.02\%.
        
        However,
        the reliance of this model on \glspl{ch} like the shirt collar concept may lead to unfair predictions, 
        \eg,
        when a woman is predicted as ``male'' due to wearing clothes associated by the model with the class ``male'', \ie here, a shirt collar.
        The impact of this is especially high in real-world applications,
        when stereotypes -- that are apparently present in the available training data -- are propagated into the inference of machine learning solutions.
        Here,
        we thus employ \gls{pclarc}, with the aim of obtaining fairer gender predictions on the Adience dataset \wrt the ``shirt collar'' \gls{ch}.
        
\begin{figure}[ht!]
            \centering
            \includegraphics[width=.92\linewidth]{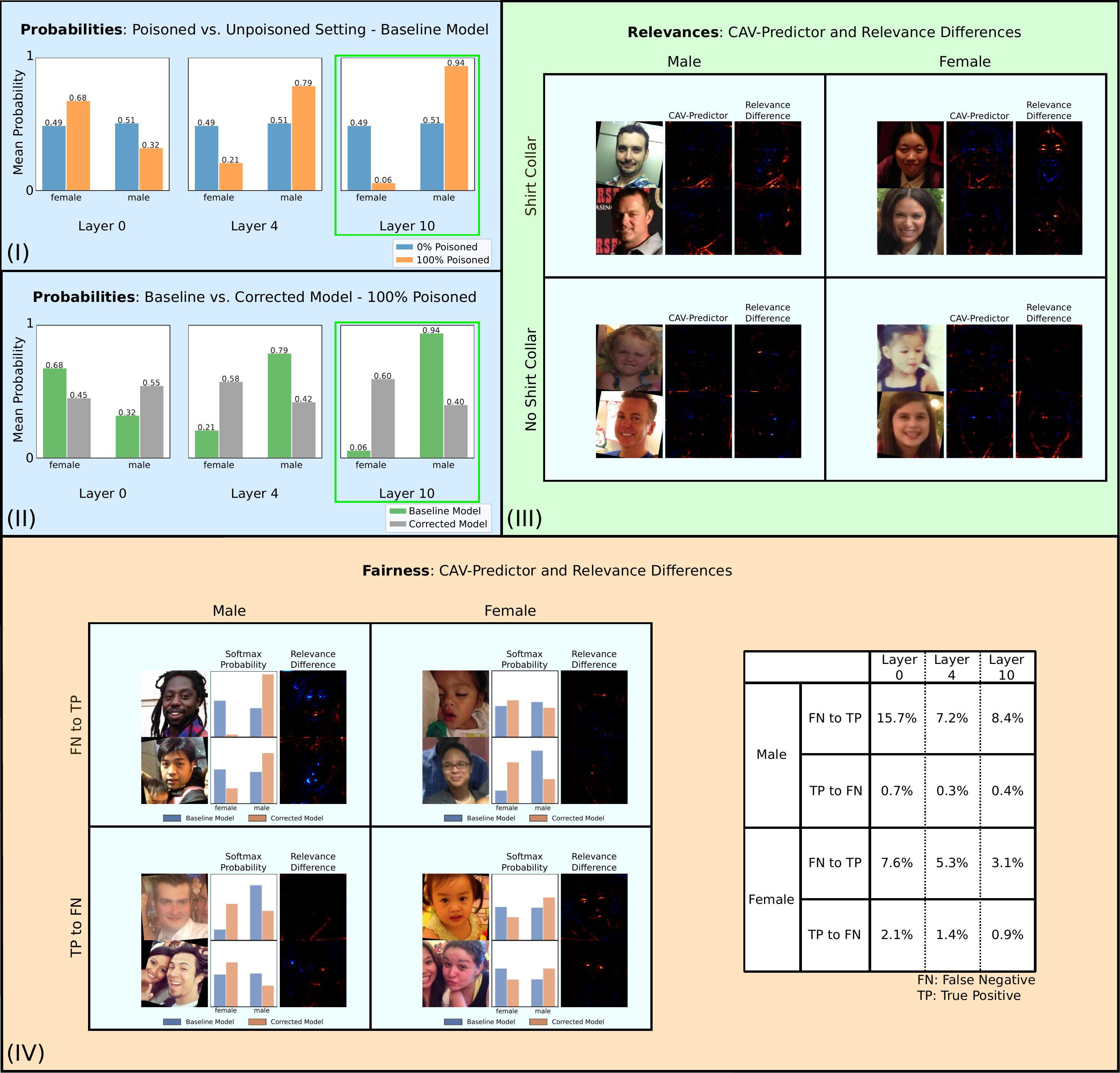}
            \caption{ %
               Application of \gls{pclarc} on the Adience dataset, with the aim to obtain less stereotypical and fairer predictions.
               The target \gls{ch} is the ``shirt collar'' concept used by the model to predict in favor of the class ``male''. 
               \textit{(I)}: By adding the computed \gls{cav} to activations at the respective intermediate layer, the prediction can be affected in such a way that confidence on class ``male'' increases, showing that the \gls{cav} describes a concept specific to ``male''.
               The layer where this works best is marked by a green border.
               \textit{(II)}: Using \gls{pclarc}, poisoning via the computed \gls{ch} is easily mitigated.
               As a result, the softmax probabilities on the class ``female'' increase, while they decrease for ``male''.
               \textit{(III)}: \gls{cav}-predictor and \gls{lrp} relevance difference heatmaps at layer 10 (the best performing layer according to (I) and (II)) for examples of both genders, with and without the target \gls{ch} ``shirt collar'' each.
               The artifact is predicted and suppressed successfully if present in class ``male'', however, in the  class ``female'', this is not always the case.
               \textit{(IV)}: Analysis of transitions between \textit{true positive} and \textit{false negative} predictions when applying \gls{pclarc}.
               Examples for layer 10 of which the predicted class is flipped are shown to the \textit{(left)}, together with softmax probabilities of each sample before and after using \gls{pclarc} and the corresponding change in relevances.
               The table to the \textit{(right)} shows the percentage of original \textit{true positives} that change to \textit{false negatives}, and vice versa.
               Generally, a higher percentage of \textit{false negatives} is corrected than \textit{true positives} are confused. Due to the original model being 94\% accurate,
               however, a larger absolute number of samples are changed from \textit{true positives} to \textit{false negatives}, leading to an overall decrease in accuracy.
            }
            \label{fig:capimage_adience}
        \end{figure}

        Figure \ref{fig:capimage_adience} shows the results of this experiment on the fold 0 test set, 
        comparing the original model to the model employing \gls{pclarc} to suppress the targeted ``shirt collar'' \gls{ch}. 
        To compute the corresponding \glspl{cav}, 
        two hand-selected subsets of the samples representing class ``male'' were used,
        one containing samples with shirt collars, 
        and one without.
        Figure~\ref{fig:capimage_adience}~\textit{(I)} and~\textit{(II)} shows for intermediate layers 0, 4, and 10,
        similar to Figures~\ref{fig:capimage} and~\ref{fig:capimage_adience},
        a quantitative evaluation of the change in mean \textit{softmax} probabilities when using the computed \gls{cav} to poison activations at the respective layer (Figure~\ref{fig:capimage_adience}~\textit{(I)}) and when applying \gls{pclarc} to mitigate that poisoning (Figure~\ref{fig:capimage_adience}~\textit{(II)}).
        This change is measured for both class labels of the dataset,
        \ie,
        ``female'' and ``male'',
        with the latter being the target class.
        That is,
        the class for which the \gls{ch} ``shirt collar'' is used by the model as an indicative feature.
        Similarly to the results obtained for the ISIC~2019 dataset, 
        we find that for layer 0,  
        the computed \gls{cav} does not seem to be able to concisely describe a feature specific to the target class, 
        since poisoning activations with it leads to a decrease the softmax probability of all classes, including class ``male''.
        To reiterate, 
        a meaningful \gls{cav} direction,
        \ie,
        a \gls{cav} that encodes for a feature of the target class,
        would lead to an increase in the model's confidence on that class.
        However,
        this is not the case here
        with scores for class ``male'' dropping from $0.51$ to $0.32$
        probably due to the raw input data
        that has not yet been affected by any learned invariant internal representation of the model,
        being too complex for the \gls{cav} to successfully describe.
        Note that the softmax scores for class ``female'' simultaneously increase in this setting (from $0.49$ to $0.68$).
        That is, however, a byproduct of the confidence decrease for class ``male''
        due to the binary classification task.
        The poisoning counteraction of \gls{pclarc} for layer 0 is comparatively successful  (Figure~\ref{fig:capimage_adience}~\textit{(II)}),
        with the original probabilities from Figure~\ref{fig:capimage_adience}~\textit{(I)} only being within a margin of error of only $0.04$.
        But since the computed \gls{cav} is not fully meaningful,
        the direction removed by \gls{pclarc} cannot correspond to the target concept for layer 0.
        In contrast, 
        for layer 4 and
        even more so
        for layer 10, 
        as indicated by the \textit{green} border,
        the poisoning in Figure~\ref{fig:capimage_adience}~\textit{(I)} yields the expected results,
        increasing the predicted probability of class ``male'' on average for, 
        \eg,
        layer 10,
        from $0.51$ to $0.94$, 
        while decreasing it for class ``female'' from $0.49$ to $0.06$.
        In Figure~\ref{fig:capimage_adience}~\textit{(II)},
        however,
        the removal of \gls{cav}-poisoning seems to overreach for layers 4 and 10:
        The original predicted probabilities of $0.49$ for class ``female'' and $0.51$ for class ``male'' are not exactly restored,
        instead,
        \eg,
        for layer 10 the confidence of class ``female'' rises from $0.06$ to $0.60$ (in stead of $0.49$ for a perfect recovery),
        while it drops from $0.94$ to $0.40$ (instead of $0.51$) for class ``male'', with 11\% discrepancy compared to the original values.
        Keeping in mind that Figure~\ref{fig:capimage_adience}~\textit{(I)} shows that the \gls{cav} is meaningful \wrt the target class,
        we thus infer that either the \glspl{cav} for layers 4 and 10 encode the targeted concept -- and removing it affects the prediction so much because the model strongly relies on that feature,
        or the \gls{cav} encodes \emph{not only} the shirt collar,
        but additionally other (possibly valid) features for class ``male'' that appear alongside shirt collars with a relatively large correlation.
        In any case,
        layer 10 is marked with a \textit{green} border,
        since the concept suppression effect is strongest there.

        For this layer 10, 
        Figure~\ref{fig:capimage_adience}~\textit{(III)} shows samples for both classes ``male'' and ``female'', 
        both with and without the target \gls{ch} ``shirt collar'',
        respectively.
        Each sample is accompanied by two types of \gls{lrp} relevance maps, 
        the first on the left showing which features are important for the \gls{cav}-predictor in \textit{red}, 
        \ie,
        which features indicate the presence of the target concept as it is represented via the computed \gls{cav},
        while features speaking against it are highlighted in \textit{blue} color.
        The second relevance map evaluates features of the respective sample that are used less by the model for its predictions after the application of \gls{pclarc} in \textit{red} color,
        and features that are used more in \textit{blue} color.
        On images of the target class ``male'' that contain the target \gls{ch} \textit{(top left)}, 
        on a first glance 
        positive relevances in the \gls{cav}-predictor heatmap seem to focus on the actual  shirt collar, indicating that the computed \gls{cav} does encode for the target concept.
        In the relevance difference maps, 
        however,
        while the relevance of the shirt collar decreases with an application of \gls{pclarc}
        and that of the facial features (\ie the features desired to be used by the model, naively summarized) increases
        some other features, 
        \eg,
        visible and uncovered ears,
        seem to also be suppressed. 
        In the \gls{cav}-predictor heatmap,
        these are assigned a small positive relevance.
        As found by \cite{lapuschkin2017understanding},
        specifically the visible ears also tend to be learned by models as an indicator for class ``male'' and
        possibly even constitute a \gls{ch}.
        Apparently,
        these features often appear alongside the positive examples for the ``shirt collar'' concept,
        thereby leading to the computed \gls{cav} not only encoding for ``shirt collar'' features,
        but additionally for other -- possibly \gls{ch} features of the class ``male'',
        further confirming our suspicions regarding the large shift in mean softmax probabilities when \gls{pclarc} is applied at layers 4 and 10 in Figure~\ref{fig:capimage_adience}~\textit{(II)}.
        As the Adience dataset is an extremely complex dataset with highly biased data,
        a noisy \gls{cav} encoding is to be expected,
        especially,
        since the precision of the \gls{cav} is highly dependent on the samples chosen for its computation.
        
        In contrast, 
        when the target \gls{ch} is not present (Figure~\ref{fig:capimage_adience}~\textit{(III)} \textit{(bottom left)}), 
        correctly no collar is identified. 
        Although,
        again,
        uncovered ears seem to receive partial positive relevance.
        For the ``female'' class,
        however,
        even though the shirt collar is identified by the \gls{cav}-predictor relevance maps
        ( (Figure~\ref{fig:capimage_adience}~\textit{(III)} \textit{(top right)}); albeit by far not as precisely as for class ``male'' -- ``collar''),
        it does not seem to diminish reliably after applying \gls{pclarc}. 
        Instead,
        \eg in the top example,
        its relevance in the prediction process even increases,
        and the concept removal seems to focus mostly on the eyes and hairline. 
        Contrary to class ``male'',
        an application of \gls{pclarc} is as successful for samples from class ``female''.
        This brings up a possible issue with using \glspl{cav} to represent the target \gls{ch} artifacts that we have previously only briefly touched upon:
        within the Adience dataset,
        the ``shirt collar'' \gls{ch} only has a significant presence within class ``male''
        -- leading to positive and negative ``shirt collar'' examples for the \gls{cav} computation only being obtainable in a reliable manner from samples of class ``male''.
        However, 
        because the \gls{cav} is only computed using samples from one class,
        and because its ability to distinguish a concept relies entirely on the data used for fitting the corresponding linear classifier, 
        it does not necessarily encode the target \gls{ch} as precisely when faced with samples from class since the domain changes for the \gls{cav} model.
        For the samples from class ``female'' without shirt collar features,
         no shirt collar is found and consecutively not removed (similar to the corresponding ``male'' samples).
        In the second example in Figure~\ref{fig:capimage_adience}~\textit{(III)} \textit{(bottom right)} the shape of the long hair seems to be identified as a shirt collar,
        showcasing another issue for this specific \gls{ch} among samples belonging to class ``female''.
       
        To summarize,
        while the concept suppression of \gls{pclarc} seems to have a similar success on the ``male'' class as we previously found for \gls{ch} in other datasets,
        albeit slightly more noisy due to the complex nature of the Adience dataset,
        applying it to the ``female'' class sheds light on various issues,
        \eg,
        a relatively strong domain dependence of the computed \glspl{cav}.
        
        Even though the previous results are relatively mixed,
        we evaluate the ability of \gls{pclarc} to achieve \textit{fairer} predictions in Figure~\ref{fig:capimage_adience}~\textit{(IV)}. 
        Here, the table to the right shows
        for layers 0, 4, and 10
    and both classes the percentage of previously mispredicted (false negatives, \ie, FN) and correctly predicted samples (true positives,\ie, TP) of which the predicted class changed after an application of \gls{pclarc},
        turning them into true positives and false negatives,
        respectively.
        Relatively,
        more false negatives turn into true positives when \gls{pclarc} is applied.
        Where we found the computed \gls{cav} for layer 0 to not be meaningful \wrt the target class,
        the FN to TP rate is comparatively high with $15.7\%$ for class ``male'' and $7.6\%$ for class ``female''. 
        At the same time,
        however,
        the TP to FN rate is also significant,
        with $0.7\%$ for class ``male'' and $2.1\%$ for class ``female''.
        In layer 4, 
        they decrease to $7.2\%$, 
        $0.3\%$, 
        $5.3\%$ , 
        and $1.4\%$,
        respectively.
        In layer 10,
        an interesting phenomenon occurs,
        with the rates growing to $8.4\%$ (FN to TP) and $0.4\%$ (TP to FN) for class ``male'',
        but still diminishing for class ``female'', 
        to $3.1\%$ and $0.9\%$.
        A large amount of samples changing from TP to FN and vice versa is not necessarily a sufficient measurement on its own,
        because many alterations to the model's inference process would have that effect,
        especially since with an accuracy of 94.02\%,
        there are far more TP than FN absolutely.
        \Eg,
        this seems to happen for \gls{pclarc} with a badly encoded \gls{ch},
        as is the case for layer 0,
        according to our findings in Figure~\ref{fig:capimage_adience}~\textit{(I)-(III)}.
        However,
        both (TP to FN) and (FN to TP) rates seem to steadily diminish with higher layers,
        presumably due to alterations later in the network not being propagated as far and thus having a lessened effect,
        except -- as noted above -- for layer 10 of (only) the class ``male'',
        where a sudden increase occurs.
        This observation corresponds to our two previous assertions,
        that the layer 10 \gls{cav} and \gls{pclarc} process for class ``male'' is quite precise \wrt the target concept ``shirt collar'' -- although some other correlating distinct ``male'' features are also affected.
        For class ``female'', however,
        the same artifact does not seem to be as well defined.
        
        In any case,
        a closer look at the affected samples is needed to come to a conclusion.
        For this purpose, 
        Figure~\ref{fig:capimage_adience}~\textit{(IV) (left)} shows
        examples of which the prediction switched after applying \gls{pclarc} in layer 4 are shown,
        along with the softmax probabilities of the respective samples before and after the attempted correction \wrt the \gls{ch} concepts, together with the corresponding attribution difference maps, for classes ``male'' and ``female'' and both types of prediction change.
        For the class ``male'',
        samples seem to be predicted from TP to FN \textit{(bottom left)} due to the target concept,
        \ie,
        ``shirt collar'' or correlating male features like uncovered ears,
        being suppressed successfully. 
        The accompanying change in softmax probabilities is quite significant,
        especially for the first example.
        Interestingly, in the second example,
        the female face visible in the image gains in attributed relevance due to the removal of features
        corresponding to class ``male''.
        Furthermore,
        the change from FN to TP \textit{(top left)} appears to happen due to more significance being attributed to facial features,
        and less to surrounding features.
        Interestingly,
        in the top example,
        part of a ``shirt collar'' is removed,
        but confidence for ``male'' is \textit{increased},
        perhaps due to the colorful expression of the visible clothing item.
        Again, 
        we note significant changes in the predicted class probabilities.
        In contrast,
        for class ``female'' ,
        probabilities often seem to only change slightly and due to the model having difficulties classifying the sample in the first place,
        as is the case,
        \eg,
        for small children (\textit{top right and bottom right}, first sample each).
        However,
        we also observe changes from FN to TP due to a shirt collar feature being withheld from the model (\textit{top right} second image),
        although the shirt collar removed here is a misinterpreted pearl necklace,
        and the corresponding alterations in relevance are by far not as distinct as for the examples labelled as ``male''.
        Even so,
        the accompanying discrepancies in softmax probabilities are notably higher for examples such as this,
        where the classification changes due to valid (\wrt the targeted \gls{ch}) reasons.
        
        To summarize,
        on the Adience dataset -- which is admittedly quite difficult to solve, due to its various inherent biases and imbalances -- ,
        we found that the influence of even highly complex \gls{ch},
        \eg,
        the ``shirt collar'' of class ``male'',
        can be successfully mitigated via \gls{pclarc},
        although not quite as precisely and significantly as achieved for,
        \eg,
        the ISIC~2019 dataset.
        Especially the issue of \gls{pclarc} not being transferable between classes without losing in precision of the \gls{ch} correction becomes clear
        if a concept is present within multiple classes
        but the \gls{cav} representation is only learned from samples of a single class.
        This,
        however,
        seems to be a problem of the representation only being computed from samples of one class -- due to a sufficient number of examples expressing the \gls{ch} sufficiently well only being available from the target class -- ,
        not the \gls{pclarc} method itself.
        Finding more accurate and generalizing representations is subject to future work.
        In terms of fairness,
        we conclude that for the target class, 
        the predictions after applying \gls{pclarc} become more focused on the desired features,
        leading to classifications for the right reasons.
        For the other class,
        this is not always the case due to the representation issue stated above,
        however, 
        \emph{if} the concept is detected and suppressed correctly,
        the resulting difference in predicted probabilities is far more significant.

\section{Conclusion}
    Deep Learning models have gained high practical usability by pre-training on large corpora and then reusing the learned representation for transferring to novel related data.
    A prerequisite for this practice is the availability of large sets of rather standardized and, most importantly, representative data.
    If artifacts or biases are present in data, then the representations formed are prone to inherit these flaws.
    This is clearly to be avoided, however, it requires either clean data or detection and subsequent removal of the influence of artifacts, biases \etc~of data bases that would cause dysfunctional representation learning.
    In this paper we have used techniques from \glsdesc{xai} (\eg, \gls{lrp}~\cite{bach2015pixel} and \gls{spray}~\cite{lapuschkin2019unmasking} with several meaningful extensions),
    and introduced the \glsdesc{clarc} framework to scalably and automatically detect,
    validate and alleviate \glsdesc{ch} behavior in multiple recent and large data corpora.
    While we mainly used \gls{lrp},
    the proposed \gls{clarc} framework is independent of the particular \gls{xai} method. 
    \Gls{clarc} encompasses a first simple intuition based  of how artifacts may harm generalization.
    As this intuitive model is based on logistic regression,
    it is rather crude,
    but it already shows the main effects caused by artifacts:
    deterioration of generalization ability.
    For neural networks it may, however,
    still serve as a reasonable guideline and indeed our large-scale experiments on various datasets show analogous effects, that can exhibit a dramatic drop of generalization for some classes.
    Based on the \gls{clarc} model of artifactual features,
    we have introduced two concrete algorithms to implement the desensitization and unlearning of undesired features in a deep neural network:
    First, we proposed \gls{aclarc}, an approach building on strategic augmentation of the data and subsequent fine-tuning of the model in order to remove the influence of artifactual confounders from inference.
    Second with, we aim at \gls{pclarc} suppressing the the representation of an artifact as a feature to prevent its use in inference.
    While the latter approach is extremely efficient as it does not involve any training beyond the modeling of the artifact itself, the former can drive the model to adapt to a different, benign set of features. 
    Both approaches can be applied on artifact representations obtained in input spaces, as well as latent space.
    
    Let us discuss the main experimental findings.
    Based on an extended \gls{spray} technique we could in toy settings verify artificially created \glsdesc{ch} artifacts, and
    automatically detect some rather unexpected \glsdesc{ch} strategies of a popular pre-trained VGG-16 deep learning model on ILSVCR2012.
    These are caused by a zoo of artifacts and biases isolated by our framework in the corpus:
    encompassing copyright tags, unusual image formatting, specific co-occurrences of unrelated objects, cropping artifacts, just to name a few.
    Detecting this zoo gives not only insight but also the possibility for relieving models and datasets from their \glsdesc{ch} moments,
    \ie, based on our theoretical findings,
    we are now able, using \gls{clarc}, to implicitly un-Hans large reference datasets such as the ImageNet corpus
    and thus provide a more consistent basis for pre-trained models.
    We demonstrated this in \emph{unlearning experiments} for several artifactual features on ImageNet, and in practical application scenarios, \ie, the ISIC~2019 dataset skin lesion prediction dataset and the Adience benchmark dataset of unfiltered faces, yielding more representative predictors for the tasks.
    In all scenarios, we observe that a precise modeling of the artifact,
    \ie the availability and use of representative data distinguishing artifactual features from desired ones,
    will have a beneficial effect on the success of both \gls{clarc} variants.

    Let us reiterate that without removing, or at least considering such data artifacts, learning models are prone to adopt \glsdesc{ch} strategies \cite{lapuschkin2019unmasking}, thus, giving the correct prediction for an artifactual/wrong reason.
    Once these artifacts are absent or appear in unusal combination with other features in the wild such \glsdesc{ch} models will experience significant loss in generalization (see, \eg, Figures~\ref{fig:experiments:aclarc:imagenet:scatter}, \ref{fig:capimage_isisc2019} and~\ref{fig:capimage_adience}). 
    This makes them especially vulnerable to adversarial attacks that can harvest all such artifactual issues in a data corpus~\cite{carlini2017towards}.
    
    Future work will therefore focus on the important intersection between security and functional cleaning of data corpora, \eg, to lower the attack risk when building on top of pre-trained models.

\section*{Acknowledgements}

We acknowledge Marina H\"ohne for valuable discussion.
This work was supported in part by the German Ministry for Education and Research (BMBF) under grants 01IS14013A-E, 01GQ1115, 01GQ0850, 01IS18056A, 01IS18025A and 01IS18037A.
This work is also supported by the Information \& Communications Technology Planning \& Evaluation (IITP) grant funded by the Korea government (No. 2017-0-001779),
as well as by the Research Training Group ``Differential Equation- and Data-driven Models in Life Sciences and Fluid Dynamics (DAEDALUS)'' (GRK 2433) and Grant Math+, EXC 2046/1, Project ID 390685689 both funded by the German Research Foundation (DFG).

\bibliographystyle{unsrtnat}
\bibliography{main_20201218210202}

\clearpage
\begin{appendix}

    \section{Neural Network Architecture and Training Setups}
    \label{sec:appendix:networks}
    
    \subsection{CIFAR-10 Training}
    \label{sec:appendix:networks:cifar}
    The simple convolutional model used to train CIFAR-10 in \ref{sec:bdhans} consists of two ReLU-activated convolutional-pooling blocks (filter sizes 16 and 32), followed by two dense layers (512 and 10 outputs, respectively).
    The model is trained for 5 epochs using \gls{sgd} with a learning rate of $0.01$ and a momentum of $0.9$.
    
    \subsection{Colored MNIST Training}
    \label{sec:appendix:networks:coloredmnist}
    
    All models on colored MNIST in Sections \ref{sec:experiments:inputspray} and \ref{sec:experiments:unlearning:aclarc} are trained using the AdaDelta algorithm with a learning rate of $1.0$, which is multiplied by $0.7$ after each epoch, for 10 epochs.
    The \emph{a posteriori ClArC} is trained for 10 epochs on top of the \emph{native model}, which has also been trained for 10 epochs.
    The network consists of 2 convolutional layers, followed by a max-pooling, and finally 2 fully connected layers.
    Dropout is used after the max pooling and after the first fully connected layer, with 25 percent and 50 percent dropout probabilities respectively.
    ReLU activations follow all linear layers except the final one.
    
    The model used for \ref{sec:experiments:unlearning:pclarc} is trained with SGD, a learning rate of $0.001$ for 5 epochs. The architecture, however, is the same as for the other colored MNIST models.
    
    \subsection{\gls{aclarc} on ImageNet}
    \label{sec:appendix:networks:aclarcimagenet}
    In Section \ref{sec:experiments:unlearning:aclarc} we employ \gls{aclarc} using a VGG-16 model with the pretrained weights obtained from the Pytorch model zoo. 
    For the input space \gls{aclarc} experiment, we use an Adam optimizer with learning rate $0.0001$ for fine-tuning.
    During feature space\gls{aclarc}, an SGD optimizer with learning rate $0.001$ and momentum $0.9$ is applied.
    In both cases, we fine-tune over 10 epochs.
    
    \subsection{\gls{pclarc} on ISIC 2019 and Adience Training}
    \label{sec:appendix:networks:isicadience}
    We again employ the VGG-16 model in Section \ref{sec:experiments:unlearning:pclarc} with the pretrained weights obtained from the Pytorch model zoo to train on both ISIC 2019 and Adience datasets, replacing the last fully connected layer of the classifier to fit the number of classes, i.e., 9 and 2, respectively. Both models are then trained over 100 epochs, using an SGD optimizer with learning rate $0.001$ and momentum $0.9$.

    \FloatBarrier
\clearpage
\printnoidxglossary[type=\acronymtype]
\end{appendix}
\end{document}